\documentclass{article}

\usepackage[eandd, final]{neurips_2026}

\usepackage[utf8]{inputenc} % allow utf-8 input
\usepackage[T1]{fontenc}    % use 8-bit T1 fonts
\usepackage{hyperref}       % hyperlinks
\usepackage{url}            % simple URL typesetting
\usepackage{booktabs}       % professional-quality tables
\usepackage{amsfonts,amsmath}       % blackboard math symbols
\usepackage{nicefrac}       % compact symbols for 1/2, etc.
\usepackage{microtype}      % microtypography
\usepackage{xcolor}         % colors
\usepackage{siunitx}
\usepackage{graphicx}
\usepackage{csquotes}
\usepackage{subcaption}
\usepackage{microtype}
\usepackage{lscape}
\usepackage{makecell}

\title{Long-Rollout Stability in AI Weather Models: A Quantitative Benchmark and Analysis}

\author{%
  Fanny Lehmann \\
  ETH AI Center\\
  Seminar for Applied Mathematics\\
  ETH Zurich\\
  \texttt{fanny.lehmann@ai.ethz.ch} \\
  \And
  Firat Ozdemir \\
  Swiss Data Science Center\\
  ETH Zurich \& EPFL\\
  \And
  Yun Cheng\\
  Swiss Data Science Center\\
  ETH Zurich \& EPFL\\
  \And
  Torsten Hoefler \\
  Scalable Parallel Computing Lab\\
  ETH Zurich\\
  \And
  Sebastian Schemm \\
  Dep. of Applied Mathematics and Theoretical Physics\\
  University of Cambridge\\
  \And
  Benedikt Soja \\
  Institute of Geodesy and Photogrammetry\\
  ETH Zurich\\
  \And
  Siddhartha Mishra\\
  Seminar for Applied Mathematics\\
  ETH Zurich\\
}

\begin{document}

\maketitle

\begin{abstract}
    While AI weather models excel at short-to-medium range forecasts (up to 15 days), they frequently suffer from ill-defined \textquote{instabilities} when rolled out over longer horizons. 
    This work addresses the lack of a formal taxonomy by categorizing these failures into three distinct regimes — blow-up, loss of seasonality, and small-scale energy exceedance — through year-long rollouts of eight state-of-the-art AI weather models.
    Our analysis reveals that stability hinges on the treatment of small spatiotemporal scales: unstable models amplify high-frequency energy, while stable models act as denoisers when noise is added to their inputs. Far from reducing these models to mere stochastic \textit{parrots}, our findings highlight that stable models generate unique weather trajectories, conditioned on the initial state. 
    Ablation studies using state-of-the-art Vision Transformer (ViT) and Spherical Fourier Neural Operator (SFNO) identify absolute time information, long skip connections, absolute targets, and sufficient model capacity as key factors for stable and meaningful long rollouts. 
\end{abstract}

\section{Introduction}
    AI weather models have surpassed numerical solvers on standard metrics at weather time scales (up to 15 days), but producing realistic predictions over longer horizons remains a fundamental challenge.
    A common approach for long-term prediction applies the model autoregressively, feeding its own output back as input at each step. This yields rollouts of arbitrary length, but predictions frequently \textit{blow up} or become unrealistically blurry as the number of autoregressive steps grows, and no formal characterization of these failure modes exists \cite{bonev_spherical_2023, bonavita_limitations_2024}. 

    While long-term stability remains elusive for many AI weather models, recent climate-scale emulators demonstrate remarkable autoregressive robustness, an effect that appears architecture-agnostic. Success has been observed across a variety of backbone architectures, including Spherical Fourier Neural Operators (ACE2 \cite{watt-meyer_ace2_2025}, LUCIE \cite{guan_lucie_2024}), UNet Convolutional Neural Networks (DLESyM \cite{cresswell-clay_deep_2025}), and 3D Swin Transformers (ArchesWeather \cite{couairon_archesweather_2024}).  
    Typically, the main difference between AI weather models and AI climate emulators is the spatiotemporal resolution of the data with which they are trained and operate.
    While most state-of-the-art weather models operate at a spatial resolution of \ang{0.25} and temporal resolution of 6 hours, climate emulators lower the spatial resolution to \ang{1.0}, at best (see Table~\ref{tab:sota_models}) and sometimes increase the lead time to one day or more. This work investigates the influence of spatiotemporal resolution on long-term stability as a potential explanation for the success of climate emulators on decadal time scales.

    Prior work has proposed several hypotheses to explain long-rollout instability. Some works suggest that models need to account for the spherical structure of the Earth to correctly represent all spatial scales, using spherical harmonics or HEALPix meshes for this purpose \cite{bonev_spherical_2023, karlbauer_advancing_2024}. Chattopadhyay et al. identify spectral bias as the root cause of instability \cite{chattopadhyay_challenges_2024}. They propose a new framework (FouRKS) for spectral regularization but only demonstrate its benefits on low-resolution data (\ang{2} and 24h lead time), leaving it unclear whether these benefits generalize to the high-resolution (\ang{0.25}) regimes common to state-of-the-art AI weather models.
    
    AI Earth system models are increasingly positioned as tools for studying climate change, given their demonstrated ability to reproduce atmospheric dynamics. Beyond the physical constraints required to design meaningful climate emulators, a prerequisite for any climate application is that models sustain long rollouts without unphysical drift, blow-up, or spectral artifacts. This work supports active efforts in the AI for Weather and Climate community to address time scales longer than weather, such as the Weather Quest targeting subseasonal to seasonal time scales \cite{loegel_ai_2025} and the AIMIP experiment focusing on climatic time scales \cite{henn_aimip_2026}.

\section{Related Work}
    Several AI models achieve decadal stability: ACE2 sustains rollouts up to 1000 years \cite{watt-meyer_ace2_2025} and NeuralGCM up to 40 years \cite{kochkov_neural_2024}. However, this stability is not unconditional: NeuralGCM yielded stable 40-year rollouts for only 22 of 37 tested initial conditions \cite{kochkov_neural_2024}. The Triton model achieves 1-year stability through a multi-grid architecture that addresses spectral bias \cite{wu_advanced_2025}, but its predictions exhibit significant smoothing after several months, degrading physical realism. Challenges related to long-term stability have already been raised but only described visually \cite{chattopadhyay_challenges_2024} or at a low spatial resolution \cite{gallusser_exploring_2025}. McCabe et al. give an in-depth analysis of stability failure modes in FNO and propose several architectural changes related to spectral convolutions that improved stability up to 400 rollout steps \cite{mccabe_stability_2023}. Models retrained at a coarser resolution (\ang{5.625}) with a reduced set of variables can remain stable for several months~\cite{karlbauer_comparing_2026}, although this setting differs from the operational \ang{0.25} checkpoints evaluated here.
	
	Outside of the weather domain, several approaches have been proposed to stabilize autoregressive models. Multi-step loss functions improve long-term stability but are memory-prohibitive for large models, motivating backpropagation through only the final rollout step \cite{brandstetter_message_2022, zhou_improving_2025}.
    Instability often arises when autoregressive predictions drift out of the training distribution. Robustness can be improved by injecting noise during rollout \cite{lippe_pderefiner_2023}, denoising predictions via diffusion models as in Thermalizer \cite{pedersen_thermalizer_2025}, or adding regularization terms such as a Jacobian penalty to the loss function \cite{nie_jaws_2026}.
	Finally, Armegioiu highlights the influence of unresolved variables on rollout stability for generative models and shows the benefits of keeping a memory of the physical system to stabilize rollouts \cite{armegioiu_memoryconditioned_2026}.

\begin{figure*}[t]
        \centering
        \includegraphics[width=1\textwidth]{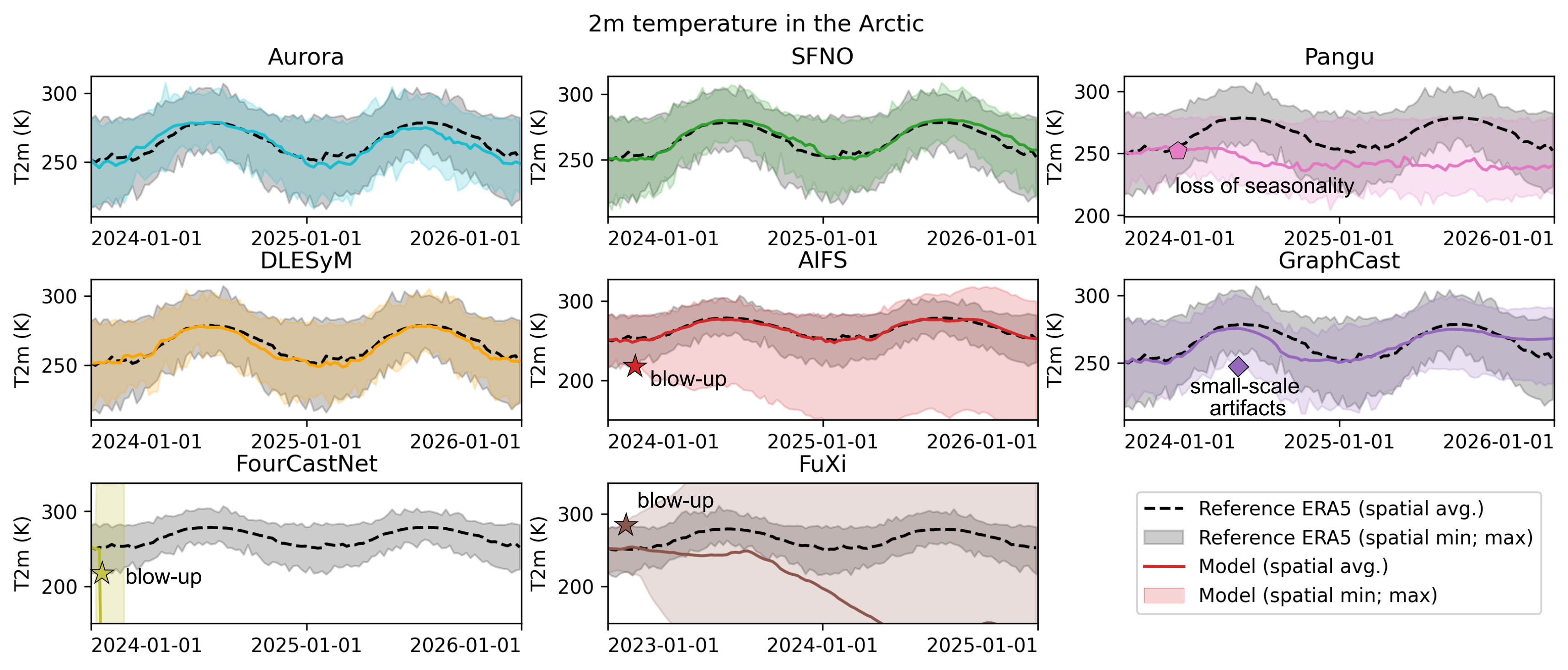}
        \caption{Rollouts initialized on January 1, 2024 for eight AI weather models. The lines show 2-meter temperature averaged across the Arctic for the reference ERA5 (black) and the model (color). The shaded areas extend from the minimum to the maximum temperature over the Arctic for ERA5 (grey) and the model (color). FourCastNet results are shown only until they reach unphysical values. FuXi is initialized on January 1, 2023 due to data availability (Appendix~\ref{sec:model_defs}).}
        \label{fig:rollout_timeseries}
    \end{figure*}
    
\section{A Taxonomy of Rollout Instabilities}
\label{sec:definitions}
    We analyze eight state-of-the-art weather models spanning model sizes from 3.5M to 4.7B parameters and four architecture families (see Table~\ref{tab:sota_models}). For each model, we run a 2-year rollout (2,920 autoregressive steps at 6-hour lead time), initialized on January 1, 2024 — outside the training period of all models. Sensitivity to the initial conditions is examined in Appendix~\ref{appendix:sensitivity_metrics}. The ERA5 reanalysis \cite{hersbach_era5_2020} serves as reference data, on which all models are trained. 

    Figure~\ref{fig:rollout_timeseries} shows 2-meter temperature time series over the Arctic, a particularly sensitive diagnostic region where spherical geometry compresses grid spacing at high latitudes, often causing instabilities to manifest there before propagating to lower latitudes. Figure~\ref{fig:rollout_timeseries} reveals different behavior across models: Aurora, SFNO, and DLESyM track the ERA5 seasonal cycle closely throughout, while the minimum temperature predicted by AIFS, FourCastNet, and FuXi diverges rapidly from ERA5.

    \begin{table*}[t]
      \caption{Number of days until the model predictions blow up, defined as an exponential growth of the spatial extremes (see Appendix~\ref{sec:def_blow_up} for details). \textquote{-} denotes variables absent from the models. Sensitivity to the initial date is reported in Table~\ref{tab:blowup_sensitivity}.}
      \label{tab:blowup}
      \begin{center}
        \begin{small}
            \begin{tabular}{lccccccccr}
              \toprule
              Model & T2m & U10m & MSLP & Z500 & T500 & Q500 & U300 & T100 & Q100 \\
              \midrule
                Aurora & $>$ 730 & $>$ 730 & $>$ 730 & $>$ 730 & $>$ 730 & $>$ 730 & $>$ 730 & $>$ 730 & $>$ 730 \\
                SFNO & $>$ 730 & $>$ 730 & $>$ 730 & $>$ 730 & $>$ 730 & $>$ 730 & $>$ 730 & $>$ 730 & $>$ 730 \\
                Pangu-Weather & $>$ 730 & $>$ 730 & $>$ 730 & $>$ 730 & $>$ 730 & $>$ 730 & $>$ 730 & $>$ 730 & $>$ 730 \\
                AIFS & 268 & $>$ 730 & $>$ 730 & 21 & $>$ 730 & $>$ 730 & $>$ 730 & 567 & 515 \\
                GraphCast & 370 & 453 & 352 & 326 & 315 & 325 & 346 & 338 & 359 \\
                FourCastNet & 10 & 10 & 10 & 12 & 11 & - & - & - & - \\
                DLESyM & $>$ 730 & - & $>$ 730 & $>$ 730 & - & - & - & - & - \\
                FuXi & 5 & 8 & $>$ 730 & 158 & 33 & - & 14 & 2 & - \\
              \bottomrule
            \end{tabular}
        \end{small}
      \end{center}
      \vskip -0.1in
    \end{table*}

    Beyond visual inspection, we introduce quantitative metrics that systematically characterize rollout behavior across variables and spatial scales. Although results are presented for one rollout, Tables~\ref{tab:blowup_sensitivity} and \ref{tab:seasonality_sensitivity} show that metrics are robust to different initialization dates.
    
    \textbf{The blow-up time} is defined as the onset of unbounded exponential growth in spatial extremes. Specifically, we identify the first 30-day window where the global minimum or maximum value yields a high log-linear correlation, signifying a transition from physical oscillation to numerical divergence (Appendix~\ref{sec:def_blow_up}). 
    Table~\ref{tab:blowup} extends the observations from Fig.~\ref{fig:rollout_timeseries} by showing that Aurora, SFNO, Pangu-Weather, and DLESyM do not blow up for any of the nine variables over different pressure levels. AIFS behavior is inconsistent across variables, with geopotential at 500~hPa consistently blowing up within 50~days at the poles while other variables blow up after nine months if at all (Table~\ref{tab:blowup_sensitivity}, Fig.~\ref{fig:blowup_aifs_z500}). We noticed earlier blow-up with the probabilistic AIFS-CRPS \cite{lang_aifscrps_2024} (Fig.~\ref{fig:aifs_ens_2t}).

    \textbf{Seasonality} is quantified from the decomposition of the Mean Squared Error (MSE) 
    \begin{equation}
    MSE = \overline{(\hat{u} - u)^2} = \overline{(\hat{u} - c)^2} + \overline{(u - c)^2} - 2 \overline{(\hat{u} - c)(u - c)}
    \label{eq:decompose_mse}
    \end{equation}
    where $\hat{u}$ denotes predictions, $u$ the ERA5 reference, $c$ climatology, and $\overline{\hspace{0.25em}\cdot\hspace{0.25em}}$ a spatial average \cite{murphy_skill_1988}. The variance ratio $\overline{(\hat{u} - c)^2} / \overline{(u - c)^2}$ quantifies the model agreement with climatology. Based on a variable-dependent threshold determined from shifted time series of ERA5, we define the loss of seasonality as the first day when the model variance ratio exceeds the threshold (details in Appendix~\ref{sec:def_seasonality}). 
    This definition confirms that Pangu-Weather rollouts lose seasonality and collapse to a time-invariant global state (Table~\ref{tab:seasonality} and Fig.~\ref{fig:rollout_timeseries}), likely because Pangu-Weather lacks a time embedding and therefore has no access to long-term temporal information. 
    Although counter-intuitive, AIFS and GraphCast can preserve seasonality while blowing up since blow-up manifests locally, whereas seasonality is computed from large-scale mean climatological deviations that are insensitive to spatially limited effects.
    Top-of-atmosphere variables (above 100~hPa) are the only case where we observed stability failure modes for Aurora (Fig.~\ref{fig:rollout_q100}) and, to a lesser extent, SFNO (Table~\ref{tab:seasonality_sensitivity}).

    \begin{table*}[t]
      \caption{Number of days until loss of seasonality, defined as exceedance of the variance ratio (see the definition in Appendix~\ref{sec:def_seasonality} and sensitivity in Table~\ref{tab:seasonality_sensitivity}). \textquote{-} denotes absent variables.}
      \label{tab:seasonality}
      \begin{center}
        \begin{small}
            \begin{tabular}{lccccccccr}
              \toprule
              Model & T2m & U10m & MSLP & Z500 & T500 & Q500 & U300 & T100 & Q100 \\
              \midrule
                Aurora & $>$ 730 & $>$ 730 & $>$ 730 & $>$ 730 & $>$ 730 & $>$ 730 & $>$ 730 & 467 & 161 \\
                SFNO & $>$ 730 & $>$ 730 & $>$ 730 & $>$ 730 & $>$ 730 & $>$ 730 & $>$ 730 & $>$ 730 & $>$ 730 \\
                Pangu & 126 & $>$ 730 & 191 & 137 & 121 & 533 & 161 & 257 & 260 \\
                AIFS & 516 & $>$ 730 & $>$ 730 & 588 & 597 & 530 & $>$ 730 & 496 & 524 \\
                GraphCast & 156 & $>$ 730 & 265 & 165 & 147 & 670 & 196 & 155 & 178 \\
                FourCastNet & 15 & 17 & 18 & 18 & 16 & - & - & - & - \\
                DLESyM & $>$ 730 & - & $>$ 730 & $>$ 730 & - & - & - & - & - \\
                FuXi & 109 & 180 & 148 & 147 & 119 & - & 151 & 70 & - \\
              \bottomrule
            \end{tabular}
        \end{small}
      \end{center}
      \vskip -0.1in
    \end{table*}

    \begin{table*}[t]
      \caption{Number of days until the small-scale energy spectrum exceeds a climatological threshold (see the definition in Appendix~\ref{sec:def_artefacts} and sensitivity in Table~\ref{tab:smallscale_sensitivity}). \textquote{-} denotes absent variables.}
      \label{tab:small_scales}
      \begin{center}
        \begin{small}
            \begin{tabular}{lccccccccr}
              \toprule
              Model & T2m & U10m & MSLP & Z500 & T500 & Q500 & U300 & T100 & Q100 \\
              \midrule
                Aurora & $>$ 730 & $>$ 730 & $>$ 730 & $>$ 730 & $>$ 730 & $>$ 730 & $>$ 730 & $>$ 730 & $>$ 730 \\
                SFNO & $>$ 730 & $>$ 730 & $>$ 730 & 142 & $>$ 730 & $>$ 730 & $>$ 730 & $>$ 730 & $>$ 730 \\
                Pangu & $>$ 730 & $>$ 730 & $>$ 730 & $>$ 730 & $>$ 730 & $>$ 730 & $>$ 730 & $>$ 730 & $>$ 730 \\
                AIFS & 226 & $>$ 730 & 257 & 15 & 469 & $>$ 730 & $>$ 730 & 536 & $>$ 730 \\
                GraphCast & 37 & 636 & 22 & 10 & 75 & 456 & 392 & 171 & 438 \\
                FourCastNet & 11 & 15 & 12 & 0 & 8 & - & - & - & - \\
                DLESyM & $>$ 730 & - & $>$ 730 & 14 & - & - & - & - & - \\
                FuXi & 12 & 27 & 11 & 5 & 9 & - & 13 & 14 & - \\
              \bottomrule
            \end{tabular}
        \end{small}
      \end{center}
      \vskip -0.1in
    \end{table*}

    \textbf{The small spatial scales} 
    exhibit one of three distinct behaviors throughout long rollouts:
    (i) they are smoothed out and forecasts become blurry, (ii) they preserve realistic sharpness compared to the reference, (iii) they get amplified and lead to unrealistic small-scale artifacts. The spectral content is a common tool to distinguish these cases \cite{maheshHugeEnsemblesPart2025} that we formalize into a threshold-based metric. The small-scale energy spectra (wavelengths smaller than 400~km) are used to quantify these trends by detecting when the mean small-scale spectrum exceeds the mean small-scale ERA5 spectra over 1990-2019 (Appendix~\ref{sec:def_artefacts}). Table~\ref{tab:small_scales} shows that Aurora, SFNO, Pangu, and DLESyM maintain small-scale energy below the ERA5 threshold for all variables, except geopotential at 500hPa for SFNO and DLESyM. The two graph-based models (AIFS and GraphCast) tend to amplify the high-frequency content, as the majority of variables trigger the small-scale failure mode detection. Due to numerical blow-up, FourCastNet rapidly exceeds the small-scale energy threshold.
    
    However, we also observe that some small-scale artifacts in GraphCast U10m rollout predictions could not be detected in our energy spectra analysis (Fig.~\ref{fig:map_artefacts_graphcast}), mainly due to the limited spatial extent of these perturbations that does not impact the calculation of high-frequency Fourier coefficients.

    \textbf{The set of three metrics is irreducible} as they evaluate different failure modes and give different insights into potential remedies. Failure times depend greatly on the variables (Fig.~\ref{fig:summary_failures}), showing early instability for some variables while others maintain realistic predictions for months, confirming prior work about incorrect correlation between physical variables in AI weather models \cite{bano-medina_are_2025}.

\section{Mechanisms of Stability: Denoising and Generalization}
\label{sec:generalization_denoising}
\subsection{Stability under noise addition}
\label{sec:stability_noise}

    \begin{figure*}[t]
        \centering
        \includegraphics[width=0.85\textwidth]{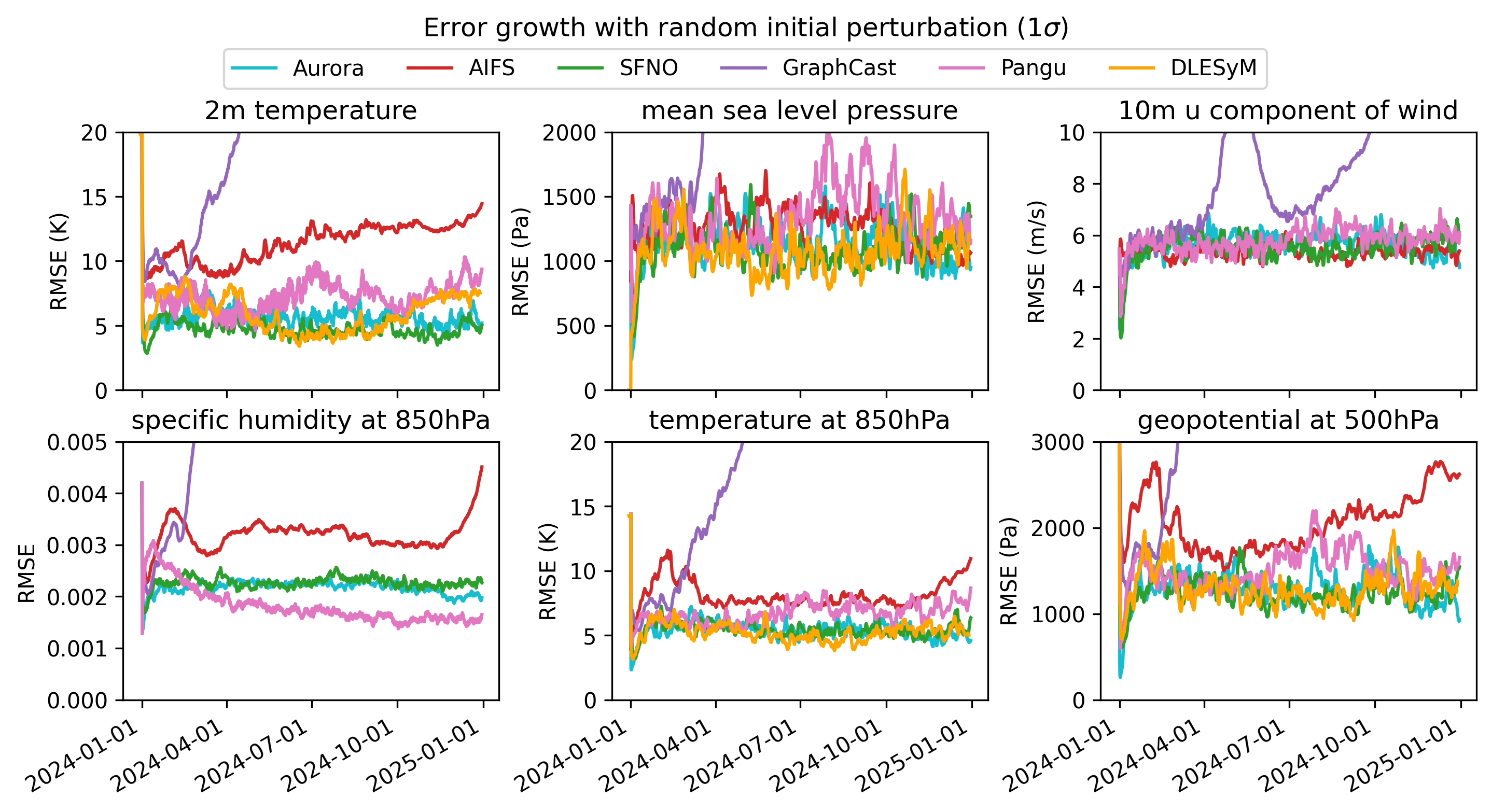}
        \caption{Error between the reference rollout and the rollout initialized with a noisy input (additive noise with $1\sigma$ standard deviation) for six models: Aurora (cyan), AIFS (red), SFNO (green), GraphCast (purple), Pangu (pink), DLESyM (orange), and six variables. The other models diverge rapidly and are shown in Fig.~\ref{fig:rollout_noise_all}.}
        \label{fig:rollout_noise}
    \end{figure*}

    When analyzing the energy spectra of GraphCast and AIFS over time, we find that the instability originates at small spatial scales where the model creates excessive energy content, and propagates to intermediate scales (Fig.~\ref{fig:spectra_graphcast_aifs}). 
    Based on this finding, we perturb the input fields of multiple models by adding a Gaussian noise $\varepsilon \sim \mathcal{N}(0, \sigma^2)$ where $\sigma$ is the standard deviation of each physical variable. We then compare the noisy rollout with the clean rollout to quantify how each model responds to the initial perturbation. Figure~\ref{fig:rollout_noise} shows that Aurora, Pangu, and SFNO stabilize to a constant error, while GraphCast yields an exponentially increasing error early on, and AIFS displays a mixed behavior. However, the models respond very differently to noisy inputs. Aurora and SFNO tend to denoise the inputs and produce physically realistic fields that evolve over time, while Pangu-Weather reduces the noise but converges to physically unrealistic fields, and the predictions of AIFS and GraphCast are severely perturbed by the initial noise (Fig.~\ref{fig:rollout_noise_map_2t}). Notably, the error follows a V-shape over the first ten days: it initially decreases as the model partially denoises the input, then rises as the noisy and clean rollouts diverge toward different trajectories (Fig.~\ref{fig:rollout_noise}). The initial variability decrease is consistent with earlier findings of a lack of \textquote{butterfly effect} \cite{selzCanArtificialIntelligenceBased2023a}. This V-shape marks a transition from dependence on physical initial conditions to fully generative behavior.

    When the initial noise is increased from $\sigma$ to $10\sigma$, the residual between noisy and clean SFNO predictions rapidly diverges, while Aurora and AIFS maintain convergence to a stable value (Fig.~\ref{fig:rollout_noise10}). Pangu-Weather also converges to a stable error comparable to Aurora's, though after a longer denoising period. These results demonstrate that both ViT models (Aurora and Pangu) progressively denoise inputs even under strong white noise and indicate Aurora's superior stability relative to SFNO.  

    Because Aurora is the most robust model in our benchmark, we probe its behavior under increasingly demanding perturbations. 
    Perturbing all dynamic variables (atmospheric and surface), Aurora successfully denoises spatially structured noise and even recovers realistic forecasts when initialized from a cat image (Fig.~\ref{fig:rollout_noise1_corr10_aurora}). Perturbing only the static variables (land-sea mask, soil type, and orography) causes rollouts to degenerate (Fig.~\ref{fig:rollout_noise_decouple}), though Section~\ref{sec:minimal_aurora} shows this reflects Aurora's sensitivity to corrupted inputs rather than an architectural necessity for static variables.

\subsection{Absence of memorization}
    The ability to generate realistic weather from noise warrants a deeper investigation into whether the models have learned generalizable atmospheric dynamics or are merely overfitting to specific sequences in the training set. This section focuses on Aurora and SFNO experiments as the previous section showed that they denoise inputs to realistic fields.
    First, we run five rollouts initialized from different pure white Gaussian noise and compute the pixel-wise standard deviation across the rollouts to assess trajectory diversity. We obtain values comparable to the training set's standard deviation for surface variables, while atmospheric variables are approximately three times smaller (Fig.~\ref{fig:rollout_std}). This sustained variability demonstrates that each randomly initialized rollout generates a unique trajectory, rather than collapsing to a single memorized sequence.

    \begin{figure}[t]
        \centering
        \includegraphics[width=0.8\linewidth]{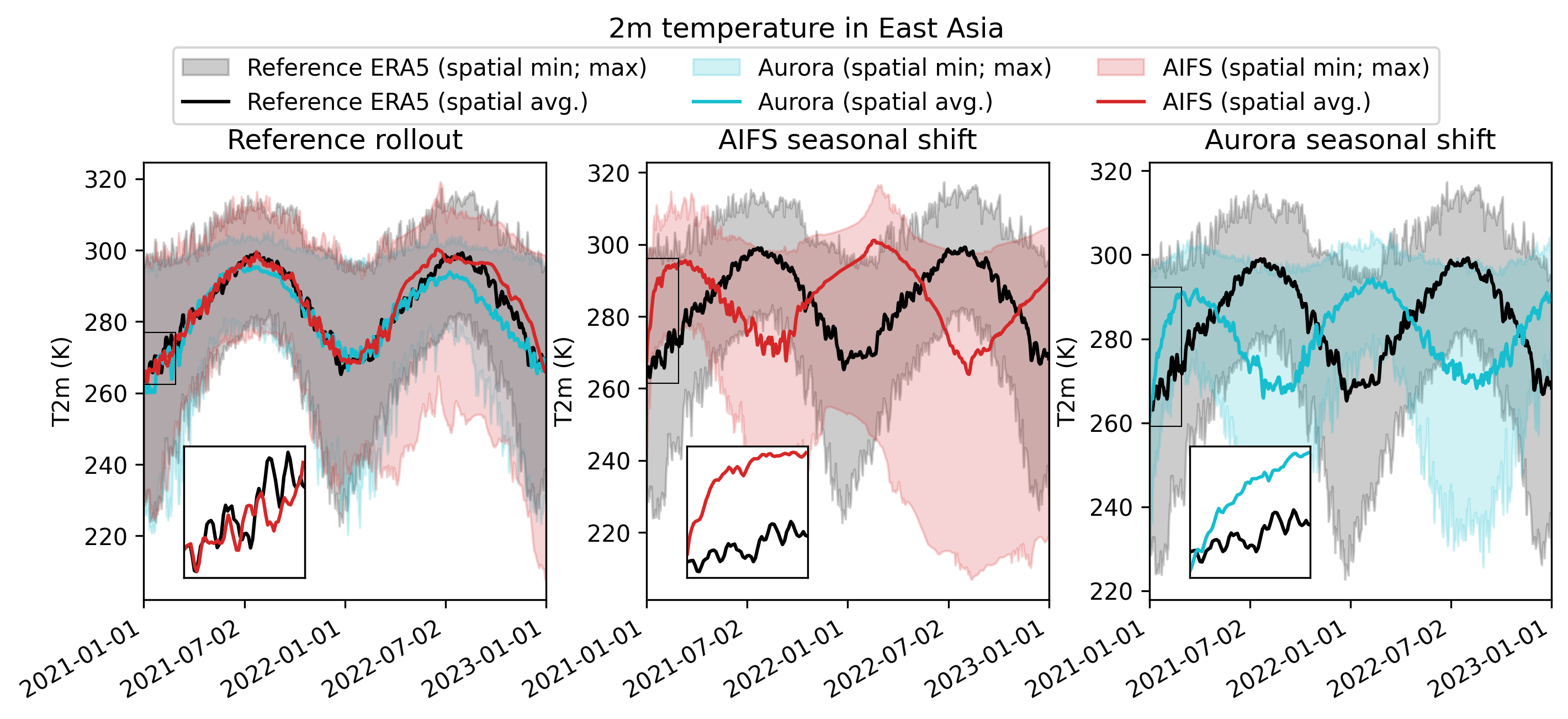}
        \caption{2-meter temperature in East Asia for Aurora and AIFS initialized on January 1, 2021 (left column), AIFS (middle column), and Aurora (right column) initialized with physical values from January 1, 2021 and time information of June 1st, 2021. The solid lines show the regional average, the shaded areas extend from the minimal to the maximal temperature over the region. The subplots show the first two months. Time series are depicted at a daily resolution (at T00) for easier visualization.}
        \label{fig:seasonal_shift}
    \end{figure}

    The second test examines whether stable models simply replay memorized training samples. Following previous work on memorization in diffusion models \citep{bonnaire_why_2025, gu_memorization_2023, yoon2023diffusion}, a sample $X_t$ is said to be memorized if it is significantly closer to its first neighbor in the training dataset than to its second neighbor (Appendix~\ref{sec:memorization}).
    Figure~\ref{fig:distance_ratio} shows that the distance ratio remains close to 1 throughout the one-year rollout for both Aurora and SFNO, indicating no systematic preference for the nearest training neighbor and thus no evidence of memorization. In summary, Aurora and SFNO converge toward the seasonal cycle as a dynamic attractor, but this behavior reflects genuine generalization rather than memorization of training trajectories.

\subsection{Influence of the time embedding}
    Most AI weather models are given time information as input, both the absolute time of the input and the lead time of the prediction. % Aurora encodes absolute time using a Fourier feature expansion of the scalar timestamp (hours since beginning of any given year) followed by a learned linear projection to the embedding dimension. 
    To test whether the time embedding anchors the model's output to the correct season, we conduct a seasonal shift experiment. We initialize 2-year Aurora and AIFS rollouts with physical inputs from January 1, 2024, but provide the model with time information corresponding to June 1, 2024. Figure~\ref{fig:seasonal_shift} shows that the predictions diverge from the reference within a few days, converging instead to a trajectory consistent with the June seasonal cycle.
    This indicates that Aurora and AIFS rapidly adjust the physical state to match the provided time embedding, overriding the January initial conditions within days. 
    In addition, keeping the date fixed throughout the rollout similarly drives the model toward a time-invariant state with no seasonal cycle, confirming that the time embedding drives long-term temporal structure (not shown).

\section{Design Choices for Achieving Realistic Rollouts}
\label{sec:minimal_aurora}
    Having established that Aurora and SFNO exhibit superior long-rollout stability among the eight models studied, we now investigate which design choices drive this behavior through systematic ablation experiments. We train a small Aurora variant (denoted Aurora\textsubscript{S}, 113M parameters) and a small SFNO variant (97M~parameters) from random initialization across multiple configurations. We find that stability is remarkably insensitive to most architectural design choices we tested but does depend on two training choices: skip connections and absolute targets.
    Both models are trained on ERA5 data from 1979 to 2019, with 2020 serving as validation, and 2021-2024 as test set. Training is conducted on 8~GPUs for 60k steps, with a per-GPU batch size of 1, and no model parallelization. Unless otherwise mentioned, they are trained with 6-hour lead time and \ang{0.25} resolution. More details on the models and training are given in Appendix~\ref{appendix:minimal_aurora} and \ref{appendix:minimal_sfno}. 

    \begin{figure*}[t]
        \centering
        \includegraphics[width=0.48\textwidth]{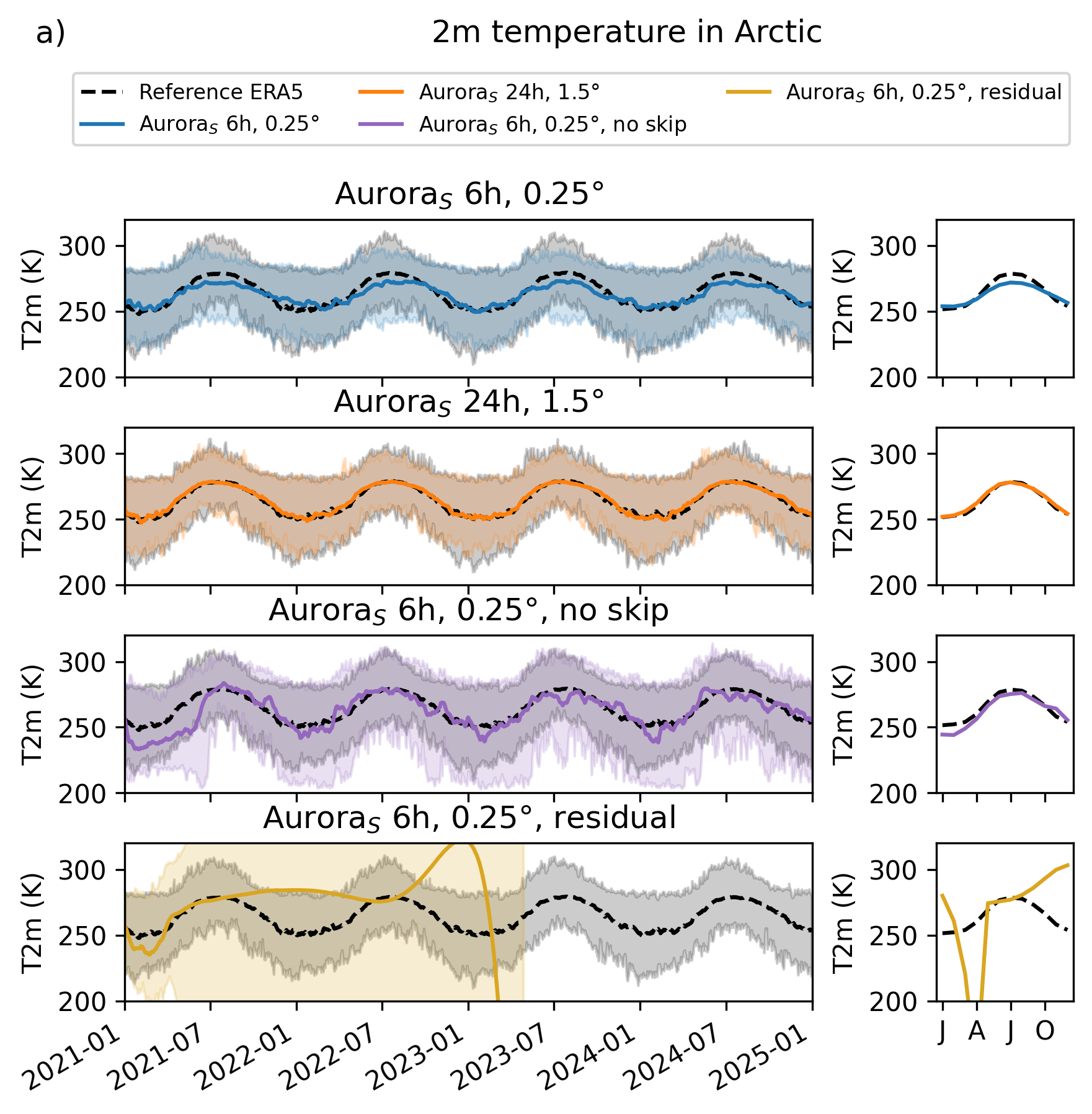}
        \includegraphics[width=0.48\textwidth]{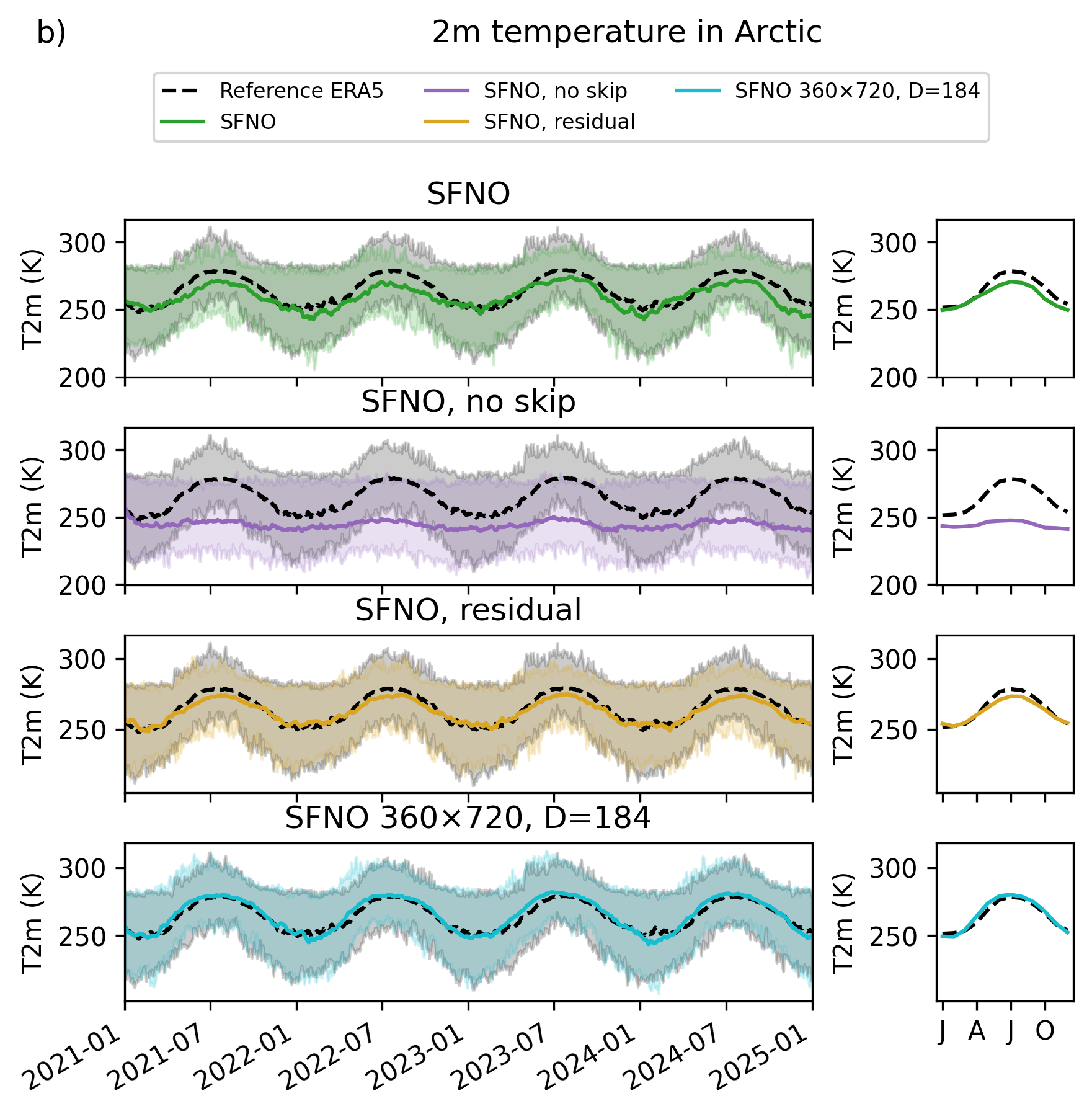}
        \caption{4-year rollout of 2-meter temperature for a) four small Aurora models trained from random initialization with different spatial resolutions and lead times, without skip connection, and with residual objective, b) small SFNO in its baseline configuration, without skip connection, with residual objective, with increased internal grid. The right panels show the monthly seasonal cycle.}
        \label{fig:ablations_aurora_sfno_2t}
    \end{figure*}

\subsection{Residual objective and skip connections: determinant for meaningful stability}
    First, we check that our baseline SFNO and Aurora\textsubscript{S} models trained from random initialization are stable (except T100 for SFNO with some initial conditions, already present in the official SFNO, Table~\ref{tab:blowup_sensitivity}) and preserve the seasonal cycle (Figure~\ref{fig:ablations_aurora_sfno_2t} top panels, Tables~\ref{tab:sfno_ablation_season} and \ref{tab:aurora_ablation_season}). 

    Both Aurora and SFNO contain skip connections, although at different levels of their architecture: two skip connections inside the UNet backbone in Aurora, one \textquote{big} skip connection from the input to the output in SFNO. Training each model without these skip connections leads to a visibly degraded seasonal cycle (Figure~\ref{fig:ablations_aurora_sfno_2t}) and our metric detects a loss of seasonality, as early as 8 days for Aurora\textsubscript{S} T2m and 78 days for SFNO T2m (Tables~\ref{tab:sfno_ablation_season} and \ref{tab:aurora_ablation_season}). However, the absence of skip connections only degrades the validation loss by around 5\% in SFNO and 11\% in Aurora\textsubscript{S}, confirming the weak correlation between 1-step training target and rollout accuracy. 

    We additionally train the baseline models (with skip connections) with a residual target, \textit{i.e.}, predicting $X(t+\Delta t) - X(t)$ instead of the absolute target $X(t+\Delta t)$. This corresponds to the configuration of GraphCast and AIFS, where the model outputs the residual (although the training loss is evaluated on the reconstructed absolute target for AIFS). With a residual objective, Aurora\textsubscript{S} and SFNO show two opposite behaviors: Aurora\textsubscript{S} rapidly blows up (72 days for T2m, Figure~\ref{fig:ablations_aurora_sfno_2t}) while SFNO accuracy improves both on short-range and on the seasonal cycle (Table~\ref{tab:sfno_ablation_season}). This discrepancy may be explained by the presence of the big skip connection in SFNO that gives the model a direct access to the initial value.

\subsection{Architectural invariants: what does not determine stability}
\label{sec:minimal_aurora_ablation} 
    In Aurora\textsubscript{S} architectural ablations, we first vary the components of the Swin3D attention mechanism. Changing the window size from the default (2, 6, 12) to (2, 3, 6) or (2, 12, 24), removing window shifting, or varying the patch size across 1, 2, and 4 all produce stable rollouts with accurate seasonality. % This rules out any explanation of stability based on the specific locality structure or receptive field of the attention mechanism.
    We next test normalization, motivated by the expectation that Layer Normalization prevents value divergence and thus promotes stability. Replacing Layer Normalization with RMSNorm does not affect stability (Fig.~\ref{fig:ablations_aurora_2t}b, second row). Reducing the number of atmospheric levels from 13 to 3 likewise preserves stable and seasonal rollouts, despite substantially reducing the amount of training data and the vertical resolution of the atmospheric representation (Fig.~\ref{fig:ablations_aurora_2t}b, first row).
    
    Static variables (land-sea mask, soil type, and orography) might be expected to play a stabilizing role, since Section \ref{sec:stability_noise} shows that corrupting them at inference time causes full degeneration. However, training Aurora\textsubscript{S} entirely without static variables also produces a stable model that preserves seasonality and retains denoising ability. This apparent contradiction is resolved by noting that the two experiments ask different questions: a model trained without static variables learns not to rely on them, while a model trained with them becomes sensitive to their corruption at inference time.
    Finally, removing the absolute time embedding suppresses all explicit information about long-term dynamics. The rollout does not blow up but the model converges to a physically stable state without seasonality (Fig.~\ref{fig:ablations_aurora_2t}b, third row). Time embedding is therefore not a condition for dynamical stability, but is necessary for climatological accuracy.

    Notably, reducing Aurora\textsubscript{S}'s parameter count from 113M to 21M — by shrinking the encoder depth from (2, 6, 2) to (2, 2, 2) and the embedding dimension from 256 to 128 — eliminates the realistic seasonal cycle (Fig.~\ref{fig:ablations_aurora_2t}b, fourth row), suggesting a minimum model capacity threshold for seasonality. The effect is dominated by the width reduction, as our narrow and deep model (embedding dimension of 128, encoder depth 2, 2, 2) preserves the seasonal cycle for most variables (Table~\ref{tab:season_aurora_width_depth}, Fig.~\ref{fig:ablations_aurora_2t_width_depth}). 

    Regarding SFNO, the architectural ablations consisted of increasing the internal grid, as SFNO stability has been attributed to a truncation of high Fourier coefficients in the spectral layers \cite{bonev_spherical_2023, mccabe_stability_2023}. However, increasing the internal grid from the baseline $180 \times 360$ to the maximum $360 \times 720$ possible at \ang{0.25} resolution did not lead to any blow-up or significant loss of seasonality (only Q500 was detected, Table~\ref{tab:sfno_ablation_season}). Therefore, SFNO confirms the stability insensitivity to many architectural choices.

\subsection{Temporal and spatial scales during training}
\label{sec:minimal_aurora_resolution}
    To examine the influence of spatiotemporal resolution during training, we train Aurora\textsubscript{S} from random initialization on four configurations that combine spatial resolutions \ang{0.25} and \ang{1.5}, and lead times 6~hours and 24~hours. 
    Figure~\ref{fig:ablations_aurora_2t}a shows that the four Aurora\textsubscript{S} models produce stable 4-year rollouts with seasonality. The seasonality metric defined in Section~\ref{sec:definitions} indicates that both 24-hour models preserve accurate seasonality across the full four-year rollout. 
    For the 6-hour models, the metric is conservative and flags a loss of seasonality on T100 after 101 days (\ang{0.25}) and 145 days (\ang{1.5}, Table~\ref{tab:aurora_ablation_season}). The Fourier spectra (Fig.~\ref{fig:spectra_from_scratch}) confirm that \ang{1.5} models produce smoother predictions than \ang{0.25} models, though the degree of smoothing does not significantly worsen during rollout. By contrast, \ang{0.25} models accumulate small-scale spectral energy beyond the ERA5 reference during rollout, though neither blow up. This effect is stronger in the 24-hour model than the 6-hour model.

    \begin{figure*}[t]
        \centering
        \includegraphics[width=0.4\textwidth]{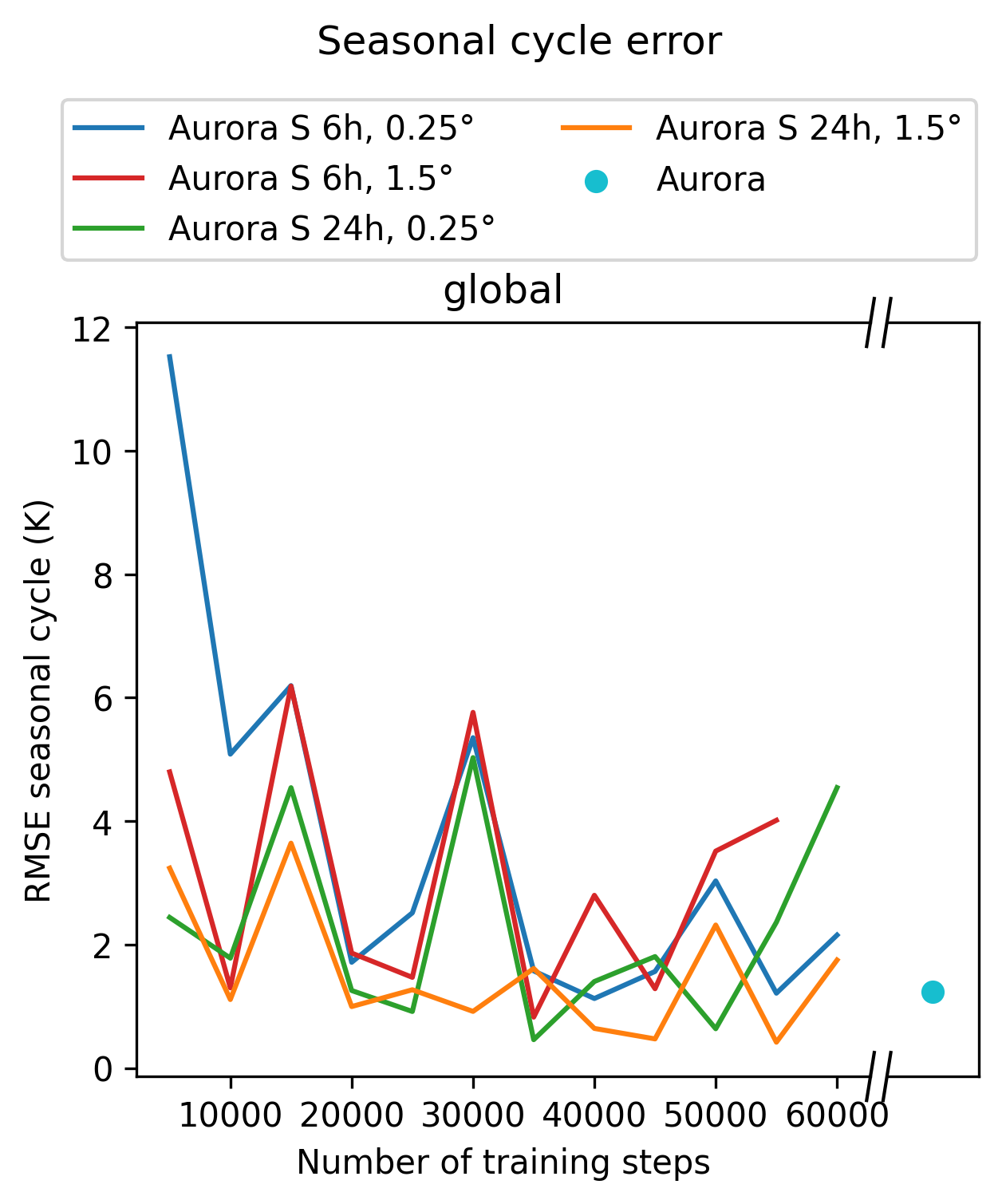}
        \includegraphics[width=0.48\textwidth]{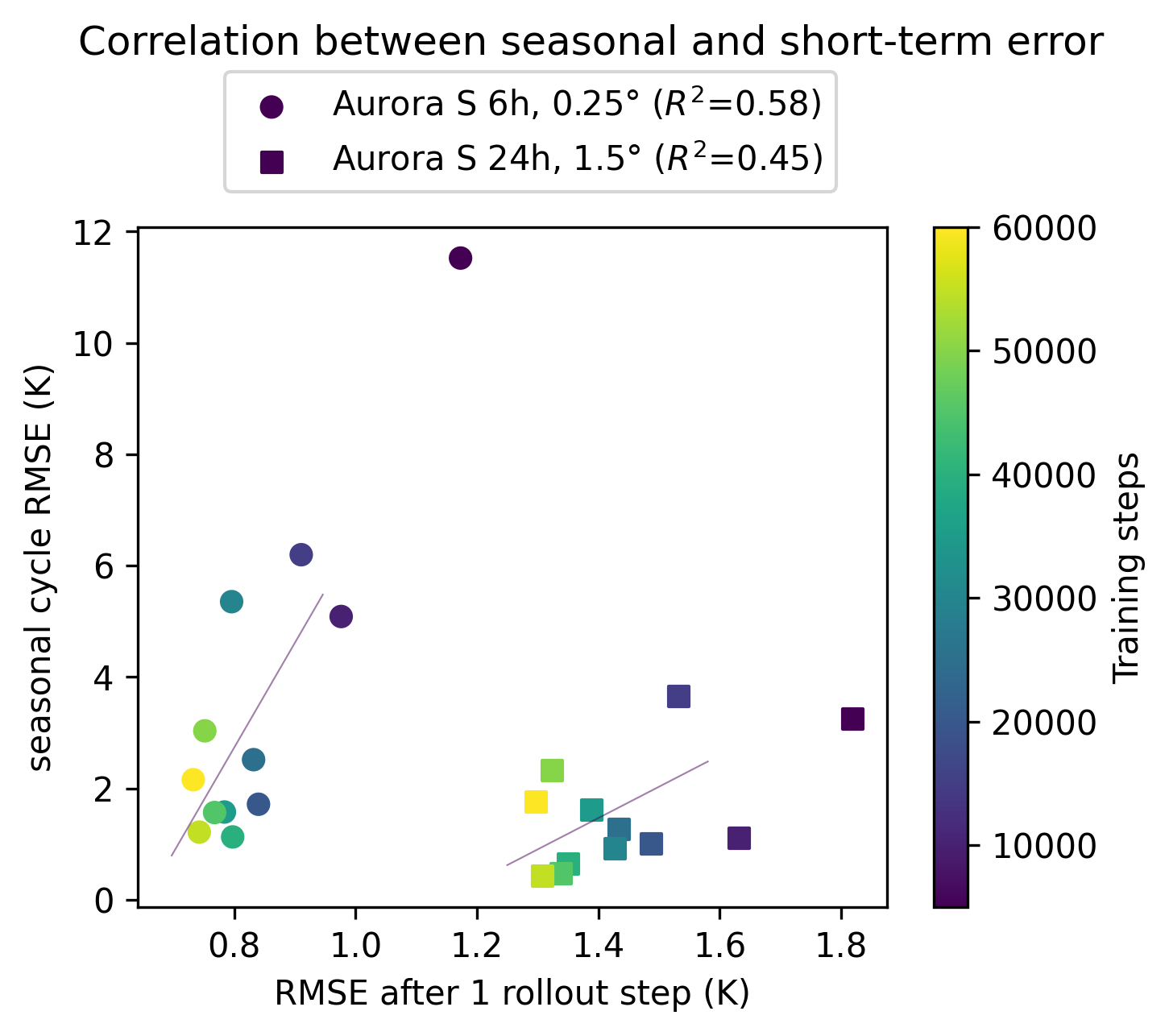}
        \caption{(left) RMSE between the monthly seasonal cycle of the Aurora\textsubscript{S} models and ERA5, as a function of the number of training steps. The seasonal cycle from the large Aurora model \cite{bodnar_foundation_2025} is shown with the cyan dot. (right) On the horizontal axis, 2-meter temperature RMSE after one rollout step (6-hour or 24-hour lead time depending on the model). On the vertical axis, 2-meter temperature seasonal cycle RMSE. Each model is evaluated at different training steps indicated by the colored markers. The linear regression for each model is shown with the thin line and the coefficient of determination is indicated in the legend.}
        \label{fig:seasonal_cycle_cvg}
    \end{figure*}

\subsection{Stability and seasonality training dynamics}
\label{sec:minimal_aurora_dynamics}
    Since all non-ablated Aurora\textsubscript{S} rollouts preserve seasonal fluctuations, we move beyond detecting their loss to quantifying seasonal accuracy. We assess the ability of the model to reproduce long-term behavior by computing the monthly seasonal cycle over the four-year rollout and comparing it with the ERA5 monthly cycle over the same time period (2021-2024). The RMSE between the monthly seasonal cycles then serves as a quantitative measure of how accurate a model is in reproducing the true seasonal cycle. Figure~\ref{fig:seasonal_cycle_cvg} shows that the seasonal cycle error generally decreases as the training progresses, although there can be significant differences between consecutive training steps. All Aurora\textsubscript{S} models match the seasonal accuracy of the full Aurora model (1.3B parameters, \cite{bodnar_foundation_2025}) within 60k training steps, despite the large difference in model capacity. This suggests that long training of the large model does not impact the seasonality of long rollouts, although it is essential to improve short and medium-term accuracy. These results confirm prior observations that model accuracy on the weather time scale is only weakly correlated with climate performance \cite{cachay2024probablistic}. 
    
    The coarsest configuration (24-hour, \ang{1.5}) has the lowest seasonal cycle error and its accuracy is less dependent on the training step. This configuration mirrors the design of most climate emulators, which trade fine-scale detail for long-term stability.
    While Fig.~\ref{fig:seasonal_cycle_cvg}a clearly indicates that seasonal accuracy improves during training, there is only moderate correlation between short-term error and seasonal accuracy of the models at different stages in their training (Fig.~\ref{fig:seasonal_cycle_cvg}b). Short-term error steadily decreases during training, while long-term seasonality is only a byproduct.

\section{Extreme-Event Statistics on Stable Rollouts}
\label{sec:extremes}
    The preceding sections show that several models produce stable rollouts preserving seasonality over year-long time scales. Here we examine whether stable long rollouts preserve the distributional properties needed for climate applications. To this end, we focus on the all-season extreme temperature events derived from 10-year rollouts of Aurora, SFNO, and DLESyM, and compare their statistics with ERA5. All models are initialized on January 1, 2021 and rolled out for 10 years (14,700 steps). The first four years are used as a statistical comparison against ERA5 for the same time period. We define hot (cold) events as time steps where the regional maximum (minimum) exceeds the 90th (falls below the 10th) percentile of the ERA5 climatology; full details are in Appendix~\ref{appendix:extreme_events}.

\begin{figure*}[t]
        \centering
        \includegraphics[width=0.8\textwidth]{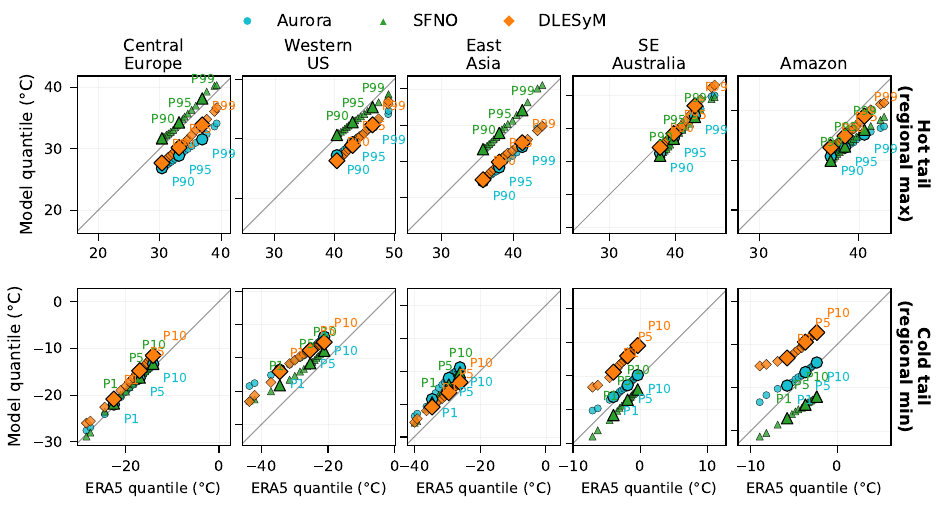}
        \caption{QQ plots comparing ERA5 and the model tail quantiles of regional spatial maximum (hot, top) and minimum (cold, bottom) across five regions. Only points below the 10th and above the 90th percentile are shown.}
        \label{fig:extreme_qq}
    \end{figure*}

    The statistical occurrence of extreme events is evaluated with quantile-quantile plots for hot and cold events (Fig.~\ref{fig:extreme_qq}). For hot events, points below the diagonal indicate that Aurora and DLESyM underestimate the temperature of hot events. For SFNO, results are region-dependent as hot events are underestimated in the Southern regions but overestimated in the three Northern regions under study. For cold events, Aurora and DLESyM have lighter tails, while SFNO exhibits a slight cold bias in the Southern regions (Fig.~\ref{fig:rollout_sfno_extreme}), leading it to produce colder extremes in Australia and the Amazon. A frequency analysis shows that no model consistently reproduces extreme event frequency of ERA5 and no model is clearly better than the others (Fig.~\ref{fig:exceedance}).
    Overall, our results show that Aurora, SFNO, and DLESyM generate physically plausible extreme events during long rollouts, though their magnitude and frequency do not perfectly match ERA5.

\section{Limitations and Conclusion}
    Most work on data-driven weather models targets short- to medium-range accuracy improvements. 
    In this study, we first thoroughly quantify, compare, and benchmark longer rollout performance of eight leading AI weather models. While our metrics provide the first quantitative description of rollout instability and are robust to the choice of initial date, some local unrealistic patterns could not be detected until they degenerate to larger scales. Nevertheless, our metrics could guide the development of AI weather models that aim at long-term predictions. The failure mode first triggering a loss of stability depends on the variable and the model, indicating that our set of metrics is not reducible to a simplified characterization. Each metric also suggests different potential remedies, such as adding a notion of absolute time in Pangu-Weather to correct the loss of seasonality.

    By perturbing the initial conditions with different types of noise, we show that the large stable models (SFNO, Aurora, and Pangu-Weather to a lesser extent) denoise inputs with progressive low-pass filtering to realistic predictions. For Aurora and SFNO, this extends to generalization capabilities from pure noise, toward a seasonal attractor that yields diverse realistic trajectories. 
    
    We then investigate Aurora and SFNO through systematic ablation studies on architecture choices and training procedures. Both models are robust to individual architectural changes, suggesting that the core architecture offers denoising abilities that maintain stable long rollouts.
    The efficiency of Aurora\textsubscript{S} training is consistent with Willard et al. \cite{willard_analyzing_2025}, who demonstrate rapid convergence of 2D Swin v2 models; both that work and the present study nonetheless capture only a partial view of model capabilities. Our ablations point to four practical design choices for stable and meaningful long rollouts: absolute time information, long skip connections, absolute targets, and sufficient model capacity. Long skip connections are crucial for seasonal accuracy in both models, while Aurora results show that predicting residuals instead of absolute targets leads to blow-up. This could be one potential explanation behind the observed instability of AIFS and GraphCast, although we did not retrain these models. None of these choices is individually sufficient, and architectures not ablated in this work may require additional ingredients.

    By analyzing consecutive checkpoints of our trained SFNO and Aurora\textsubscript{S} baseline models, we find that none of them leads to blow-up, although they differ in the accuracy of their seasonal cycle. This suggests that the non-divergence of models may be a property of the architecture, while their physical realism may depend on the selected checkpoint, consistent with ACE2 findings \cite{watt-meyer_ace2_2025}.

    The three models yielding stable and physically realistic long rollouts (SFNO, Aurora, DLESyM) produce realistic extreme events over 10-year rollouts, although they tend to be less extreme and less frequent than the reference period from ERA5, as commonly reported \cite{xu_extremecast_2024}. This shows the potential of AI weather models for studying extreme events, though this potential is not yet fully realized given the remaining biases in magnitude and frequency. 

    It is important to note that this work does not address the influence of climate change on long rollouts. Models have been trained on historical data, hence they are not expected to be robust or to give physically realistic results under distribution shifts. Nevertheless, identifying models with stable long-rollout behavior is a necessary first step toward future climate-aware AI systems.

\section*{Acknowledgements}
    This research was supported by the ETH AI Center through an ETH AI Center postdoctoral fellowship to Fanny Lehmann. This work was supported as part of the Swiss AI Initiative by a Grant from the Swiss National Supercomputing Centre (CSCS) under project IDs a122, a0196, and ab016 on Alps. 

\section*{Impact Statement}
    This work advances the understanding of long-term stability in AI weather models, a necessary step toward their use for climate-scale applications. Improved long-rollout stability could benefit society by enabling more accessible and computationally efficient tools for studying extreme weather events and climate variability. We do not anticipate direct negative societal impacts from this work, as it is foundational research focused on model diagnostics rather than deployment. However, we note that AI weather models trained on historical data are not designed to account for climate change, and their misuse for climate projections without appropriate caveats could lead to misleading conclusions.

\section*{Data and Code Availability}
    The code to compute the stability metrics is available at \url{https://github.com/lehmannfa/ai-weather-stability}. The publicly available ERA5 dataset was downloaded from WeatherBench2 \cite{rasp_weatherbench_2024}. The inference of all models is done through Nvidia Earth2Studio platform \url{https://github.com/NVIDIA/earth2studio} version 0.19.0, which guarantees that all perturbations are applied identically across models. Training of the small Aurora is done from the official repository \url{https://github.com/microsoft/aurora} with the training script replicated from \url{https://github.com/swiss-ai/ESFM}. Training of the small SFNO exactly follows \url{https://github.com/NVIDIA/makani}.
    
% Authors are \textbf{required} to include a statement of the potential broader
% impact of their work, including its ethical aspects and future societal
% consequences. This statement should be in an unnumbered section at the end of
% the paper (co-located with Acknowledgements -- the two may appear in either
% order, but both must be before References), and does not count toward the paper
% page limit. In many cases, where the ethical impacts and expected societal
% implications are those that are well established when advancing the field of
% Machine Learning, substantial discussion is not required, and a simple
% statement such as the following will suffice:

% ``This paper presents work whose goal is to advance the field of Machine
% Learning. There are many potential societal consequences of our work, none
% which we feel must be specifically highlighted here.''

% The above statement can be used verbatim in such cases, but we encourage
% authors to think about whether there is content which does warrant further
% discussion, as this statement will be apparent if the paper is later flagged
% for ethics review.

\clearpage 

\bibliography{references.bib}

%%%%%%%%%%%%%%%%%%%%%%%%%%%%%%%%%%%%%%%%%%%%%%%%%%%%%%%%%%%%%%%%%%%%%%%%%%%%%%%
%%%%%%%%%%%%%%%%%%%%%%%%%%%%%%%%%%%%%%%%%%%%%%%%%%%%%%%%%%%%%%%%%%%%%%%%%%%%%%%
% APPENDIX
%%%%%%%%%%%%%%%%%%%%%%%%%%%%%%%%%%%%%%%%%%%%%%%%%%%%%%%%%%%%%%%%%%%%%%%%%%%%%%%
%%%%%%%%%%%%%%%%%%%%%%%%%%%%%%%%%%%%%%%%%%%%%%%%%%%%%%%%%%%%%%%%%%%%%%%%%%%%%%%
\newpage
\appendix
\setcounter{figure}{0}
\renewcommand\thefigure{A.\arabic{figure}}
\setcounter{table}{0}
\renewcommand\thetable{A.\arabic{table}}

\onecolumn
\section{Weather and Climate Models}
\subsection{Summary of models}

    \begin{table}[h]
      \caption{Non-exhaustive summary of state-of-the-art AI weather and climate models referenced in this work. Spatial resolutions in degrees can be approximated as: \ang{0.25} $\simeq$ 25km, \ang{1.0} $\simeq$ 110km. Training data come from ERA5 if not specified otherwise.}
      \label{tab:sota_models}
      \begin{center}
        \begin{small}
            \begin{tabular}{lp{3.0cm}p{1.7cm}p{2.6cm}p{2.0cm}}
			\toprule
			Model reference & Architecture & \# param. & Spatial res. & Training period \& Lead time  \\
			\midrule
			AIFS \cite{lang_aifs_2024} & Graph Transformer & 253M & N320 grid $\simeq$ 31km & 1979-2022 \newline 6h \\
			Aurora \cite{bodnar_foundation_2025} & 3D Swin Transformer & 1.3B & \ang{0.25} & 1979-2020 \newline 6h \\
            FourCastNet \cite{pathak_fourcastnet_2022} & Adaptive FNO & 75M & \ang{0.25} & 1979-2015 \newline 6h \\
            FuXi \cite{chen_fuxi_2023} & 3D Swin Transformer & 4.7B & \ang{0.25} & 1979-2015 \newline 6h \\
			GraphCast \cite{lam_learning_2023} & Graph Neural Network & 36.7M & \ang{0.25} & 1979-2017 \newline 6h \\
			PanguWeather \cite{bi_accurate_2023} & 3D Swin Transformer and Earth-specific Transformer & 256M & \ang{0.25} & 1979-2017 \newline 1h-24h \\
			SFNO \cite{bonev_spherical_2023} & SFNO & 289M & \ang{0.25} & 1979-2015 \newline 6h \\
            \midrule
            ACE2 \cite{watt-meyer_ace2_2025} & SFNO & 450M & \ang{1.0} & 1940-1995, 2011-2019, 2021-2022 \newline 6h \\
            ArchesWeather \cite{couairon_archesweather_2024} & 3D Swin Transformer & 84M & \ang{1.5} & 1979-2018 \newline 24h \\
            DLESyM \cite{cresswell-clay_deep_2025} & ConvNeXt and GRU & 3.5M (atmos.) & HEALPix 64 ($\simeq$ \ang{1}) & 1983-2016 \newline 6h (atmos.) \\
            DLWP-HPX \cite{karlbauer_advancing_2024} & ConvNeXt and GRU & 9.8M & HEALPix 64 ($\simeq$ \ang{1}) & 1979-2012 \newline 3h \\
            GenCast \cite{price_probabilistic_2024} & Diffusion model & 57M & \ang{0.25} & 1979-2018 \newline 12h \\
            LUCIE \cite{guan_lucie_2024} & SFNO &  & T30 grid ($\simeq$ \ang{3.75}) & 10 years  \\
		      \bottomrule
            \end{tabular}
        \end{small}
      \end{center}
      \vskip -0.1in
    \end{table}

\subsection{Running the models}
\label{sec:model_defs}
    All models were run through Earth2Studio v0.19.0 with their default checkpoint. They are all initialized with the ARCO data source, except FuXi that requires relative humidity at 50hPa and is initialized with the WeatherBench2 data source, and FourCastNet, which requires accumulated precipitation and is initialized with a custom data source downloaded from the Copernicus Data Store. We use AIFS Single v1.1. GraphCast refers to the operational version. PanguWeather is iterated only with the 6-hour model.

    Aurora, AIFS, DLESyM, FourCastNet, GraphCast, PanguWeather, SFNO are initialized on ten dates (2022-01-01, 2022-04-01, 2022-07-01, 2022-10-01, 2023-01-01, 2023-04-01, 2023-07-01, 2023-10-01, 2024-01-01, 2024-04-01) to get initial dates spanning multiple seasons and recent dates that are not in the training period of any model (except the 2022 dates that belong to the training period of AIFS but do not show any difference compared to the later initialization times). Due to availability of the WeatherBench2 data source, FuXi is initialized on 2022-01-01, 2022-04-01, 2022-07-01, 2022-10-01, 2023-01-01. 

    Since AIFS and GraphCast are distributed as operational models, we initialized them with IFS HRES in addition to ERA5 mentioned above. There is no appreciable difference between the stability of ERA5- and IFS-initialized rollouts. GraphCast shows similar small-scale artifacts after two to three months, leading to blow-up within one year. As with the ERA5-initialized rollouts, AIFS preserves stable and seasonal rollouts longer than GraphCast. Its degradation again starts with isolated pixels, which our metric detects as blow-up.

\subsection{Note on GenCast}
    GenCast is a denoising diffusion model trained on ERA5 at \ang{0.25} resolution with 12-hour lead time (referred to as \verb|GenCast 0p25deg <2019| in the original repository, \url{https://github.com/google-deepmind/weathernext/tree/main/docs/weathernext1_gen}). It is distributed through Earth2Studio in the \textquote{mini} version that has a reduced number of parameters and runs at \ang{1.0} resolution (\verb|GenCast 1p0deg Mini <2019|). We found that the \textquote{mini} version is stable. Although we were able to adjust the Earth2Studio code to run the full GenCast model, we noticed that our RMSE increases more rapidly than the results reported on the WeatherBench platform. Therefore, we cannot guarantee that, despite our best efforts, our code adaptation for \ang{0.25} GenCast is perfectly correct and GenCast is not evaluated in this manuscript. GenCast \textquote{mini} is not included due to our focus on models running at \ang{0.25} resolution.

\section{Definitions of stability metrics}
\subsection{Blow-up}
\label{sec:def_blow_up}
    The blow-up of a rollout is defined from the two time series of minimum and maximum values over the globe $\Omega$
    \begin{equation}
        \hat{u}_{min}(t) = \min_{x,y \in \Omega} \hat{u}(x,y,t) \quad ; \quad \hat{u}_{max}(t) = \max_{x,y \in \Omega} \hat{u}(x,y,t)
        \label{eq:spatial_min_max} \text{  ,}
    \end{equation}
    where $\hat{u}$ is the prediction of the physical variable $\mathrm{u}$. As the same procedure is applied to $\hat{u}_{min}$ and $\hat{u}_{max}$, the following describes the computation for $\hat{u}_{min}$ only. First, an average rolling window of 4 days is applied to smooth small-scale fluctuations. Then, a linear regression is applied on the log-transform of $\hat{u}_{min}$ over time windows of 30 days. If the coefficient of determination is larger than 0.9 for a given time window, $\hat{u}_{min}$ exhibits an exponential growth over this time window. Additionally, the rollout should leave the range of historical ERA5 values (minimum ;maximum interval over 1979-2020 widened by 20 \% of its width on each side) within 120 days and stay outside for at least 3 consecutive days. The beginning of the first time window satisfying these criteria is then defined as the time when the model blows up. The reported global blow-up time in Table~\ref{tab:blowup} is the earliest blow-up time between $\hat{u}_{min}$ and $\hat{u}_{max}$. 

    For FourCastNet, the window size for applying the linear regression was reduced to 6 days because the blow-up happens too quickly and the model saturates at some constant value after the blow-up. 

    \paragraph{Choice of hyperparameters} The duration of 4-day smoothing and 30-day time windows was chosen empirically to have a sustained exponential rate. The blow-up time shows limited sensitivity when varying $R^2$ from 0.8 to 0.95, e.g., blow-up is detected between 6 and 11 days for FuXi U10m, 321 and 363 days for GraphCast U300, 17 and 24 days for AIFS Z500 (Table~\ref{tab:blowup_days_R2}). Aurora, SFNO, Pangu, and DLESyM absence of blow-up is left unchanged for the vast majority of variables. Second, the minimum slope to characterize the linear fit as a blow-up was determined from the historical period of ERA5 as the minimum value that does not trigger any blow-up detection on ERA5. The minimum slope is fixed to 0.001, except specific humidity at 100hPa that shows rapid fluctuations and whose minimum slope is fixed at 0.005. 

    \begin{table}[h]
    \centering
    \caption{Number of days before blow-up for each model when varying the coefficient of determination. All rollouts are initialized on January 1, 2024, except FuXi initialized on January 1, 2023. The final $R^2$ chosen in this work is 0.90.}
    \label{tab:blowup_days_R2}
    \setlength{\tabcolsep}{5pt} % default is typically 6pt
    \begin{tabular}{l c| r r r r r r r r r}
    \toprule
    Model & $R^2$ & T2m & U10m & MSLP & Z500 & T500 & Q500 & U300 & T100 & Q100 \\
    \midrule
      AIFS & 0.95 & 393 & $>730$ & $>730$ & 24 & $>730$ & $>730$ & $>730$ & $>730$ & $>730$ \\
     AIFS & 0.90 & 268 & $>730$ & $>730$ & 21 & $>730$ & $>730$ & $>730$ & 567 & 515 \\
     AIFS & 0.85 & 266 & $>730$ & $>730$ & 18 & $>730$ & $>730$ & $>730$ & 565 & 442 \\
     AIFS & 0.80 & 266 & $>730$ & $>730$ & 17 & $>730$ & $>730$ & $>730$ & 545 & 441 \\
    \midrule
     GraphCast & 0.95 & 370 & 458 & 352 & 326 & 317 & 381 & 363 & 341 & 361 \\
     GraphCast & 0.90 & 370 & 453 & 352 & 326 & 315 & 325 & 346 & 338 & 359 \\
     GraphCast & 0.85 & 370 & 431 & 352 & 326 & 313 & 321 & 333 & 332 & 358 \\
     GraphCast & 0.80 & 370 & 403 & 352 & 326 & 312 & 319 & 321 & 328 & 355 \\
    \midrule
     FourCastNet & 0.95 & 10 & 10 & 11 & 13 & 12 & -- & -- & -- & -- \\
     FourCastNet & 0.90 & 10 & 10 & 10 & 12 & 11 & -- & -- & -- & -- \\
     FourCastNet & 0.85 & 5 & 9 & 10 & 12 & 11 & -- & -- & -- & -- \\
     FourCastNet & 0.80 & 5 & 9 & 10 & 6 & 10 & -- & -- & -- & -- \\
    \midrule
     FuXi & 0.95 & 8 & 11 & $>730$ & 158 & 33 & -- & 28 & 3 & -- \\
     FuXi & 0.90 & 5 & 8 & $>730$ & 158 & 33 & -- & 14 & 2 & -- \\
     FuXi & 0.85 & 3 & 7 & $>730$ & 158 & 33 & -- & 9 & 2 & -- \\
     FuXi & 0.80 & 2 & 6 & $>730$ & 158 & 33 & -- & 6 & 2 & -- \\
    \bottomrule
    \end{tabular}
    \end{table}

\begin{figure}[h]
        \centering
        \includegraphics[width=0.75\linewidth]{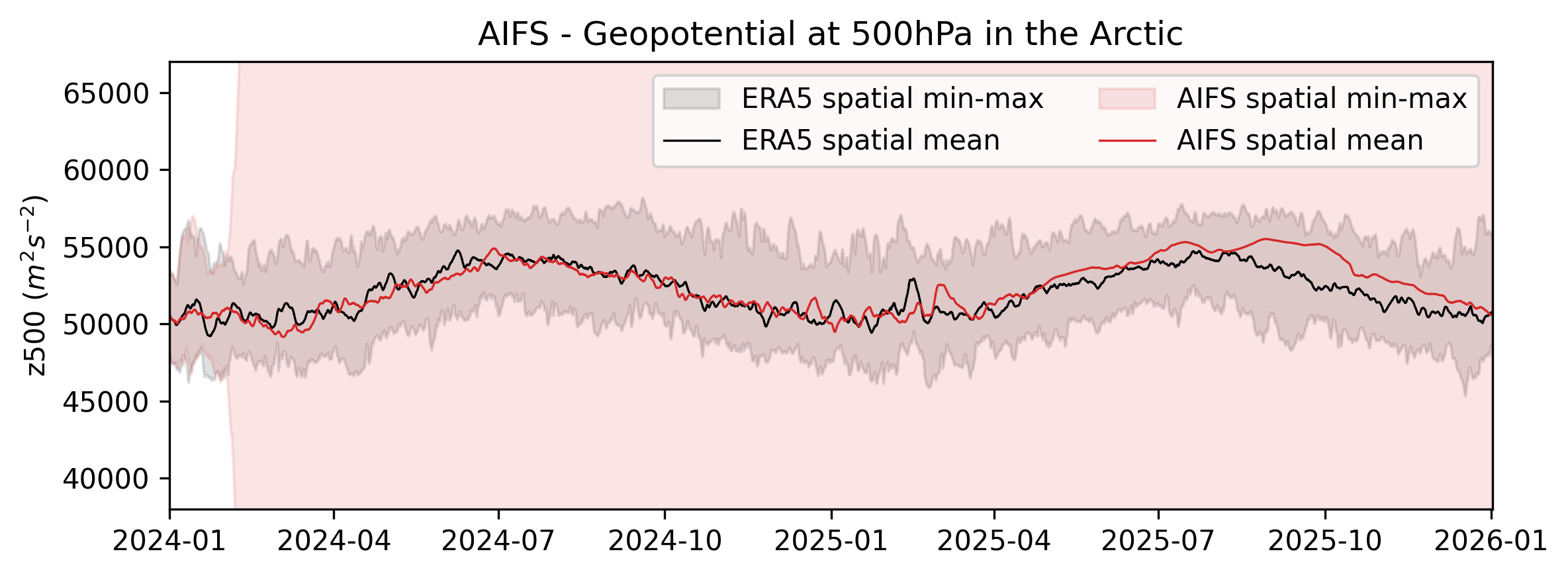}
        \includegraphics[width=0.75\linewidth]{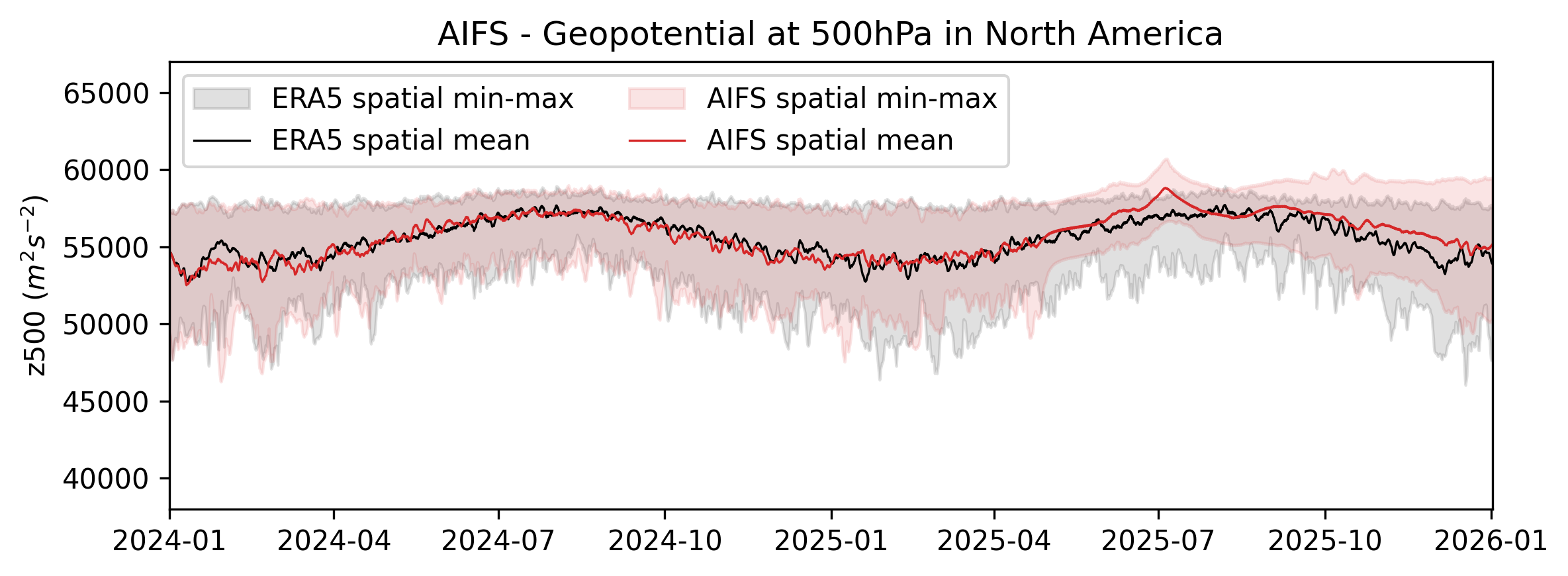}
        \caption{500~hPa geopotential rollouts with AIFS in Arctic (top) and North America (bottom). Rollouts are initialized on January 1, 2024. The solid lines show the spatial average across the globe for the reference ERA5 (black) and AIFS (red). The shaded areas extend from the minimum to the maximum geopotential over the Arctic and North America for ERA5 (grey) and AIFS (red).}
        \label{fig:blowup_aifs_z500}
    \end{figure}

    Figure~\ref{fig:blowup_aifs_z500} illustrates the early blow-up of AIFS for geopotential at 500~hPa, due to rapidly increasing values at the poles while values in other regions remain confined to physically realistic values.

\clearpage
\subsection{Loss of seasonality}
\label{sec:def_seasonality}
    Due to the large spatial extent of seasonal fluctuations, seasonality is quantified from the Mean Squared Error (MSE) decomposition (Equation~\ref{eq:decompose_mse}). The variance ratio $\overline{(\hat{u} - c)^2} /\overline{(u - c)^2}$ quantifies how much the model predictions deviate from climatology, relative to the typical deviation from the reference. 

    First, we determine the maximum variance ratio that should be acceptable, by shifting ERA5 time series by $\pm$~30~days, for 2-year time series starting on January 1 from 2000 to 2019. From these 40 shifted ERA5 time series, the maximum variance ratio observed for each variable is set as the threshold. Thresholds are higher for seasonal variables, e.g., 2-meter temperature (threshold of 5.5) and highly fluctuating variables, e.g., specific humidity at 100~hPa (threshold of 6.3), while they vary between 1.7 and 2.8 for the other variables. 

\begin{figure}[h]
        \centering
        \includegraphics[width=0.48\linewidth]{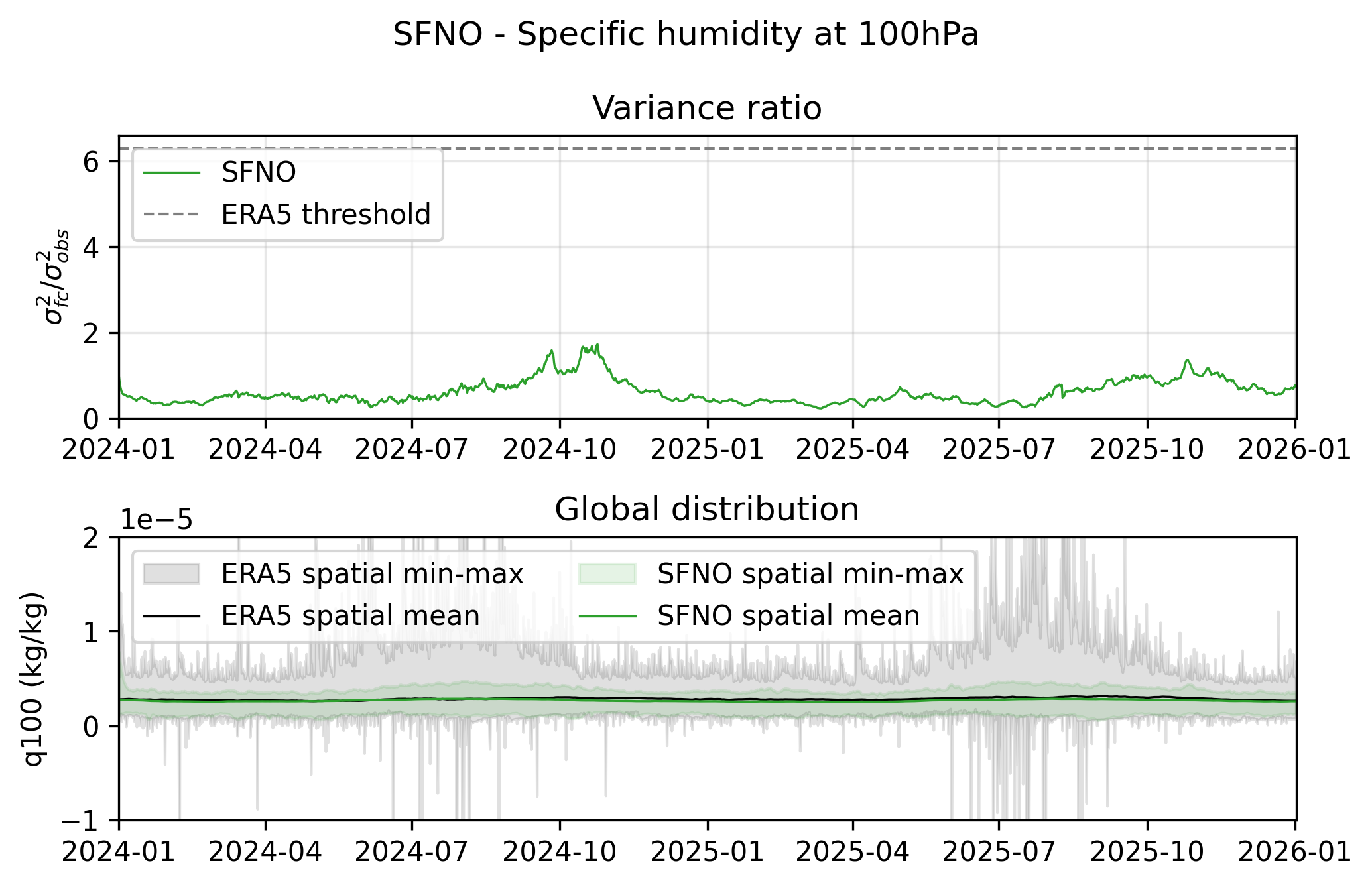}
        \includegraphics[width=0.5\linewidth]{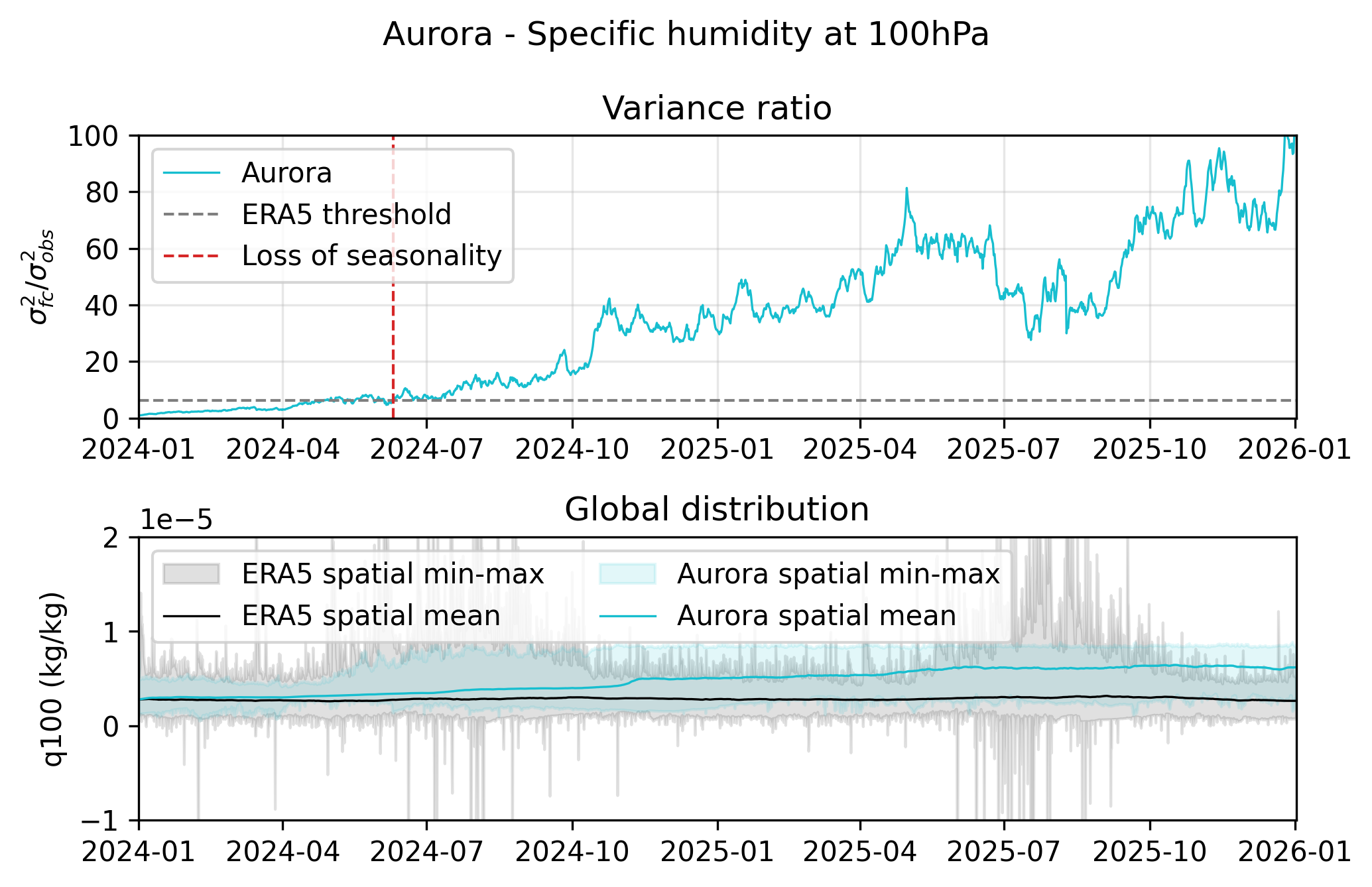}
        \caption{100~hPa specific humidity rollouts with SFNO (left) and Aurora (right). Rollouts are initialized on January 1, 2024. The upper panels show the variance ratio and the lower panels show the spatial average across the globe (solid lines) and minimum-maximum interval (shaded area) for the reference ERA5 (black) and the model (color).}
        \label{fig:rollout_q100}
    \end{figure}

    The loss of seasonality is detected as the first time when the variance ratio of the model predictions exceeds the ERA5 threshold and stays above the threshold for at least 15 days, as illustrated in Figure~\ref{fig:rollout_q100} for Aurora Q100.

\clearpage
\subsection{Small-scale artifacts}
\label{sec:def_artefacts}
    Small-scale artifacts are quantified from the energy spectrum averaged over wavelengths smaller than $\lambda_{max}$=400~km, $\mathcal{E}_{small}$. 
    The energy spectrum $\mathcal{E}$ at wavelength $\lambda$ is the amplitude of the Fourier coefficient at $\lambda$, taken along the longitude direction and averaged latitude-wise with latitude-dependent weights:
    \begin{equation}
        \hat{\mathcal{E}}_{small}(t) = \dfrac{1}{n_{small}} \sum_{\lambda \le \lambda_{max}} \hat{\mathcal{E}}(\lambda, t)  \text{  .}
    \end{equation}
    The threshold is computed as the maximum small-scale energy of the ERA5 observations between 1990 and 2019. When the predicted small-scale energy exceeds the maximum for more than 30 days, the rollout is detected as an exceedance of small-scale energy. If the small-scale energy was already higher than the threshold on the first rollout day, exceedance is not detected as it signals high energy in the model predictions, not specifically due to auto-regressive rollouts (reported as 0 in Tables~\ref{tab:small_scales}, \ref{tab:smallscale_sensitivity}, \ref{tab:smallscale_wavelength}).

    Table~\ref{tab:smallscale_wavelength} shows that the metric is moderately sensitive to the definition of the small-scale band: up to 250km, 400km, or 500km. Increasing the upper wavelength delays the time when the exceedance of small-scale energy is detected but does not change the conclusion of whether or not a model preserves small scales. This holds true except for geopotential at 500hPa where models tend to produce more refined predictions than the smooth ERA5 reference and the metric can significantly vary depending on the chosen wavelength. 

    \begin{table}[h]
    \centering
    \caption{Number of days before the small-scale energy exceeds the threshold for each model when varying the upper small-scale wavelength. All rollouts are initialized on January 1, 2024, except FuXi initialized on January 1, 2023. The final wavelength chosen in this work is 400km.}
    \label{tab:smallscale_wavelength}
    \setlength{\tabcolsep}{5pt} % default is typically 6pt
    \begin{small}
    \begin{tabular}{lr|ccccccccc}
    \toprule
    Model & $\lambda_{max}$ & T2m & U10m & MSLP & Z500 & T500 & Q500 & U300 & T100 & Q100 \\
    \midrule
    SFNO & 250km & $>$ 730 & $>$ 730 & $>$ 730 & 5 & $>$ 730 & $>$ 730 & $>$ 730 & $>$ 730 & $>$ 730 \\
    SFNO & 400km & $>$ 730 & $>$ 730 & $>$ 730 & 142 & $>$ 730 & $>$ 730 & $>$ 730 & $>$ 730 & $>$ 730 \\
    SFNO & 500km & $>$ 730 & $>$ 730 & $>$ 730 & $>$ 730 & $>$ 730 & $>$ 730 & $>$ 730 & $>$ 730 & $>$ 730 \\
    \midrule
    Aurora & 250km & $>$ 730 & $>$ 730 & $>$ 730 & 0 & $>$ 730 & $>$ 730 & $>$ 730 & $>$ 730 & $>$ 730 \\
    Aurora & 400km & $>$ 730 & $>$ 730 & $>$ 730 & $>$ 730 & $>$ 730 & $>$ 730 & $>$ 730 & $>$ 730 & $>$ 730 \\
    Aurora & 500km & $>$ 730 & $>$ 730 & $>$ 730 & $>$ 730 & $>$ 730 & $>$ 730 & $>$ 730 & $>$ 730 & $>$ 730 \\
    \midrule
    FourCastNet & 250km & 11 & 15 & 11 & 0 & 6 & - & - & - & - \\
    FourCastNet & 400km & 11 & 15 & 12 & 0 & 8 & - & - & - & - \\
    FourCastNet & 500km & 11 & 15 & 12 & 2 & 8 & - & - & - & - \\
    \midrule
    FuXi & 250km & 5 & 17 & 4 & 1 & 3 & - & 4 & 4 & - \\
    FuXi & 400km & 12 & 27 & 11 & 5 & 9 & - & 13 & 14 & - \\
    FuXi & 500km & 15 & 32 & 13 & 7 & 14 & - & 16 & 17 & - \\
    \midrule
    Pangu & 250km & $>$ 730 & $>$ 730 & $>$ 730 & $>$ 730 & $>$ 730 & $>$ 730 & $>$ 730 & $>$ 730 & $>$ 730 \\
    Pangu & 400km & $>$ 730 & $>$ 730 & $>$ 730 & $>$ 730 & $>$ 730 & $>$ 730 & $>$ 730 & $>$ 730 & $>$ 730 \\
    Pangu & 500km & $>$ 730 & $>$ 730 & $>$ 730 & $>$ 730 & $>$ 730 & $>$ 730 & $>$ 730 & $>$ 730 & $>$ 730 \\
    \midrule
    AIFS & 250km & 212 & $>$ 730 & 204 & 5 & 438 & $>$ 730 & $>$ 730 & 447 & 547 \\
    AIFS & 400km & 226 & $>$ 730 & 257 & 15 & 469 & $>$ 730 & $>$ 730 & 536 & $>$ 730 \\
    AIFS & 500km & 237 & $>$ 730 & 321 & 26 & 483 & $>$ 730 & $>$ 730 & 654 & $>$ 730 \\
    \midrule
    GraphCast & 250km & 7 & 545 & 9 & 0 & 43 & 425 & 364 & 118 & 416 \\
    GraphCast & 400km & 37 & 636 & 22 & 10 & 75 & 456 & 392 & 171 & 438 \\
    GraphCast & 500km & 43 & 669 & 32 & 23 & 90 & 474 & 400 & 224 & 451 \\
    \midrule
    DLESyM & 250km & $>$ 730 & - & $>$ 730 & $>$ 730 & - & - & - & - & - \\
    DLESyM & 400km & $>$ 730 & - & $>$ 730 & 14 & - & - & - & - & - \\
    DLESyM & 500km & $>$ 730 & - & $>$ 730 & 15 & - & - & - & - & - \\
    \bottomrule
    \end{tabular}
    \end{small}
    \end{table}

\begin{figure}[h]
        \centering
        \includegraphics[width=0.48\linewidth]{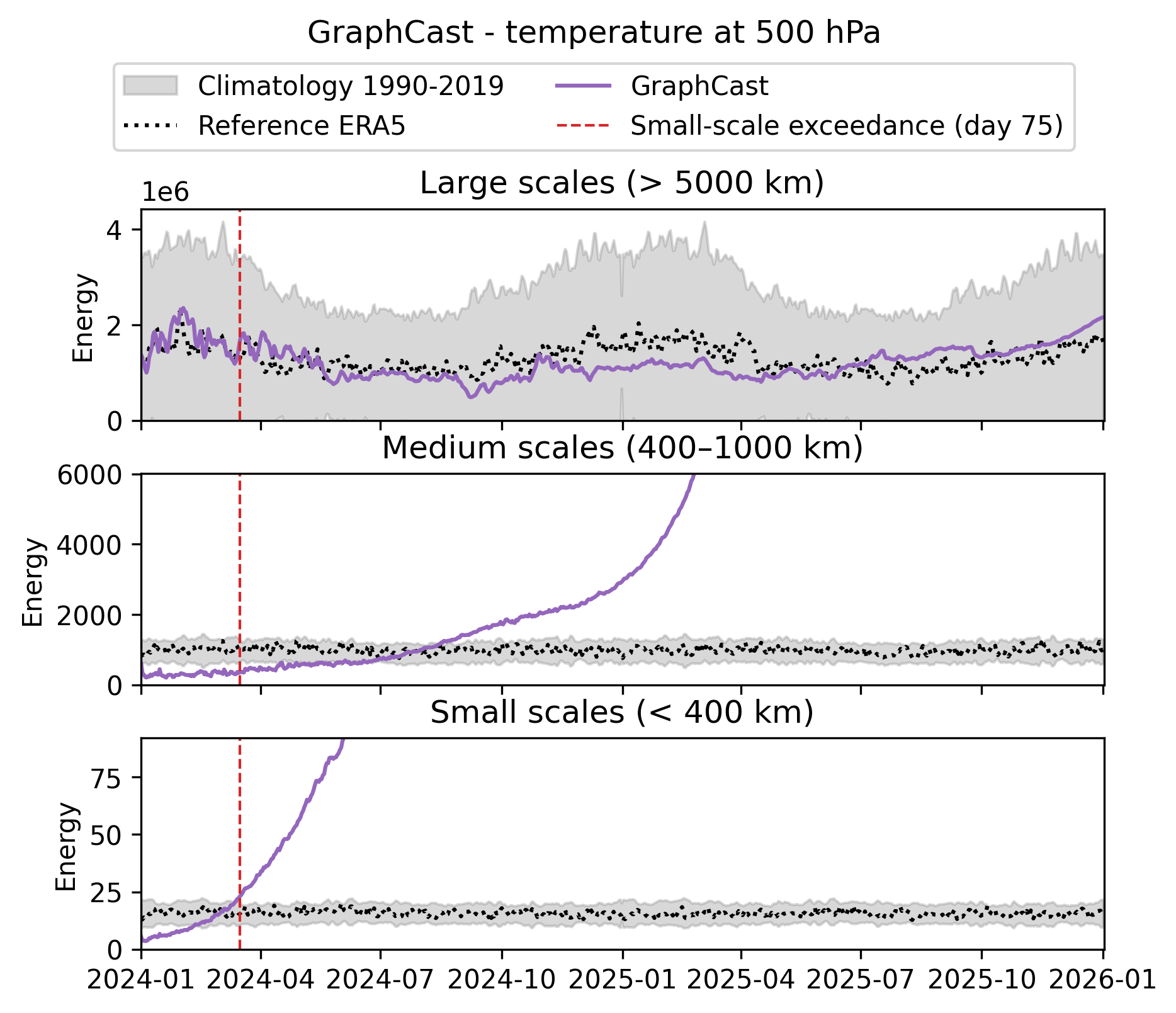}
        \includegraphics[width=0.48\linewidth]{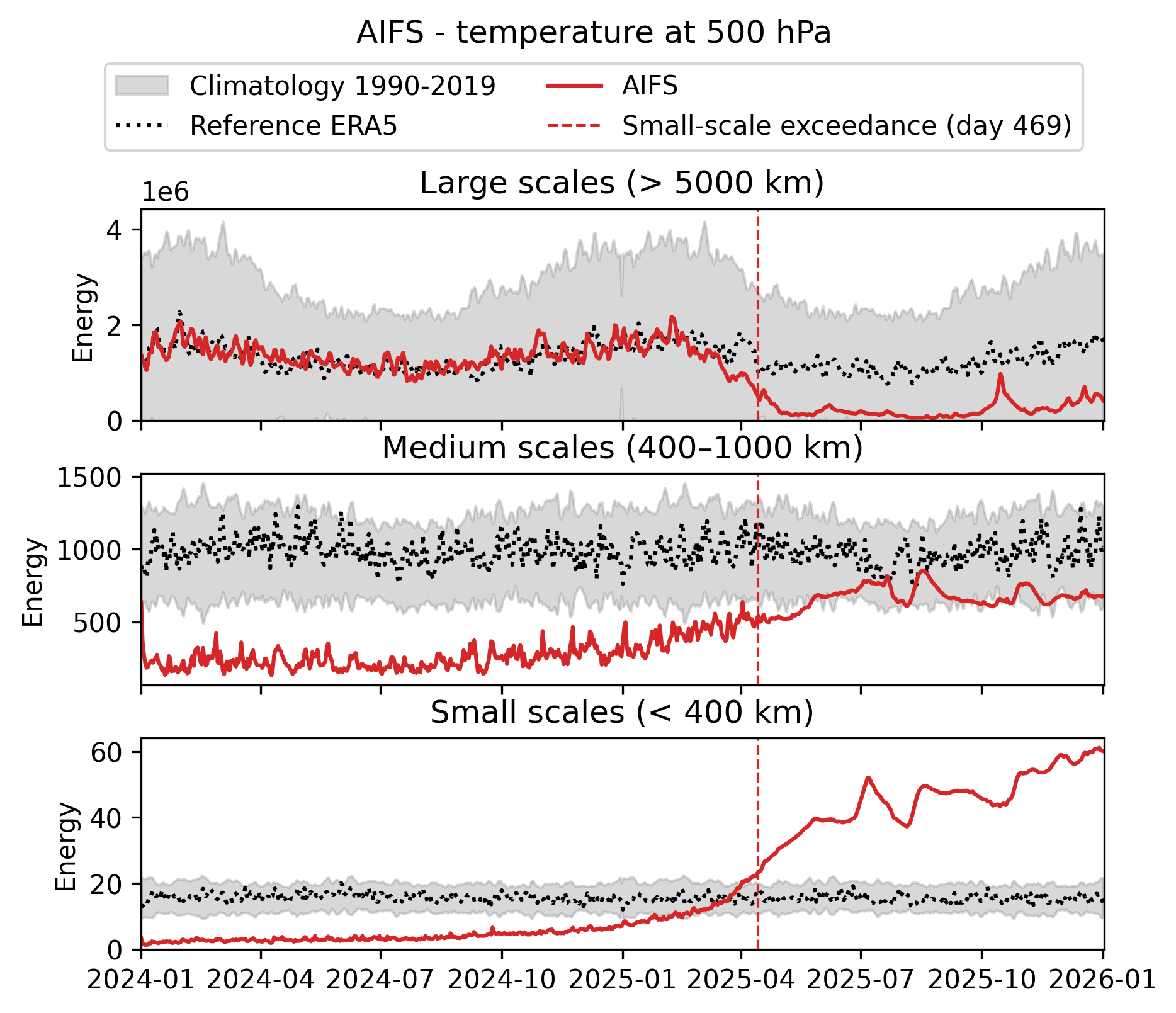}
        \caption{Energy spectra of 500~hPa temperature rollouts for GraphCast (left) and AIFS (right). The spectra are averaged into three bands: large wavelengths (larger than 5000~km), medium wavelengths (between 400 and 1000~km), and small wavelengths (smaller than 400~km). Energy spectrum from ERA5 is shown in black. For a given day of the year, the grey area extends from the minimum to the maximum observed energy spectrum on this day between 1990 and 2019.}
        \label{fig:spectra_graphcast_aifs}
    \end{figure}

\begin{figure}[h]
        \centering
        \includegraphics[width=0.48\linewidth]{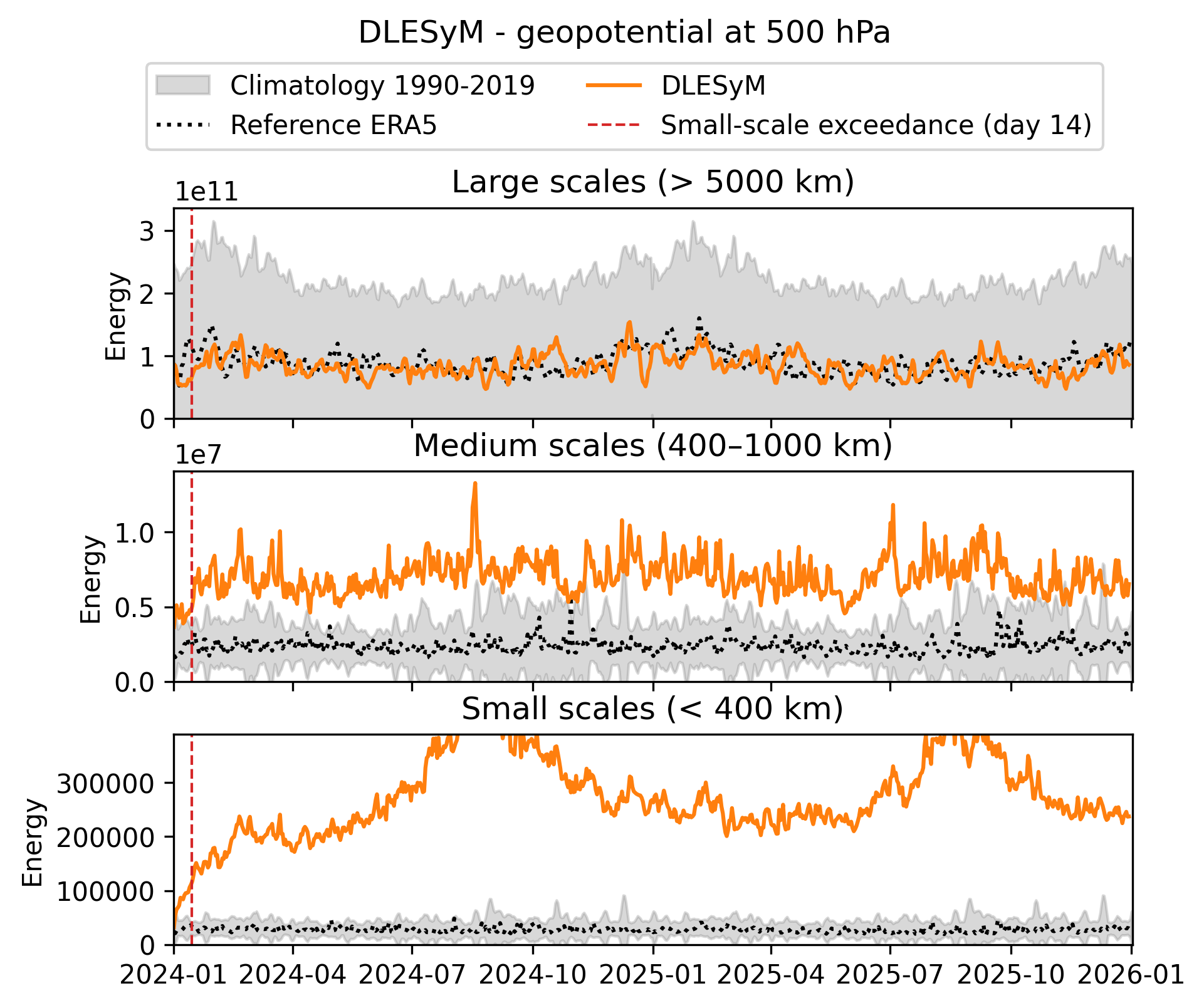}
        \includegraphics[width=0.48\linewidth]{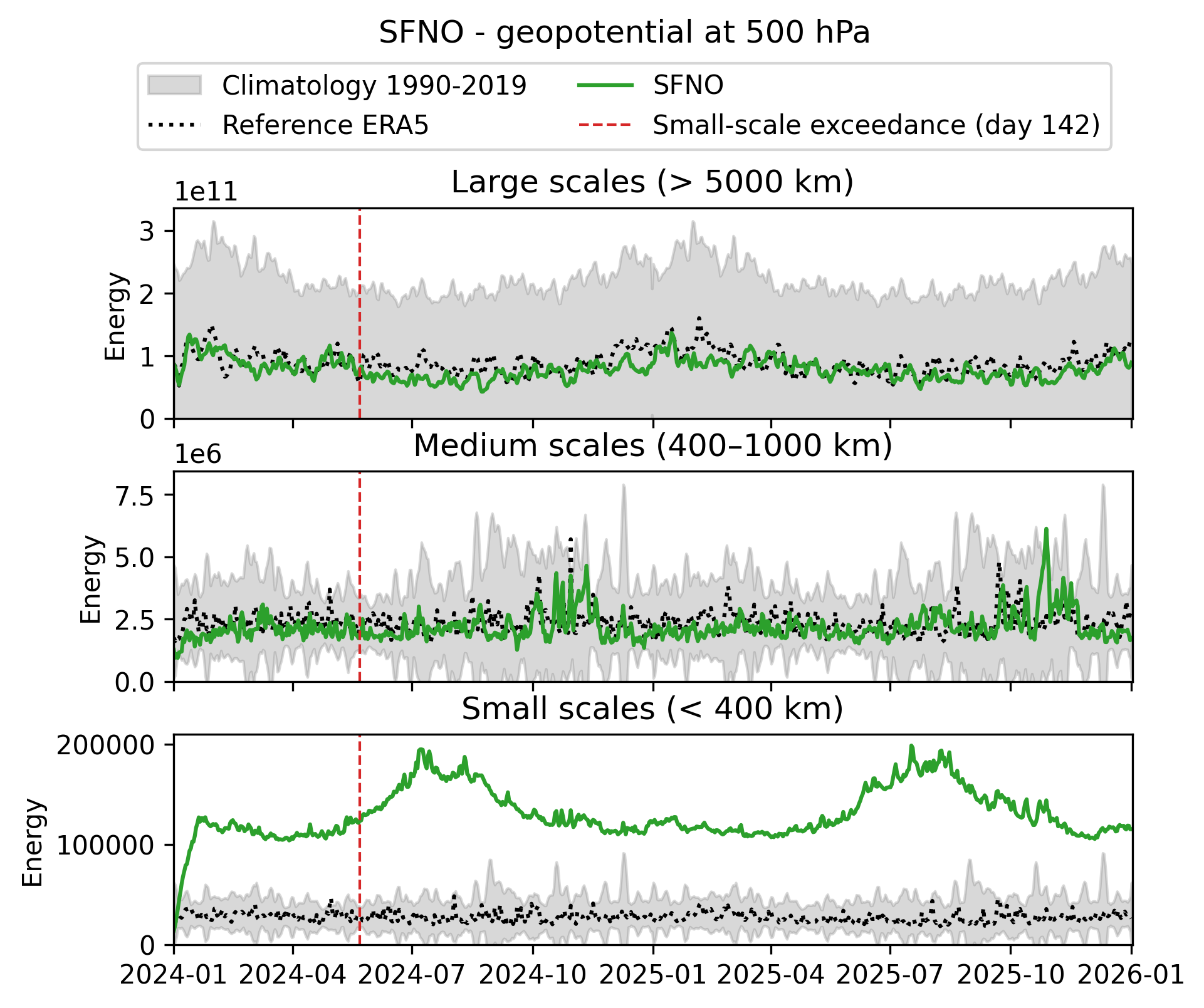}
        \caption{Energy spectra of 500~hPa geopotential rollouts for DLESyM (left) and SFNO (right).}
        \label{fig:spectra_sfno_dlesym}
    \end{figure}

    For unstable models, exceedance of small-scale energy tends to be a precursor of larger scale degradation (AIFS, GraphCast, Figure~\ref{fig:spectra_graphcast_aifs}) while stable models confine the energy increase to the small scales without an overall impact on the rollout (SFNO, DLESyM, Figure~\ref{fig:spectra_sfno_dlesym}). 

    \begin{figure}[h]
        \centering
        \includegraphics[width=0.6\linewidth]{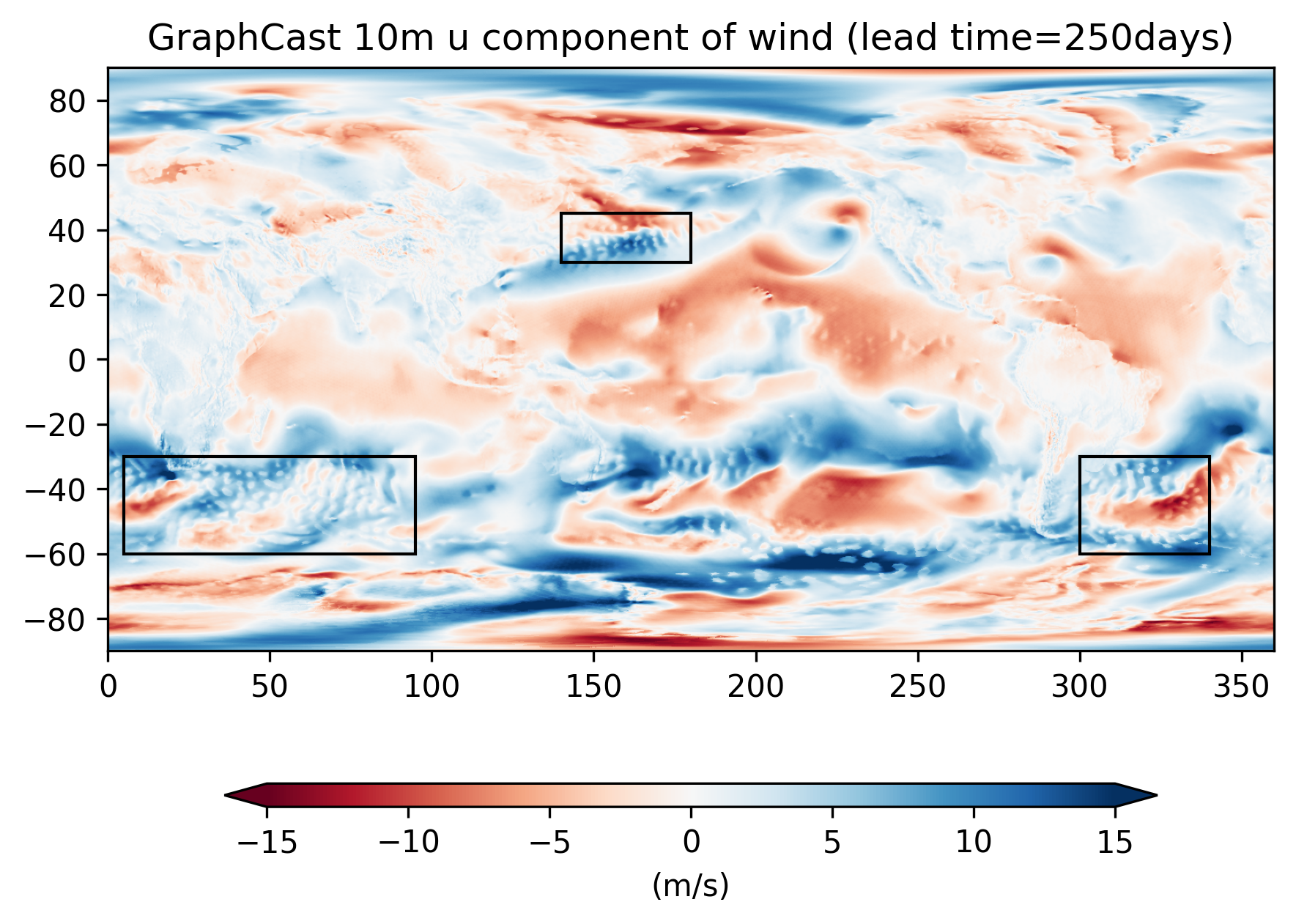}
        \caption{10-meter eastward component of wind predicted by GraphCast after 250 rollout days. Boxes highlight regions where small-scale artifacts are especially visible.}
        \label{fig:map_artefacts_graphcast}
    \end{figure}

\clearpage 
\subsection{Summary and sensitivity to initial conditions}
\label{appendix:sensitivity_metrics}
\begin{figure}[h]
        \centering
        \includegraphics[width=\linewidth]{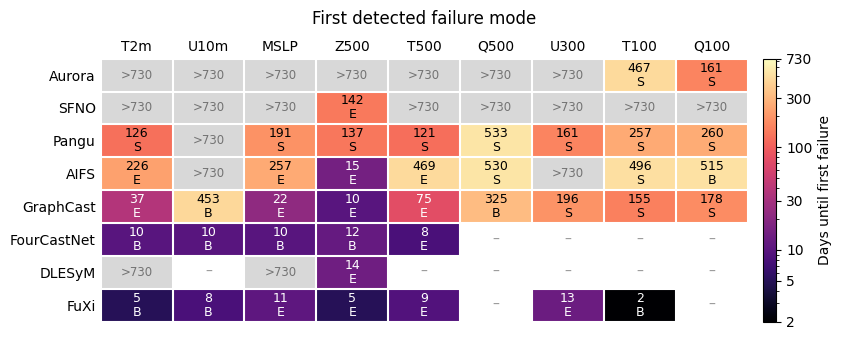}
        \caption{First day when one of the failure modes is detected (B=blow-up, S=loss of seasonality, E=exceedance of small-scale energy).}
        \label{fig:summary_failures}
    \end{figure}

    Figure \ref{fig:summary_failures} summarizes the findings of Tables~\ref{tab:blowup}, \ref{tab:seasonality}, \ref{tab:small_scales} and indicates the first day when one of the failure modes is detected.  

    This section reports the average and standard deviation of the blow-up, loss of seasonality, and small-scale artifacts metrics, computed over rollouts with ten different initial conditions (2022-01-01, 2022-04-01, 2022-07-01, 2022-10-01, 2023-01-01, 2023-04-01, 2023-07-01, 2023-10-01, 2024-01-01, 2024-04-01), limited to dates before 2023 for FuXi. 

    \begin{table*}[h]
      \caption{Number of days until the model predictions blow-up. Same as Table~\ref{tab:blowup} but with mean and standard deviation across ten rollouts initialized on different dates.}
      \label{tab:blowup_sensitivity}
      \centering
        \begin{small}
            \begin{tabular}{lccccccccr}
              \toprule
              Model & T2m & U10m & MSLP & Z500 & T500 & Q500 & U300 & T100 & Q100 \\
              \midrule
                Aurora & $>$ 730 & $>$ 730 & $>$ 730 & $>$ 730 & $>$ 730 & $>$ 730 & $>$ 730 & $>$ 730 & $>$ 730 \\
                SFNO & $>$ 730 & $>$ 730 & $>$ 730 & \makecell{706 \\ $\pm$ 76} & $>$ 730 & $>$ 730 & $>$ 730 & \makecell{546 \\ $\pm$ 233} & $>$ 730 \\
                Pangu & $>$ 730 & $>$ 730 & $>$ 730 & $>$ 730 & $>$ 730 & $>$ 730 & $>$ 730 & $>$ 730 & $>$ 730 \\
                AIFS & \makecell{338 \\ $\pm$ 121} & $>$ 730 & $>$ 730 & \makecell{25 \\ $\pm$ 10} & $>$ 730 & $>$ 730 & $>$ 730 & \makecell{602 \\ $\pm$ 148} & \makecell{474 \\ $\pm$ 135} \\
                GraphCast & \makecell{305 \\ $\pm$ 113} & \makecell{347 \\ $\pm$ 85} & \makecell{284 \\ $\pm$ 72} & \makecell{253 \\ $\pm$ 77} & \makecell{276 \\ $\pm$ 70} & \makecell{282 \\ $\pm$ 80} & \makecell{300 \\ $\pm$ 98} & \makecell{233 \\ $\pm$ 129} & \makecell{281 \\ $\pm$ 87} \\
                FourCastNet & \makecell{6 \\ $\pm$ 3} & \makecell{7 \\ $\pm$ 4} & \makecell{8 \\ $\pm$ 3} & \makecell{9 \\ $\pm$ 4} & \makecell{8 \\ $\pm$ 3} & - & - & - & - \\
                DLESyM & $>$ 730 & - & $>$ 730 & $>$ 730 & - & - & - & - & - \\
                FuXi & \makecell{3 \\ $\pm$ 1} & \makecell{10 \\ $\pm$ 1} & \makecell{580 \\ $\pm$ 212} & \makecell{121 \\ $\pm$ 28} & \makecell{23 \\ $\pm$ 15} & - & \makecell{13 \\ $\pm$ 2} & \makecell{25 \\ $\pm$ 18} & - \\
              \bottomrule
            \end{tabular}
        \end{small}
    \end{table*}

    \begin{table*}[h]
      \caption{Number of days until the model predictions lose seasonality. Same as Table~\ref{tab:seasonality} but with mean and standard deviation across ten rollouts initialized on different dates.}
      \label{tab:seasonality_sensitivity}
      \centering
        \begin{small}
            \begin{tabular}{lccccccccr}
              \toprule
              Model & T2m & U10m & MSLP & Z500 & T500 & Q500 & U300 & T100 & Q100 \\
              \midrule
                Aurora & $>$ 730 & $>$ 730 & $>$ 730 & $>$ 730 & $>$ 730 & $>$ 730 & $>$ 730 & \makecell{405 \\ $\pm$ 92} & \makecell{86 \\ $\pm$ 39} \\
                SFNO & $>$ 730 & $>$ 730 & $>$ 730 & $>$ 730 & $>$ 730 & $>$ 730 & $>$ 730 & \makecell{699 \\ $\pm$ 99} & $>$ 730 \\
                Pangu & \makecell{147 \\ $\pm$ 34} & $>$ 730 & \makecell{403 \\ $\pm$ 198} & \makecell{185 \\ $\pm$ 57} & \makecell{125 \\ $\pm$ 38} & \makecell{374 \\ $\pm$ 177} & \makecell{257 \\ $\pm$ 95} & \makecell{201 \\ $\pm$ 53} & \makecell{212 \\ $\pm$ 73} \\
                AIFS & \makecell{520 \\ $\pm$ 60} & $>$ 730 & $>$ 730 & \makecell{605 \\ $\pm$ 108} & \makecell{541 \\ $\pm$ 113} & \makecell{492 \\ $\pm$ 105} & $>$ 730 & \makecell{473 \\ $\pm$ 85} & \makecell{461 \\ $\pm$ 117} \\
                GraphCast & \makecell{152 \\ $\pm$ 31} & \makecell{619 \\ $\pm$ 186} & \makecell{322 \\ $\pm$ 162} & \makecell{175 \\ $\pm$ 39} & \makecell{133 \\ $\pm$ 31} & \makecell{407 \\ $\pm$ 195} & \makecell{184 \\ $\pm$ 44} & \makecell{156 \\ $\pm$ 32} & \makecell{233 \\ $\pm$ 190} \\
                FourCastNet & \makecell{12 \\ $\pm$ 3} & \makecell{14 \\ $\pm$ 3} & \makecell{16 \\ $\pm$ 3} & \makecell{15 \\ $\pm$ 3} & \makecell{13 \\ $\pm$ 3} & - & - & - & - \\
                DLESyM & $>$ 730 & - & $>$ 730 & $>$ 730 & - & - & - & - & - \\
                FuXi & \makecell{101 \\ $\pm$ 11} & \makecell{184 \\ $\pm$ 11} & \makecell{143 \\ $\pm$ 26} & \makecell{142 \\ $\pm$ 19} & \makecell{105 \\ $\pm$ 22} & - & \makecell{144 \\ $\pm$ 12} & \makecell{69 \\ $\pm$ 7} & - \\
              \bottomrule
            \end{tabular}
        \end{small}
    \end{table*}

    \begin{table*}[h]
      \caption{Number of days until the small-scale energy exceeds the ERA5 range. Same as Table~\ref{tab:small_scales} but with mean and standard deviation across ten rollouts initialized on different dates.}
      \label{tab:smallscale_sensitivity}
      \centering
        \begin{small}
            \begin{tabular}{lccccccccr}
              \toprule
              Model & T2m & U10m & MSLP & Z500 & T500 & Q500 & U300 & T100 & Q100 \\
              \midrule
                Aurora & $>$ 730 & $>$ 730 & $>$ 730 & $>$ 730 & $>$ 730 & $>$ 730 & $>$ 730 & $>$ 730 & $>$ 730 \\
                SFNO & $>$ 730 & $>$ 730 & $>$ 730 & \makecell{106 \\ $\pm$ 85} & $>$ 730 & $>$ 730 & $>$ 730 & $>$ 730 & $>$ 730 \\
                Pangu & $>$ 730 & $>$ 730 & $>$ 730 & $>$ 730 & $>$ 730 & $>$ 730 & $>$ 730 & $>$ 730 & $>$ 730 \\
                AIFS & \makecell{252 \\ $\pm$ 17} & $>$ 730 & \makecell{287 \\ $\pm$ 21} & \makecell{16 \\ $\pm$ 1} & \makecell{447 \\ $\pm$ 30} & $>$ 730 & $>$ 730 & \makecell{522 \\ $\pm$ 40} & $>$ 730 \\
                GraphCast & \makecell{17 \\ $\pm$ 8} & \makecell{563 \\ $\pm$ 175} & \makecell{26 \\ $\pm$ 7} & \makecell{14 \\ $\pm$ 2} & \makecell{80 \\ $\pm$ 3} & \makecell{412 \\ $\pm$ 119} & \makecell{369 \\ $\pm$ 129} & \makecell{189 \\ $\pm$ 43} & \makecell{376 \\ $\pm$ 106} \\
                FourCastNet & \makecell{10 \\ $\pm$ 2} & \makecell{12 \\ $\pm$ 3} & \makecell{10 \\ $\pm$ 2} & \makecell{0 \\ $\pm$ 0} & \makecell{8 \\ $\pm$ 1} & - & - & - & - \\
                DLESyM & $>$ 730 & - & $>$ 730 & \makecell{16 \\ $\pm$ 3} & - & - & - & - & - \\
                FuXi & \makecell{12 \\ $\pm$ 1} & \makecell{28 \\ $\pm$ 2} & \makecell{10 \\ $\pm$ 2} & \makecell{5 \\ $\pm$ 0} & \makecell{9 \\ $\pm$ 1} & - & \makecell{13 \\ $\pm$ 1} & \makecell{14 \\ $\pm$ 0} & - \\
              \bottomrule
            \end{tabular}
        \end{small}
    \end{table*}

\clearpage
\section{Stability results for AIFS Ensemble}
    Since AIFS is part of our main analysis, we extend the results to the probabilistic version, AIFS-CRPS \cite{lang_aifscrps_2024}. AIFS-CRPS shows earlier instability than the deterministic AIFS for all three failure modes (Table~\ref{tab:instability_aifsens}).

    \begin{figure}[h]
        \centering
        \includegraphics[width=\linewidth]{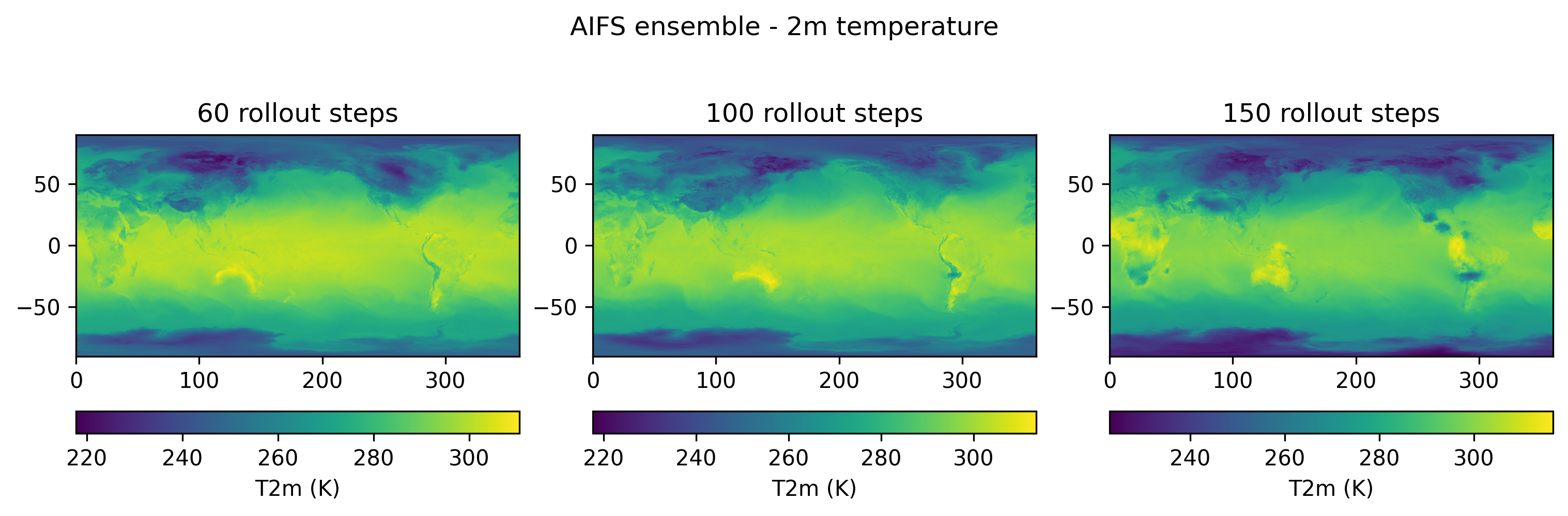}
        \caption{Predictions of 2-meter temperature after 60, 100, and 150 rollout steps with AIFS ensemble \cite{lang_aifscrps_2024}. Rollout initialized on January 1, 2024.}
        \label{fig:aifs_ens_2t}
    \end{figure}

    \begin{table*}[h]
      \caption{Number of days until AIFS-CRPS predictions blow-up (first row), lose seasonality (second row), exceed small-scale energy (third row). Rollouts are initialized on January 1, 2024 with 5 different seeds. Instability is detected for each ensemble member and the table reports the mean $\pm$ standard deviation across the five seeds.}
      \label{tab:instability_aifsens}
      \begin{center}
        \begin{small}
        \setlength{\tabcolsep}{5pt} % default is typically 6pt
            \begin{tabular}{lccccccccr}
              \toprule
              Metric & T2m & U10m & MSLP & Z500 & T500 & Q500 & U300 & T100 & Q100 \\
              \midrule
              Blow-up & 26 $\pm$ 1 & 45 $\pm$ 17 & 2 $\pm$ 0 & 6 $\pm$ 1 & 38 $\pm$ 3 & 156 $\pm$ 71 & $>$ 730 & 16 $\pm$ 1 & 477 $\pm$ 5 \\
              Seasonality & 48 $\pm$ 0 & 60 $\pm$ 0 & 35 $\pm$ 1 & 38 $\pm$ 1 & 41 $\pm$ 1 & 90 $\pm$ 5 & 158 $\pm$ 1 & 53 $\pm$ 0 & 48 $\pm$ 5 \\
              Small scales & $>$ 730 & $>$ 730 & 29 $\pm$ 0 & 23 $\pm$ 0 & 0 $\pm$ 0 & $>$ 730 & 0 $\pm$ 0 & 0 $\pm$ 0 & $>$ 730 \\
              \bottomrule
            \end{tabular}
        \end{small}
      \end{center}
      \vskip -0.1in
    \end{table*}

\clearpage
\section{Noise injection in the rollout}
\subsection{Adding Gaussian noise to the inputs}
\begin{figure}[h]
        \centering
        \includegraphics[width=0.95\linewidth]{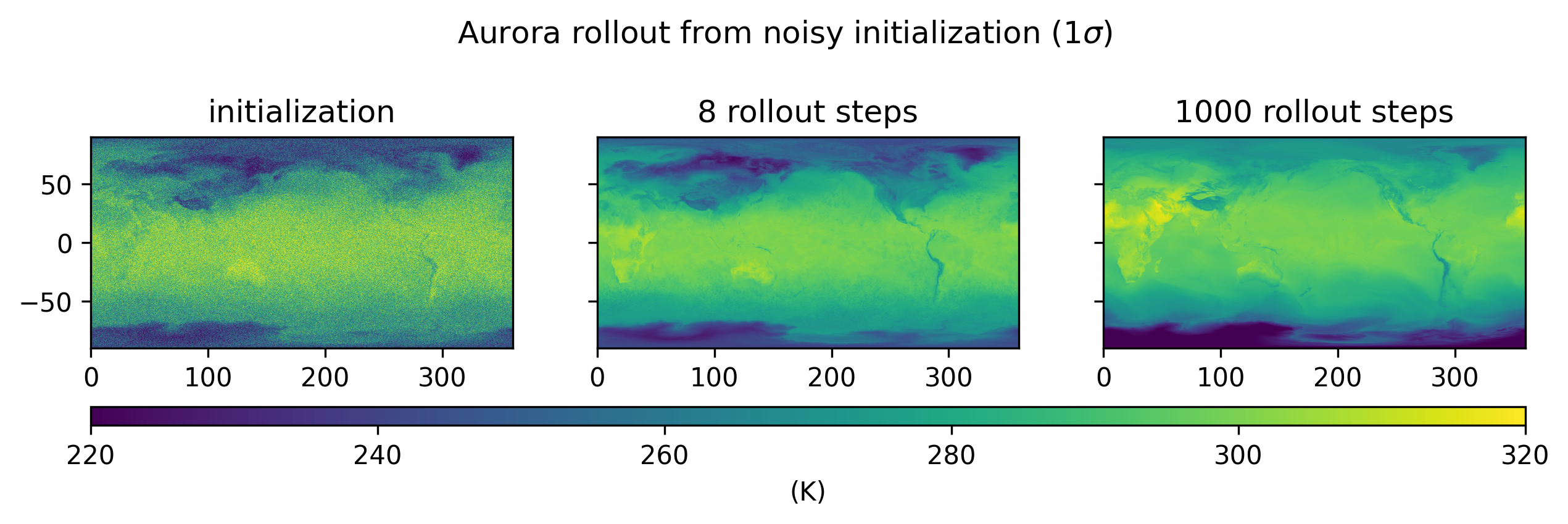}
        \includegraphics[width=0.95\linewidth]{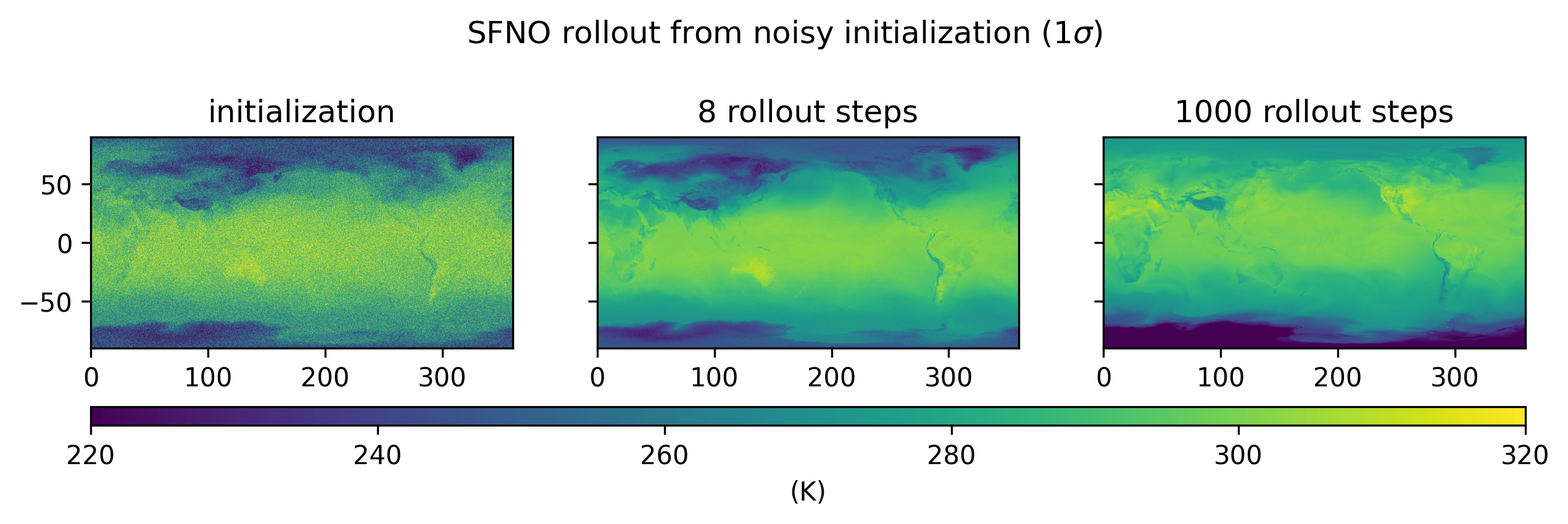}
        \includegraphics[width=0.95\linewidth]{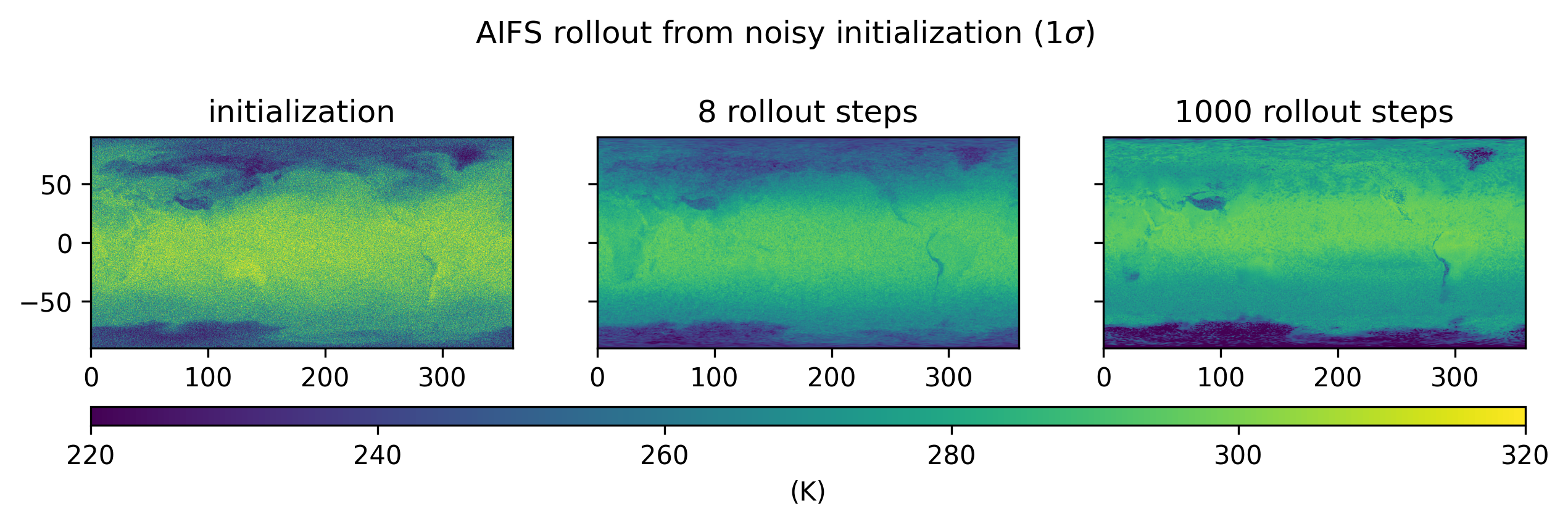}
        \includegraphics[width=0.95\linewidth]{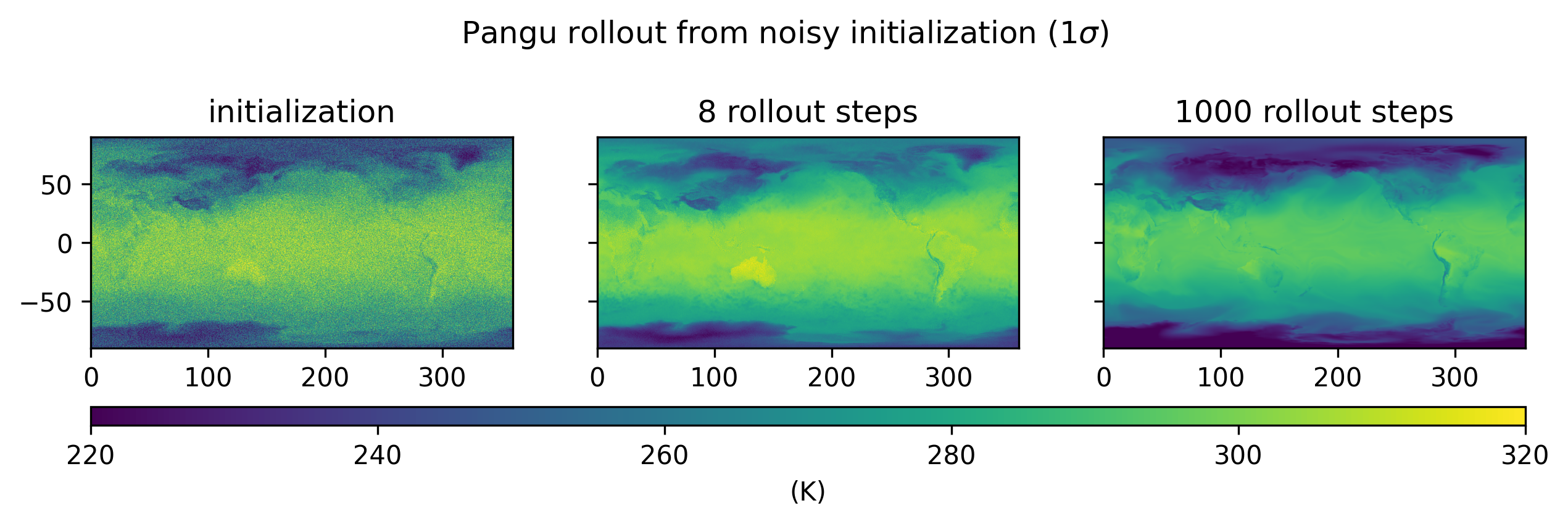}
        \includegraphics[width=0.95\linewidth]{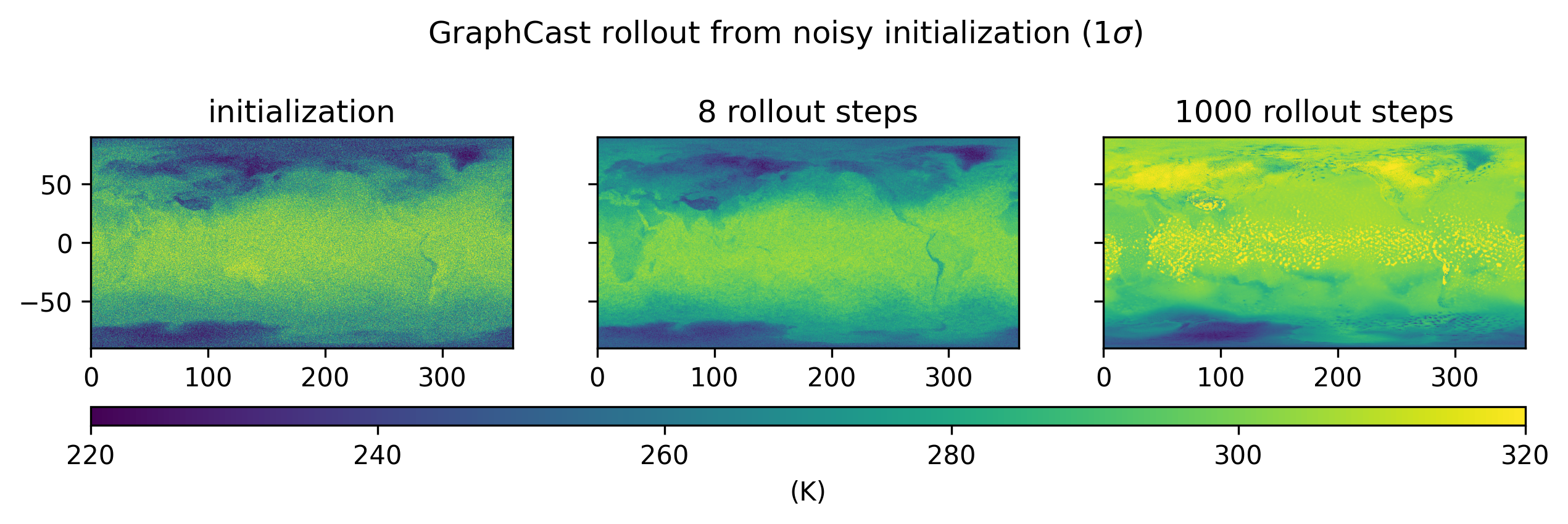}
        \caption{2-meter temperature at initialization (left), predicted after 8 (middle) and 1000 (right) rollout steps when inputs are perturbed with additive Gaussian noise of standard deviation $\sigma$.}
        \label{fig:rollout_noise_map_2t}
    \end{figure}

    First, Gaussian noise is added to the inputs (surface and atmospheric variables), with a standard deviation equal to the standard deviation of the physical variable (Fig.~\ref{fig:rollout_noise_map_2t}). This tests the model sensitivity to small-scale random perturbations.  
    
    \begin{figure*}[h]
        \centering
        \includegraphics[width=0.85\textwidth]{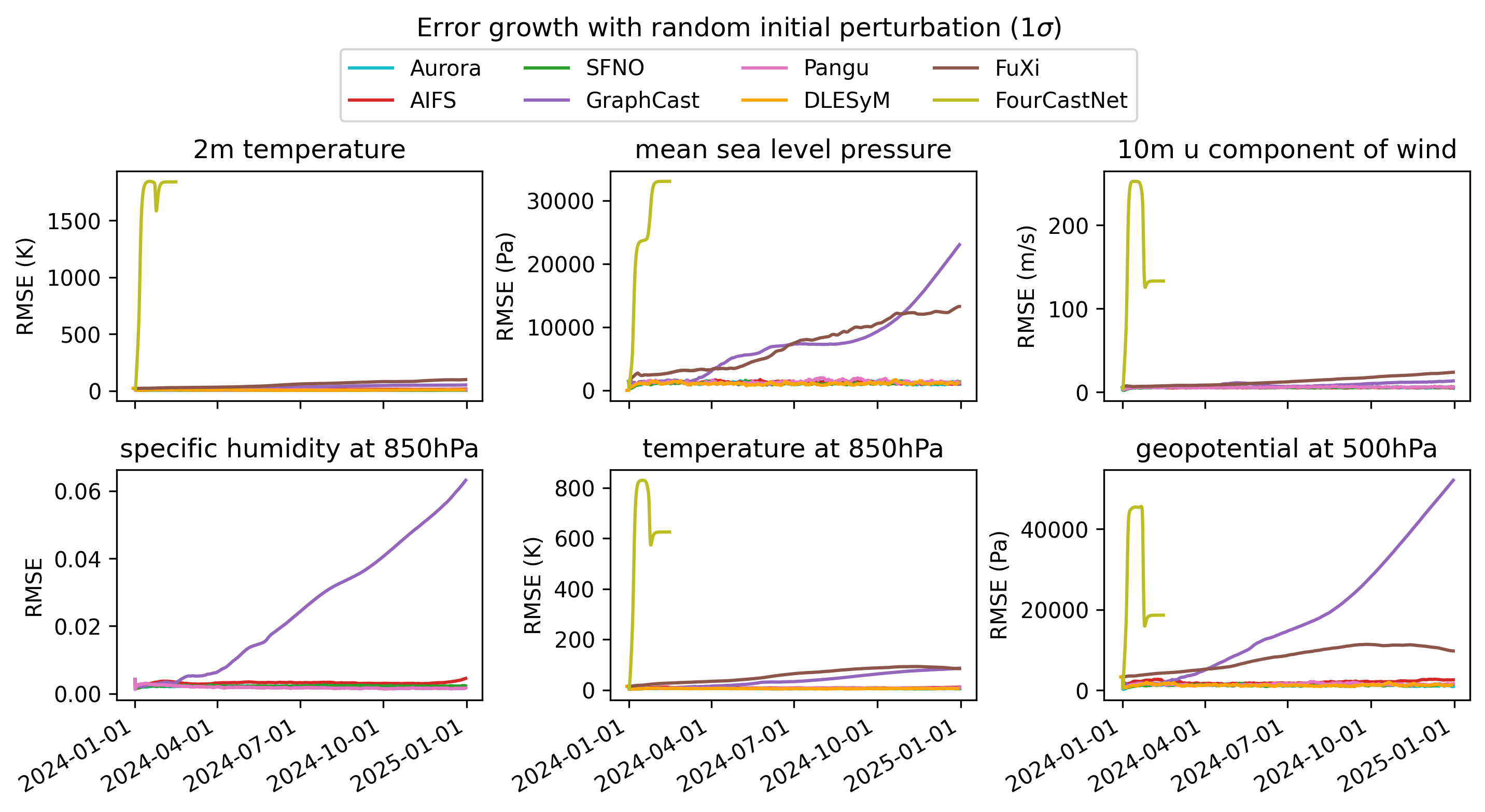}
        \caption{Same as Fig.~\ref{fig:rollout_noise} but showing all models. Error between the reference rollout and the rollout initialized with a noisy input (additive noise with $1\sigma$ standard deviation) for all models and six variables.}
        \label{fig:rollout_noise_all}
    \end{figure*}

    \begin{figure}[h]
        \centering
        \includegraphics[width=\textwidth]{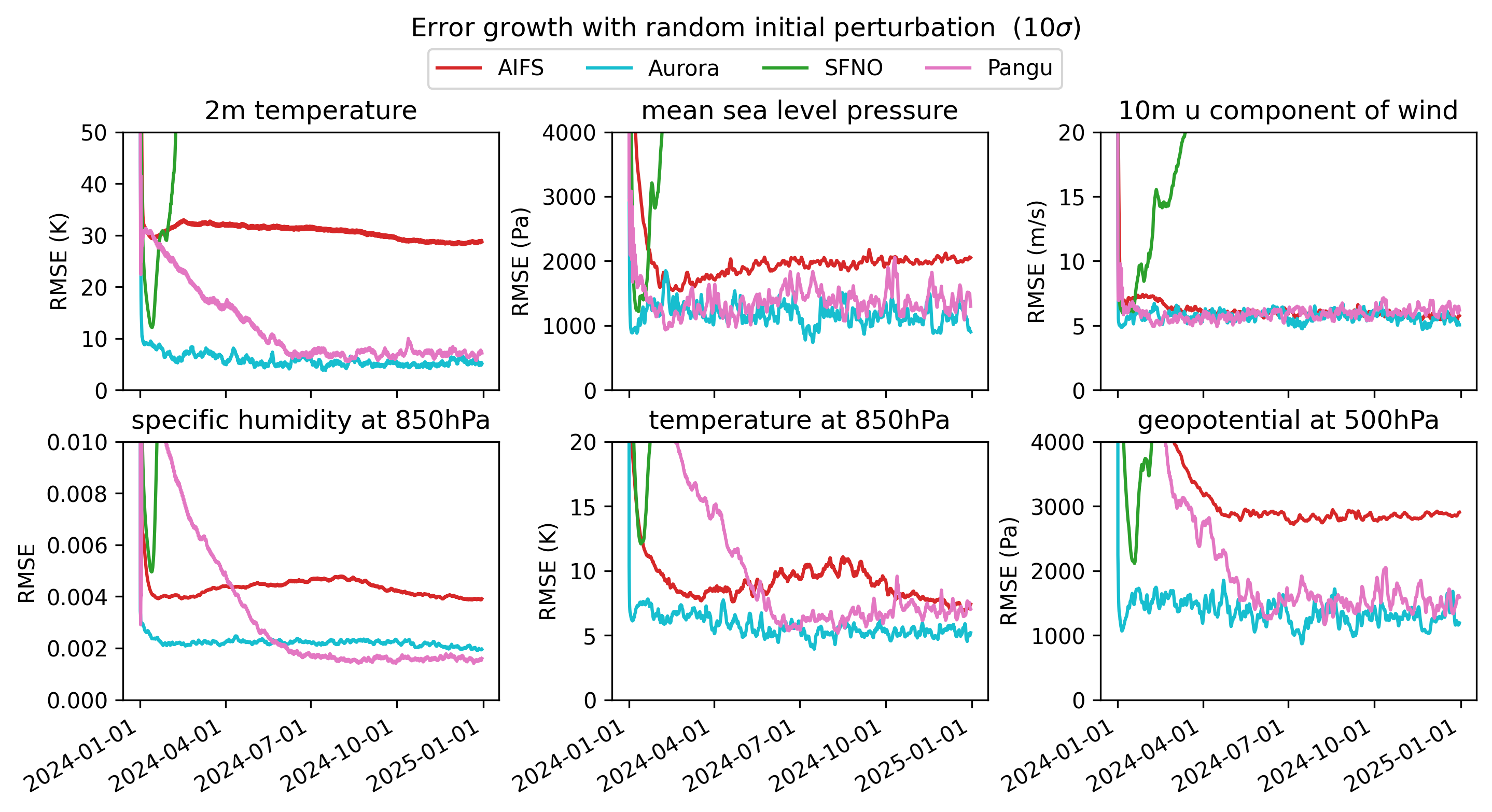}
        \caption{Error between the reference rollout and the rollout initialized with a noisy input (additive noise with $10\sigma$ standard deviation) for four models: Aurora (cyan), AIFS (red), SFNO (green), Pangu (pink) and six variables.}
        \label{fig:rollout_noise10}
    \end{figure}

\begin{figure}[h]
        \centering
        \includegraphics[width=1\linewidth]{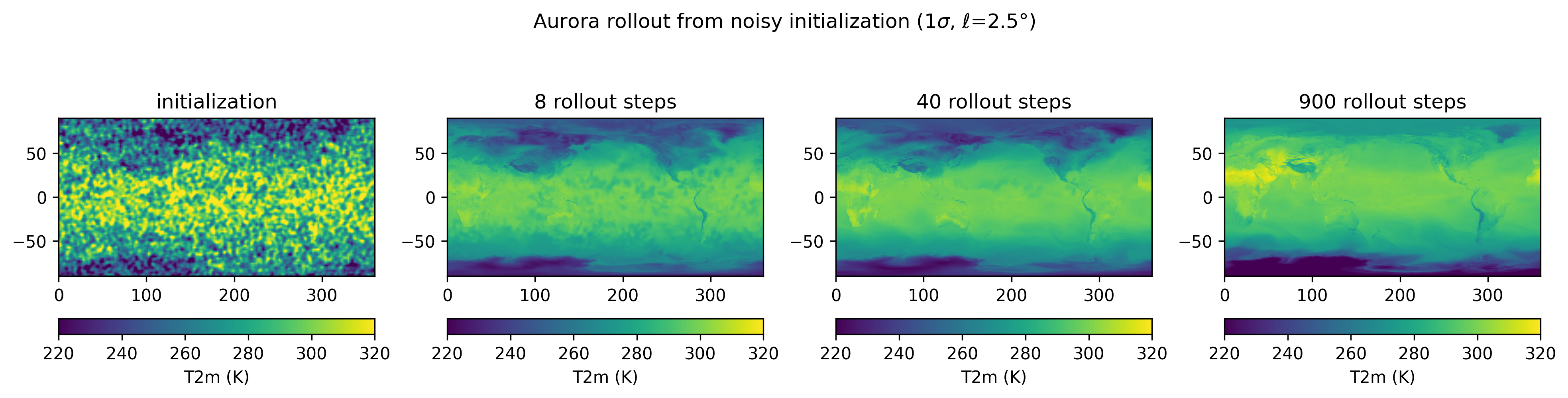}
        \includegraphics[width=\linewidth]{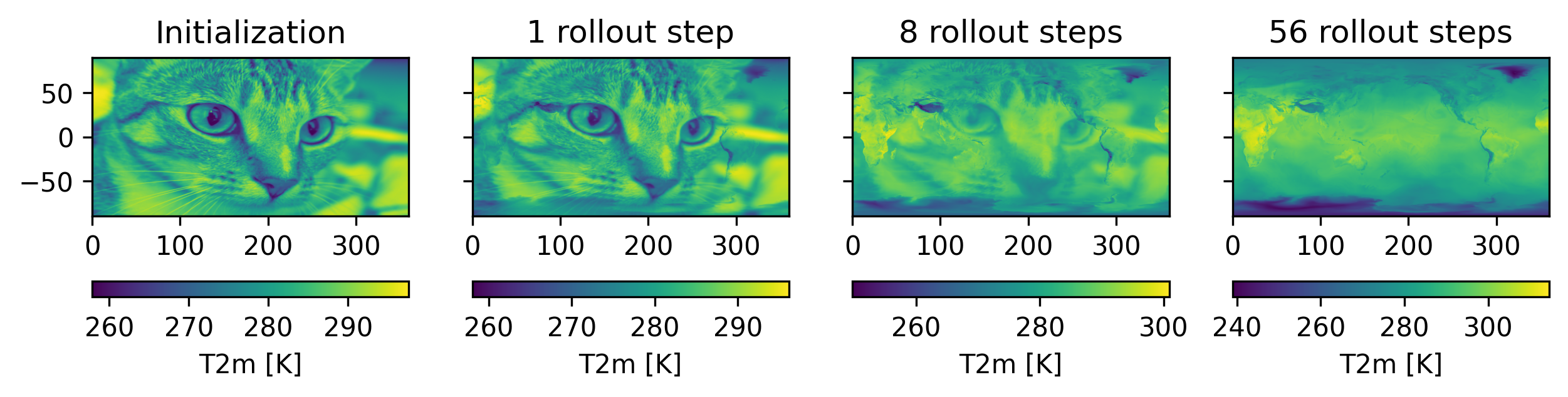}
        \caption{2-meter temperature predicted at different time stamps when (top) inputs are perturbed with an additive Gaussian random field with variance $\sigma^2$ (where $\sigma$ is the standard deviation of the physical variable) and correlation length of 10 pixels (equivalent to \ang{2.5}), (bottom) surface and atmospheric variables are initialized with a cat picture.}
        \label{fig:rollout_noise1_corr10_aurora}
    \end{figure}

    Then, we explore more specifically Aurora sensitivity to different types of input perturbations. Figure~\ref{fig:rollout_noise1_corr10_aurora} shows Aurora's denoising ability when surface and atmospheric variables are initialized with Gaussian random fields creating large-scale patterns (top) and a cat picture (bottom). In both cases, Aurora denoises the inputs within approximately 50 rollout steps and generates realistic rollout trajectories following seasonality. 

\begin{figure}[h]
        \centering
        \includegraphics[width=\linewidth]{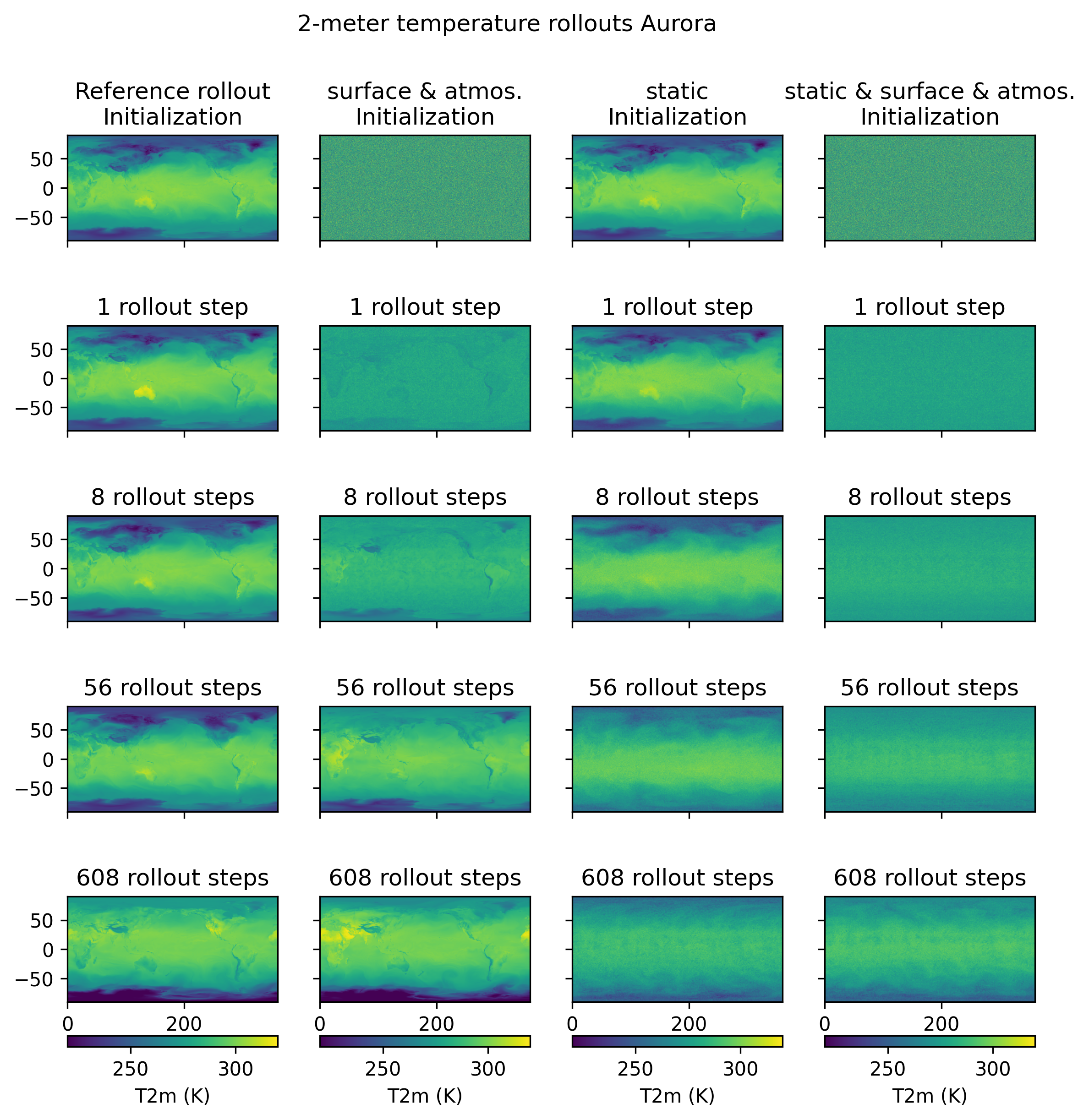}
        \caption{2-meter temperature predictions of four Aurora rollouts with different initialization, depicted at five rollout time steps. (1st column) Reference rollout. (2nd column) Initialize with random surface and atmospheric variables, static variables are left unchanged. (3rd column) Initialize with random static variables, surface and atmospheric variables are left unchanged. (4th column) Initialize with random static, surface, and atmospheric variables.}
        \label{fig:rollout_noise_decouple}
    \end{figure}

    Next, we decouple the influence of noise in static and dynamic (surface and atmospheric) variables. Figure \ref{fig:rollout_noise_decouple} shows Aurora rollouts initialized with static and/or dynamic variables set as random. Whenever static variables are the true physical ones, Aurora denoises the inputs, while input remain/become noisy when static variables are set as random. The following results show rollouts initialized on January 1, 2021.

\clearpage
\subsection{Initializing from pure noise}
\begin{figure}[h]
        \centering
        \includegraphics[width=1\linewidth]{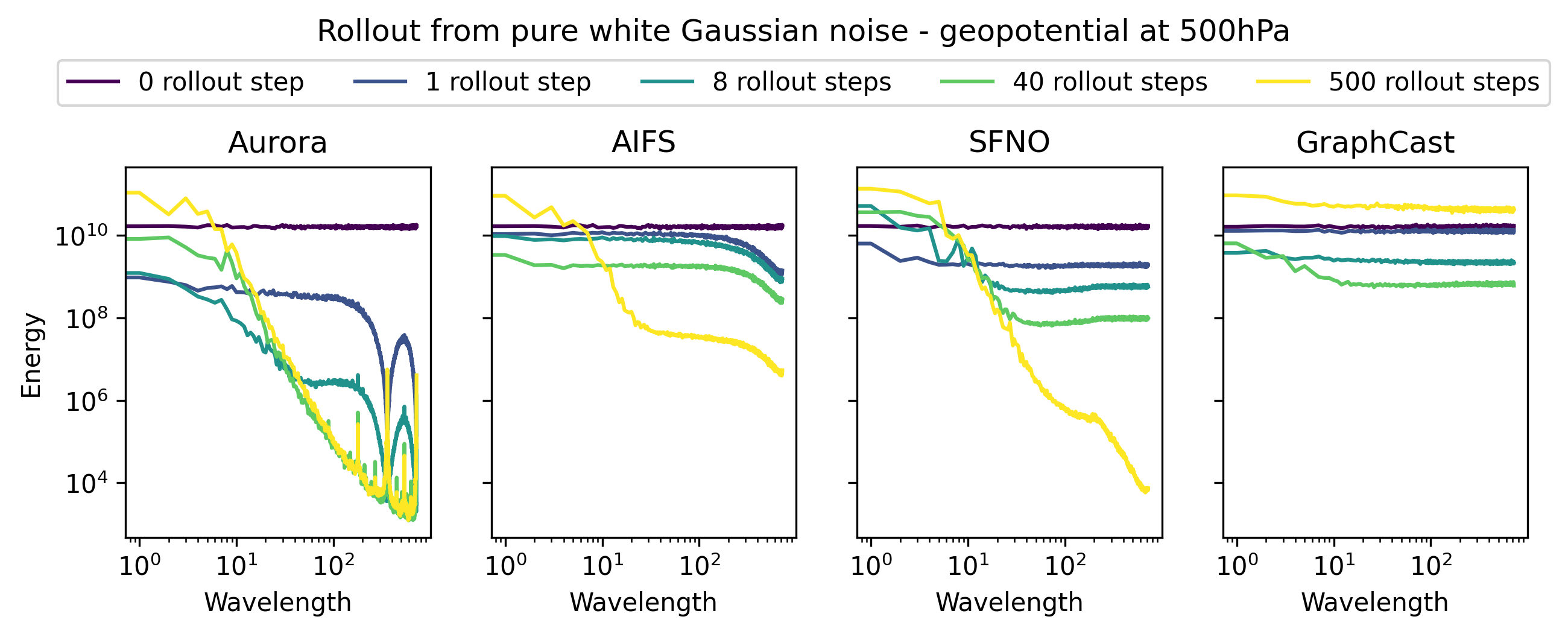}
        \caption{Energy spectra of geopotential at 500hPa for four models (Aurora, AIFS, SFNO, GraphCast) at five iterations of the rollout when initialized from pure noise (colored lines).}
        \label{fig:spectra_filtering}
    \end{figure}

\begin{figure}[h]
        \centering
        \includegraphics[width=\linewidth]{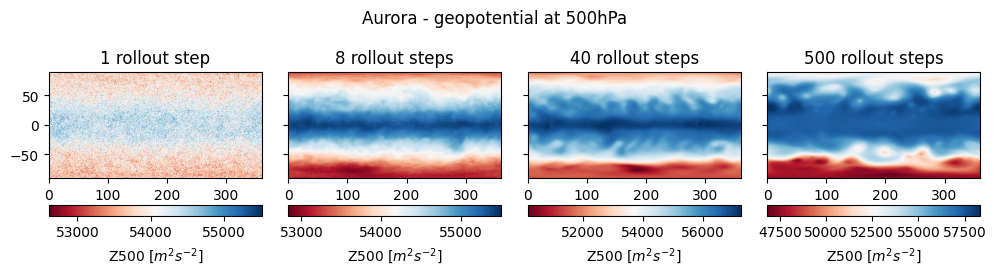}
        \includegraphics[width=\linewidth]{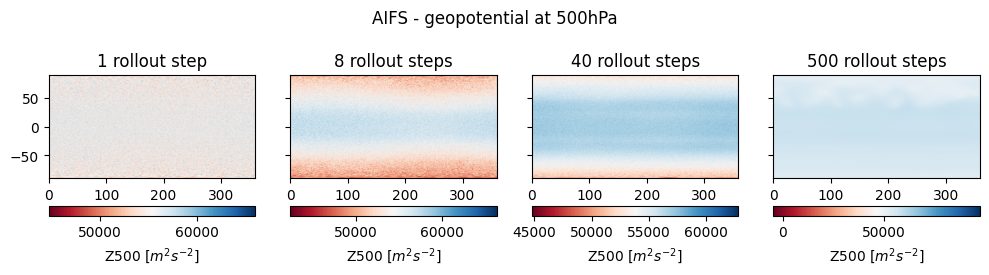}
        \includegraphics[width=\linewidth]{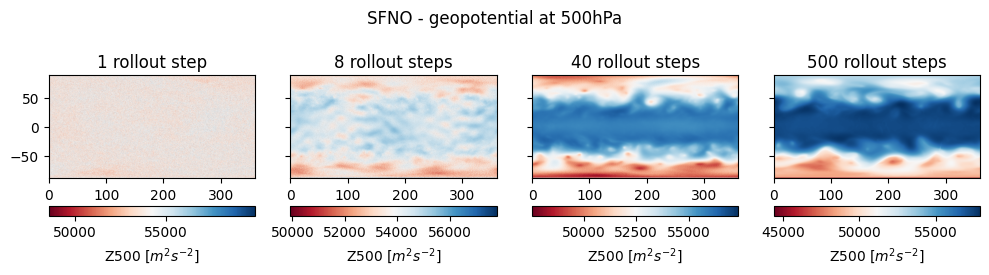}
        \includegraphics[width=\linewidth]{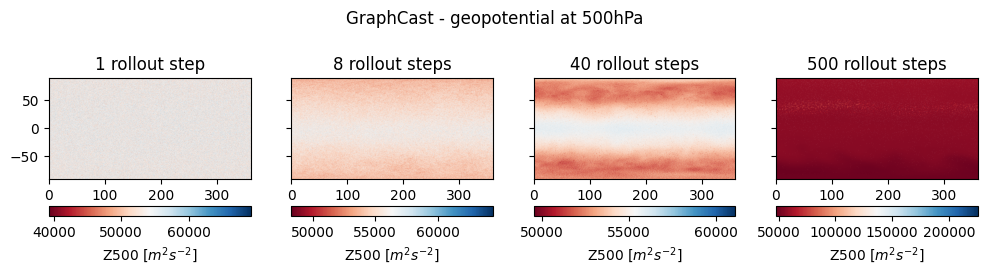}
        \caption{Geopotential at 500hPa when the rollout is initialized with pure noise, depicted at four rollout iterations, for Aurora (first row), AIFS (second row), SFNO (third row), GraphCast (fourth row).}
        \label{fig:pure_noise_maps}
    \end{figure}

    We initialize four models (Aurora, AIFS, SFNO, GraphCast) with pure white Gaussian noise, $\mathcal{N}(\mu, \sigma^2)$, where $\mu$ (resp. $\sigma$) is the mean (resp. standard deviation) of each physical variable. $\mu$ and $\sigma$ are scalar values and not two-dimensional fields to remove all their spatial content. Figure~\ref{fig:spectra_filtering} shows that each model performs a low-pass filtering of the noise. However, Aurora and SFNO denoise until realistic fields while AIFS and GraphCast show only unstructured noisy fields (Fig.~\ref{fig:pure_noise_maps}).
    
    \begin{figure}[h]
        \centering
        \includegraphics[width=\linewidth]{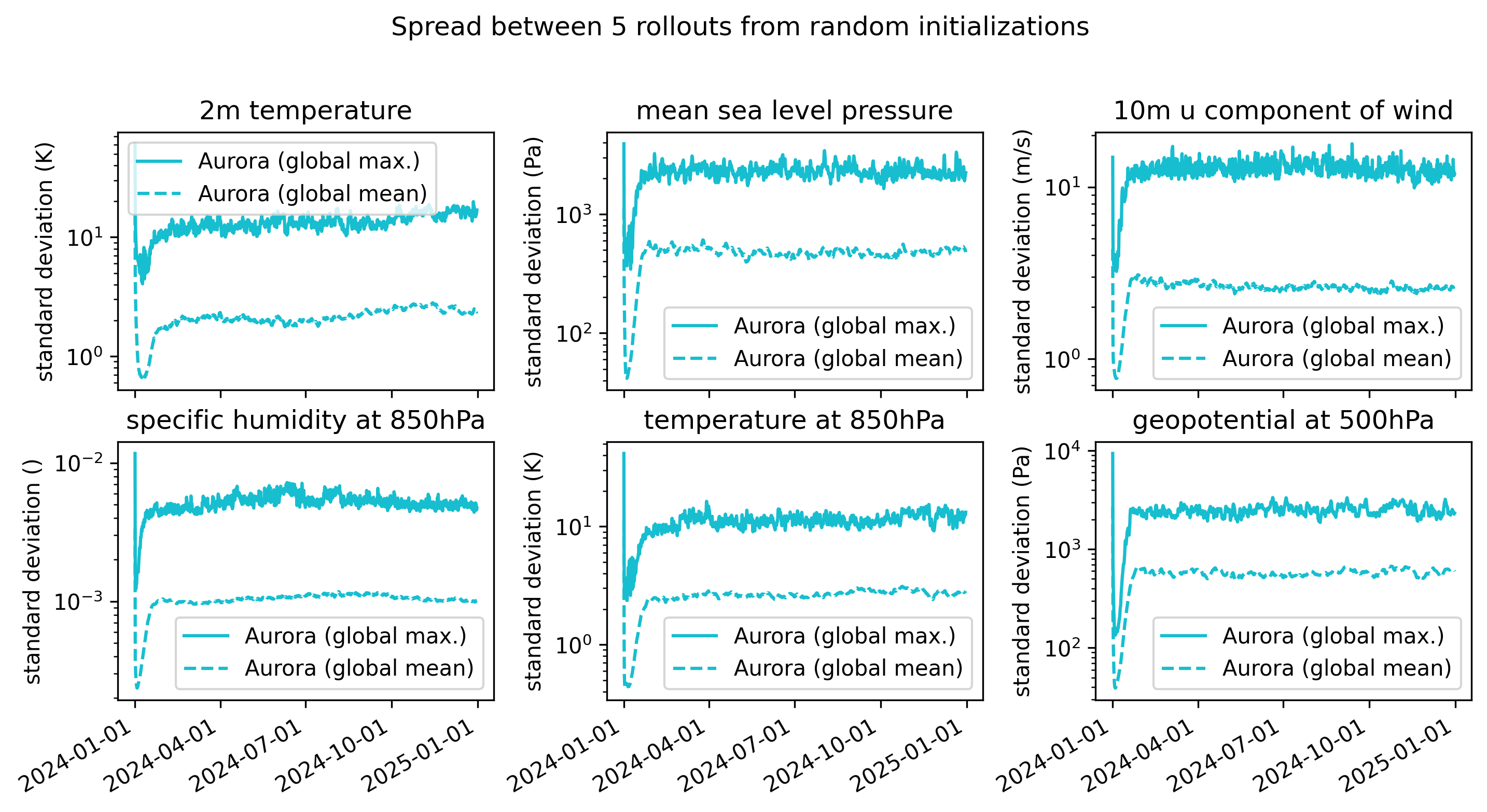}
        \caption{Standard deviation across five Aurora rollouts initialized from pure white Gaussian noise. The plot shows the average of the standard deviation over all pixels (dashed line) and the maximum over the pixels (solid line). SFNO produces spread of the same order of magnitude.}
        \label{fig:rollout_std}
    \end{figure}

\clearpage
\subsection{Memorization in stable models}
\label{sec:memorization}
\begin{figure}[h]
        \centering
        \includegraphics[width=0.45\linewidth]{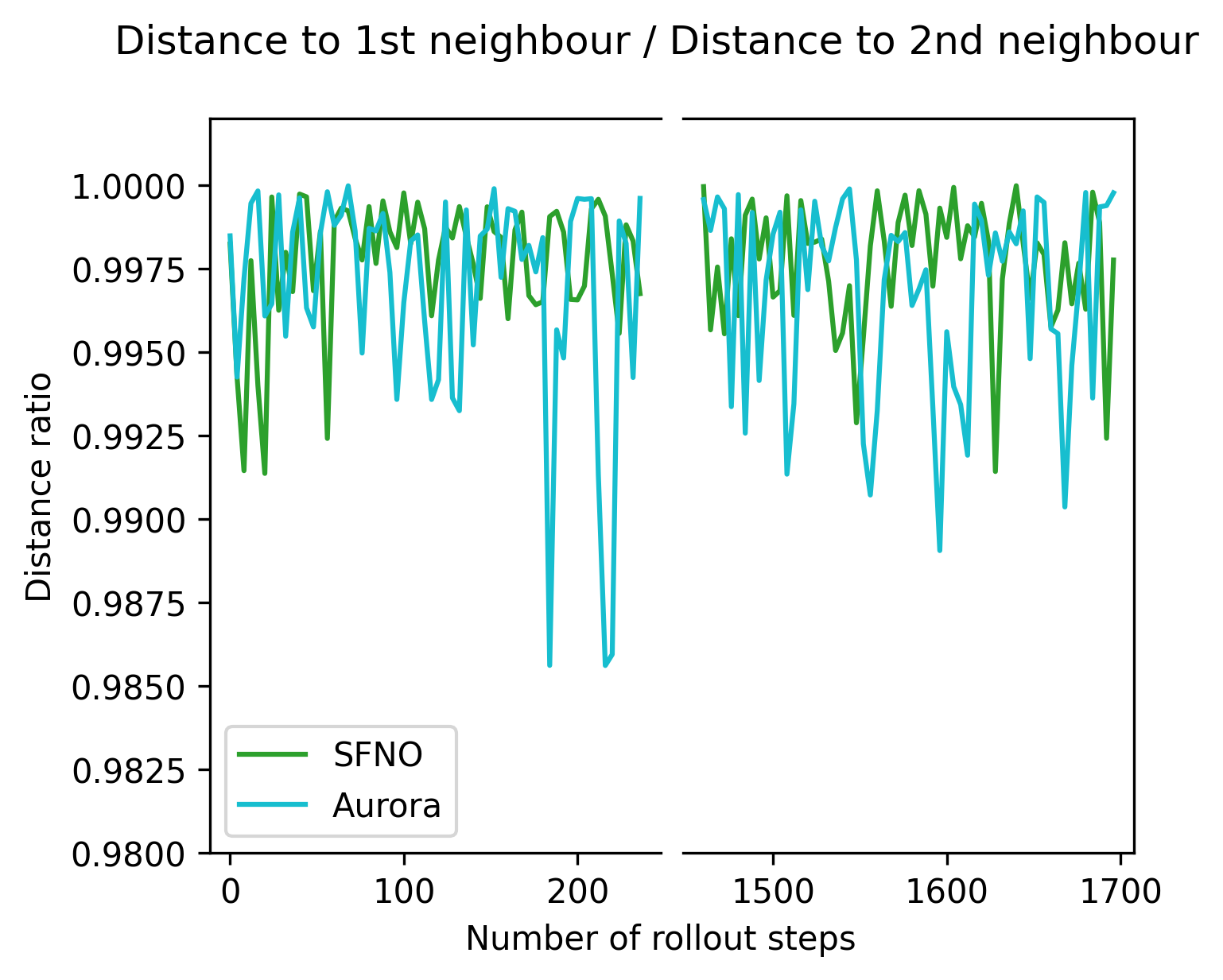}
        \caption{Distance ratio between model predictions and ERA5, computed over all four surface variables, five atmospheric variables at levels 500~hPa, 850~hPa, 1000~hPa.}
        \label{fig:distance_ratio}
    \end{figure}

    To test the memorization ability of stable models, we follow previous work \citep{bonnaire_why_2025, gu_memorization_2023, yoon2023diffusion} and say that a sample $X_t$ is memorized if it is significantly closer to its first neighbor $y_1(X_t)$ in the training dataset 
    \begin{equation}
        y_1(X_t) = \arg \min_{X_{\tau} \in \mathcal{T}} \| X_t - X_{\tau}\|, \quad 
        \mathcal{T} = \{ \tau %\in [\text{1979-01-01}, \text{2019-12-31}]
        , |\text{doy}(\tau) - \text{doy}(t)| \le 10 \ \text{days} \}
    \end{equation}
    than to its second neighbor $y_2(X_t) := \arg \min_{X_{\tau} \in \mathcal{T} - \{y_1\} } \| X_t - X_{\tau}\|$, as quantified by the distance ratio,
    \begin{equation}
        \dfrac{\|X_t - y_1(X_t)\|}{\|X_t - y_2(X_t)\|} \le 0.5 \text{ ,}
    \end{equation}
    where doy is day of the year.
    Note that the set $\mathcal{T}$ contains all days from the training period 1979-2019 that are within ten calendar days of the reference date $t$ to avoid mixing different seasons. To reduce the computational costs of looking up the entire ERA5 dataset, three atmospheric levels (500~hPa, 850~hPa, 1000~hPa) are selected among the 13 used for training. Figure~\ref{fig:distance_ratio} shows that the distance ratio is close to 1, indicating that no memorization is happening.

\clearpage
\section{Models trained from random initialization}
\subsection{Small Aurora models}
\label{appendix:minimal_aurora}
    Aurora\textsubscript{S} is trained on 8 GH200 GPUs for 60k steps, with no gradient checkpointing and no model parallelism. The per-GPU batch size is 1, optimizer is AdamW, the learning rate follows a linear warmup for 1k steps up to $5\times 10^{-4}$ followed by cosine decay. The training loss is latitude-weighted MAE with the same variable-specific weights as \cite{bodnar_foundation_2025}. Training time is between 13 hours (\ang{1.5} resolution) and 22 hours (\ang{0.25} resolution) due to the difference in data reading speed. The same batch size is used for all spatial resolutions for the sake of consistency between experiments but it could be increased for the lower resolution data.

\begin{figure*}[h]
        \centering
        \includegraphics[width=0.48\textwidth]{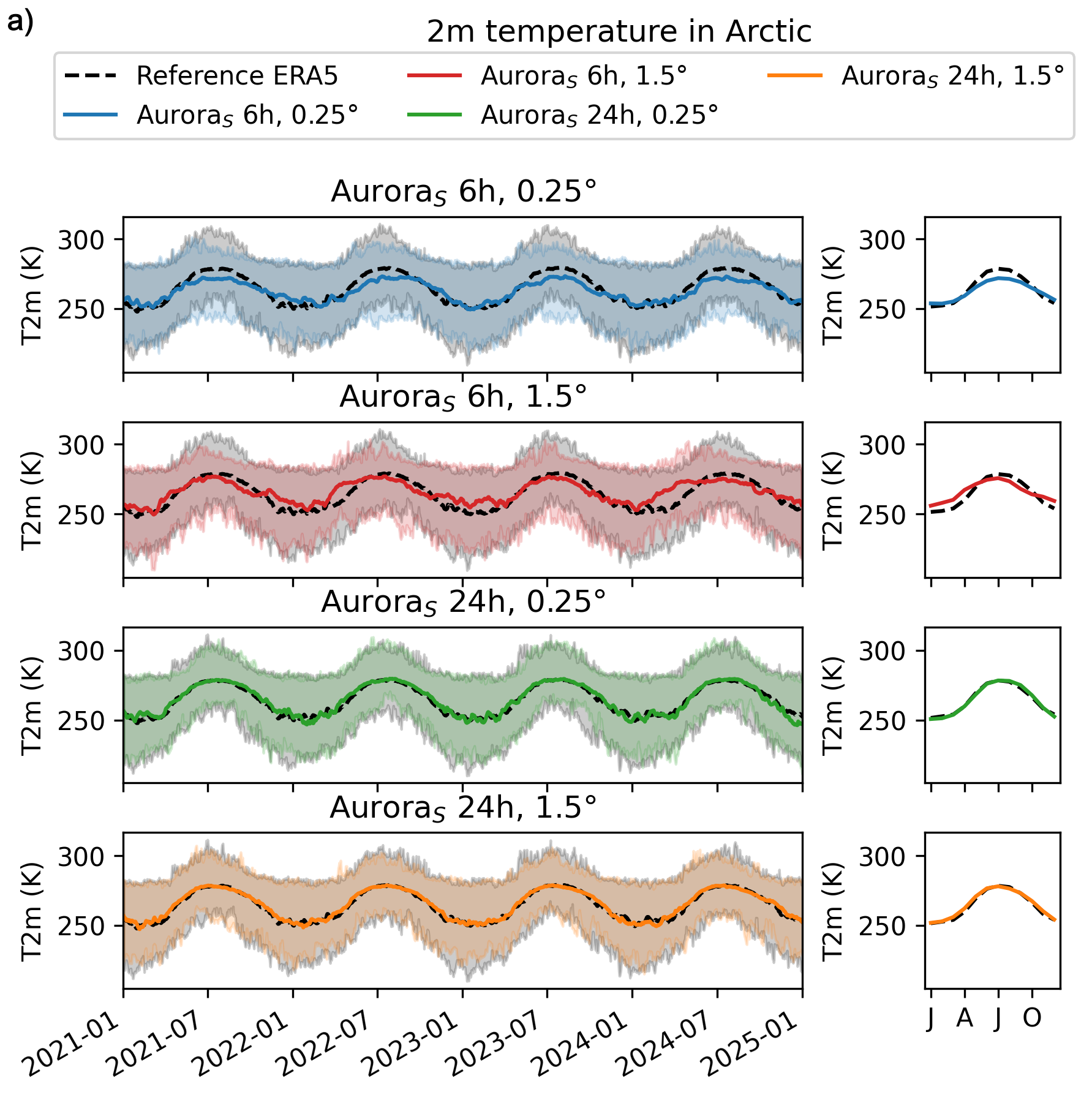}
        \includegraphics[width=0.48\textwidth]{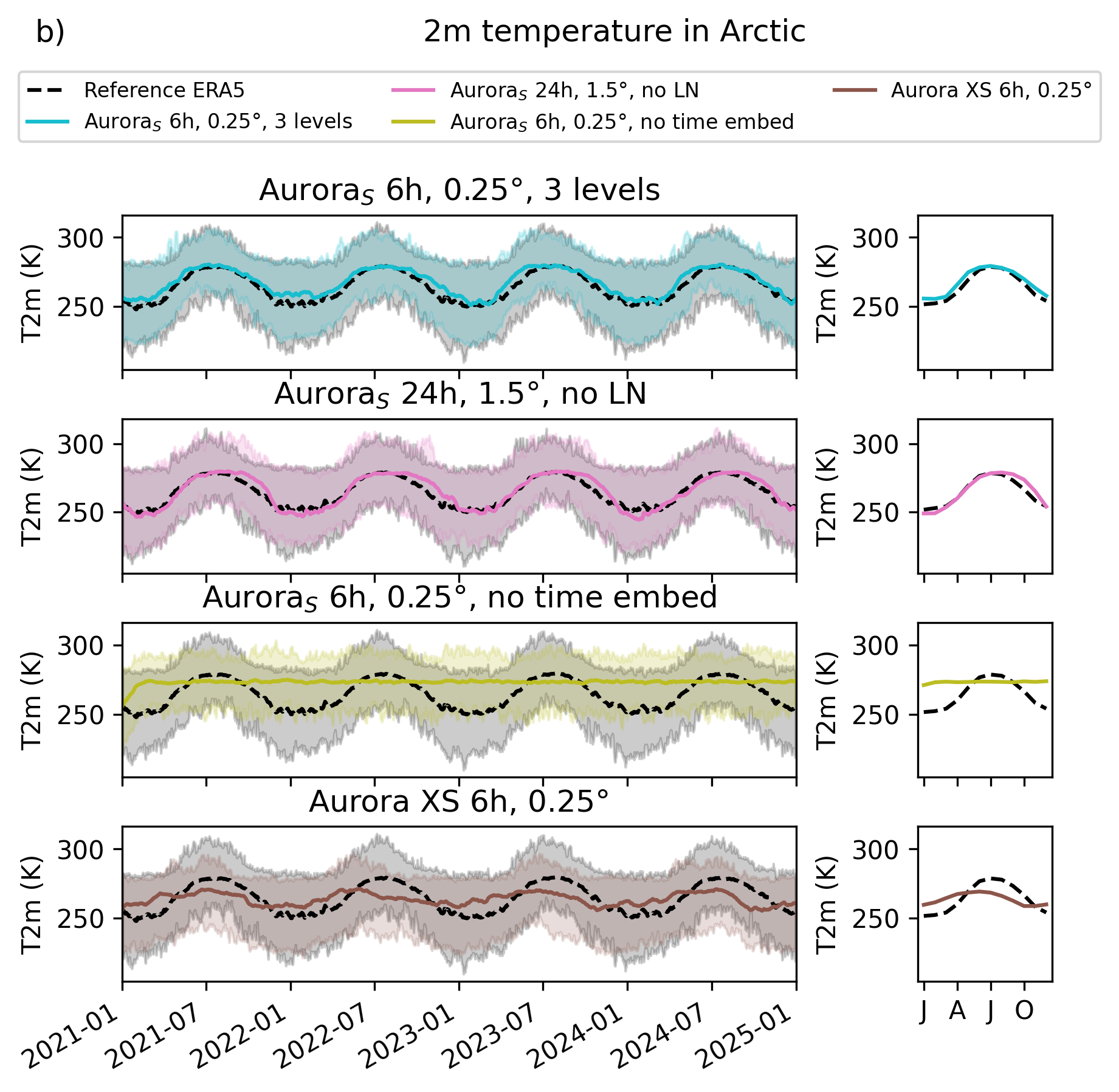}
        \caption{4-year rollout of 2-meter temperature for a) four small Aurora models trained from random initialization with different spatial resolutions (\ang{0.25} and \ang{1.5}) and lead times (6~hours and 24~hours), b) small Aurora with three atmospheric levels (500~hPa, 850~hPa, 1000~hPa) instead of 13, with RMS normalization instead of Layer Normalization, without time embedding, extra-small model (21M parameters). The right panels show the monthly seasonal cycle.}
        \label{fig:ablations_aurora_2t}
    \end{figure*}

\begin{figure*}[h]
        \centering
        \includegraphics[width=0.48\textwidth]{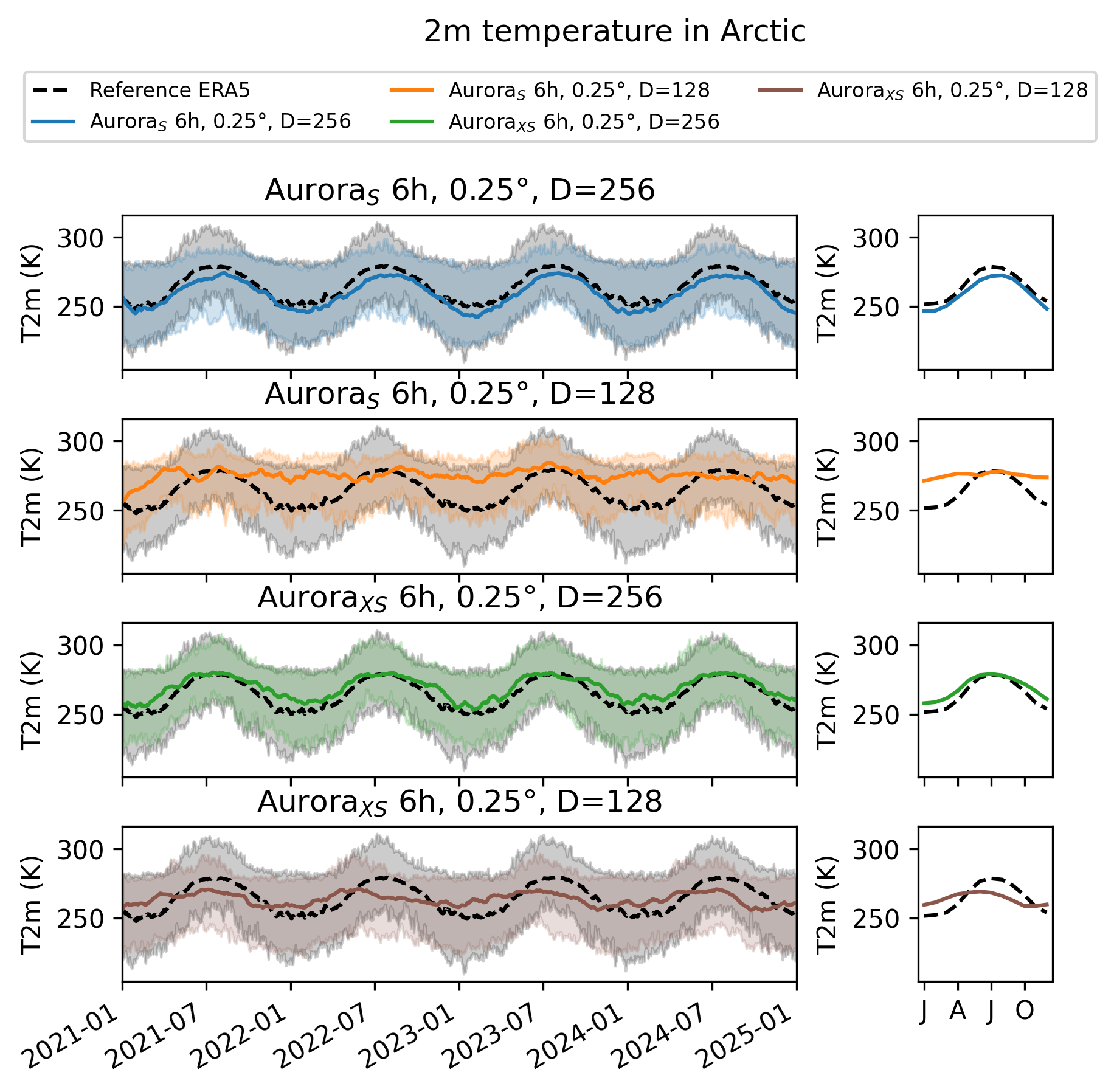}
        \caption{4-year rollout of 2-meter temperature for Aurora trained from random initialization. Aurora\textsubscript{S}: \textquote{deep} model (2-6-2 layers in the encoder part of the UNet backbone) and Aurora\textsubscript{XS}: \textquote{shallow} model (2-2-2 layers in the encoder part of the UNet backbone). Wide models (embedding dimension, D=256) and narrow models (D=128). The right panels show the monthly seasonal cycle.}
        \label{fig:ablations_aurora_2t_width_depth}
    \end{figure*}

    Table~\ref{tab:aurora_ablation_season} and Figure~\ref{fig:ablations_aurora_2t} present the results of the ablation experiments with skip connections, residual objective, and various spatiotemporal resolutions while Table~\ref{tab:season_aurora_width_depth} and Figure~\ref{fig:ablations_aurora_2t_width_depth} show the results of varying the width and depth.

    \begin{table*}[h]
      \caption{Number of days before the loss of seasonality, for the small Aurora models trained from random initialization. Aurora\textsubscript{S} 6h, \ang{0.25} is referred to as the baseline in the main text.}
      \label{tab:aurora_ablation_season}
      \centering
        \begin{small}
        \setlength{\tabcolsep}{3pt} % default is typically 6pt
            \begin{tabular}{lccccccccc}
            \toprule
            Model & T2m & U10m & MSLP & Z500 & T500 & Q500 & U300 & T100 & Q100 \\
            \midrule
            Aurora\textsubscript{S} no skip & 8 & 63 & 6 & 7 & 4 & 9 & 363 & 8 & 10 \\
            Aurora\textsubscript{S} residual & 91 & 86 & 82 & 58 & 41 & 54 & 46 & 54 & 37 \\
            Aurora\textsubscript{S} 6h, \ang{0.25} & $>$ 1460 & $>$ 1460 & $>$ 1460 & $>$ 1460 & $>$ 1460 & $>$ 1460 & $>$ 1460 & 101 & 19 \\
            Aurora\textsubscript{S} 6h, \ang{1.5} & $>$ 1460 & $>$ 1460 & $>$ 1460 & 308 & $>$ 1460 & $>$ 1460 & $>$ 1460 & 145 & $>$ 1460 \\
            Aurora\textsubscript{S} 24h, \ang{0.25} & $>$ 1460 & $>$ 1460 & $>$ 1460 & $>$ 1460 & $>$ 1460 & $>$ 1460 & $>$ 1460 & $>$ 1460 & $>$ 1460 \\
            Aurora\textsubscript{S} 24h, \ang{1.5} & $>$ 1460 & $>$ 1460 & $>$ 1460 & $>$ 1460 & $>$ 1460 & $>$ 1460 & $>$ 1460 & $>$ 1460 & $>$ 1460 \\
            \bottomrule
            \end{tabular}
        \end{small}
    \end{table*}

    \begin{table}[h]
    \centering
    \caption{Number of days before the loss of seasonality, for the small Aurora models trained from random initialization with varying width and depth. \textquote{Deep} models (2-6-2 layers in the encoder part of the UNet backbone) and \textquote{shallow} models (2-2-2 layers in the encoder part of the UNet backbone). Wide models (embedding dimension, D=256) and narrow models (D=128).}
    \label{tab:season_aurora_width_depth}
    \begin{small}
    \setlength{\tabcolsep}{3pt} % default is typically 6pt
    \begin{tabular}{lll|ccccccccc}
    \toprule
    Depth & Width & Params [M] & T2m & U10m & MSLP & Z500 & T500 & Q500 & U300 & T100 & Q100 \\
    \midrule
    2-6-2 & 256 & 112.8 & $>$ 1460 & $>$ 1460 & $>$ 1460 & $>$ 1460 & $>$ 1460 & $>$ 1460 & $>$ 1460 & 101 & 19 \\
    2-6-2 & 128 & 28.4 & 67 & 275 & 115 & 66 & 69 & 15 & 183 & 187 & 49 \\
    2-2-2 & 256 & 83.4 & $>$1460 & $>$1460 & $>$1460 & $>$1460 & $>$1460 & 69 & $>$1460 & 150 & 34 \\
    2-2-2 & 128 & 21.0 & 180 & $>$1460 & 7 & 227 & 95 & 175 & 19 & 150 & 147 \\
    \bottomrule
    \end{tabular}
    \end{small}
    \end{table}

    \begin{figure*}[h]
        \centering
        \includegraphics[width=\textwidth]{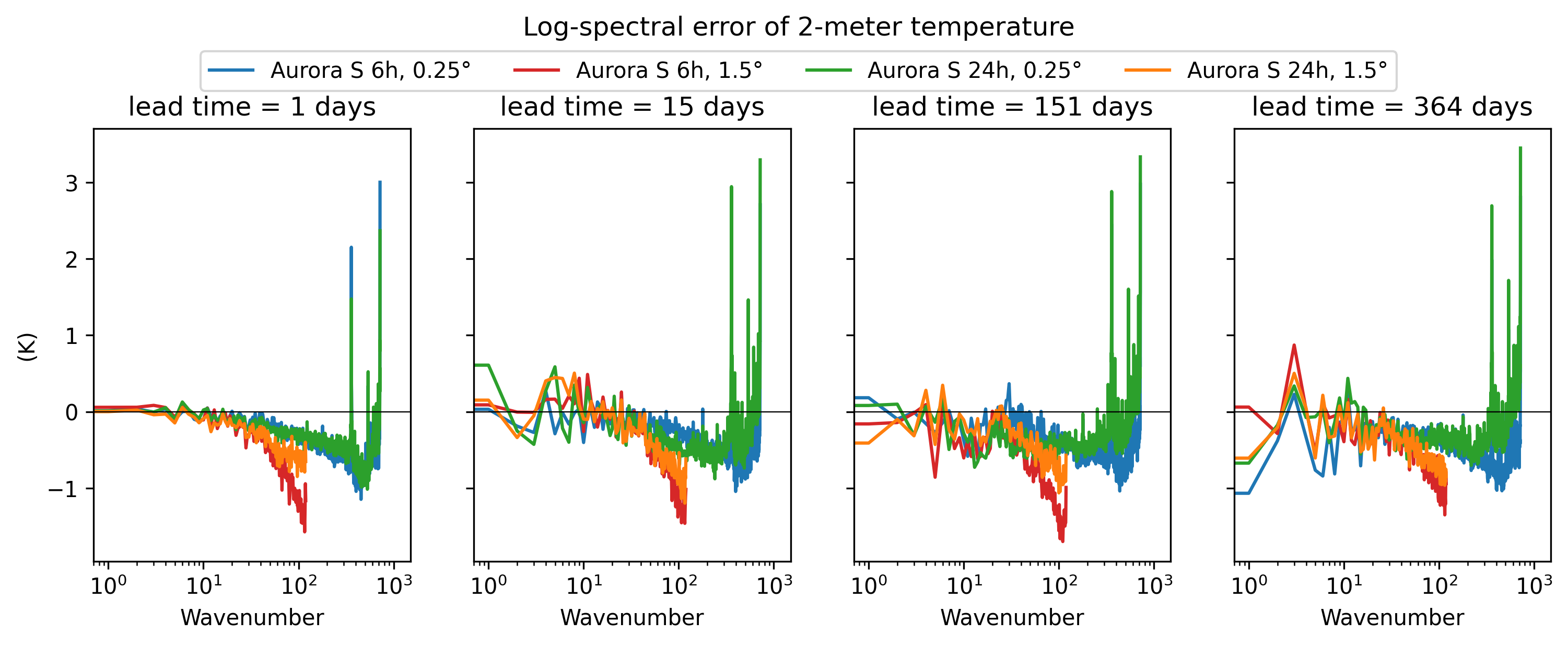}
        \caption{Difference (in log scale) between the Fourier spectrum of the Aurora\textsubscript{S} models and the reference spectrum for ERA5, at different lead times.}
        \label{fig:spectra_from_scratch}
    \end{figure*}

\subsection{Small SFNO models trained from random initialization}
\label{appendix:minimal_sfno}
    The training dataset is the same as used for the small Aurora training runs: ERA5 at \ang{0.25} resolution and 6-hour sampling, with training period 1979-2019, validation 2020, test 2021-2024. The lead time is six hours and the models were trained for 8 epochs (60k steps on 8 GPUs) without rollout finetuning. We use the official NVIDIA github repository for training (\url{https://github.com/NVIDIA/makani}) and made a few changes to the reference configuration file (\verb|config/sfnonet.yaml|) to create our baseline SFNO:
    \begin{itemize}
        \item We increased the scaling factor from 3 to 4, which coarsens the internal grid on which the spherical harmonic transforms are computed (from $240 \times 480$ to $180 \times 360$).
        \item We reduced the embedding dimension from 384 to 256. These two choices decrease the number of model parameters from 289M to 97M, bringing SFNO close to the small Aurora we trained.
        \item The learning rate scheduler is also modified from a Reduce on Plateau to a cosine decay preceded by a linear warmup for 1000 steps up to the reference $10^{-3}$. This ensures the same learning rate scheduler across all training runs. 
        \item We omitted four surface variables for comparability with Aurora: 100m eastward and northward wind velocity, surface pressure, total column water vapour.
    \end{itemize}

    \begin{table}[h]
        \centering
        \caption{Number of days before the loss of seasonality, for the SFNO ablations. Rollout initialized on 2021-01-01. SFNO baseline denotes our small model trained from random initialization.}
        \label{tab:sfno_ablation_season}
        \setlength{\tabcolsep}{5pt} % default is typically 6pt
        \begin{small}
        \begin{tabular}{lccccccccc}
            \toprule
            Model & T2m & U10m & MSLP & Z500 & T500 & Q500 & U300 & T100 & Q100 \\
            \midrule
            SFNO baseline & $>$ 1460 & $>$ 1460 & $>$ 1460 & $>$ 1460 & 181 & $>$ 1460 & 908 & 166 & $>$ 1460 \\
            SFNO no skip & 78 & $>$ 1460 & $>$ 1460 & 97 & 59 & 96 & 113 & 114 & 146 \\
            SFNO residual & $>$ 1460 & $>$ 1460 & $>$ 1460 & $>$ 1460 & $>$ 1460 & $>$ 1460 & $>$ 1460 & $>$ 1460 & $>$ 1460 \\
            \makecell{SFNO 360×720, \\ \quad D=184} & $>$ 1460 & $>$ 1460 & $>$ 1460 & $>$ 1460 & $>$ 1460 & 66 & $>$ 1460 & $>$ 1460 & $>$ 1460 \\
            \bottomrule
        \end{tabular}
        \end{small}
    \end{table}

\clearpage
\section{Climate extremes from rollouts}
\label{appendix:extreme_events}
    \paragraph{Data and regions.} We analyze 2-meter temperature from ERA5 (0.25\textdegree, subsampled to 6-hourly). Five regions are selected: Central Europe (45--55\textdegree N, 5--20\textdegree E), Western US (30--50\textdegree N, 105--125\textdegree W), East Asia (25--45\textdegree N, 110--135\textdegree E), SE Australia (25--40\textdegree S, 140--155\textdegree E), and the Amazon (15\textdegree S--5\textdegree N, 45--70\textdegree W). At each 6-hourly time step, we extract the regional spatial maximum and minimum of 2-meter temperature.
    The compact regions reduce variations in latitude and elevation but do not eliminate geographic heterogeneity or seasonality; they serve as illustrative regional diagnostics.

    \paragraph{Definition of extremes.} Thresholds are computed from the ERA5 climatological period 1990--2019: for each region, the 90th (P90) and 10th (P10) percentiles of the regional spatial maximum and minimum time series are computed, respectively. A \emph{hot event} occurs when the regional spatial maximum exceeds P90; a \emph{cold event} occurs when the regional spatial minimum falls below P10. This definition of all-season extremes tests the absolute-temperature tails of free-running rollouts, retaining mean-state and seasonal-cycle biases. Exceedance rates are evaluated over 2021--2024 for both ERA5 and the Aurora, SFNO, and DLESyM rollouts. 

\begin{figure*}[h]
        \centering
        \includegraphics[width=\textwidth]{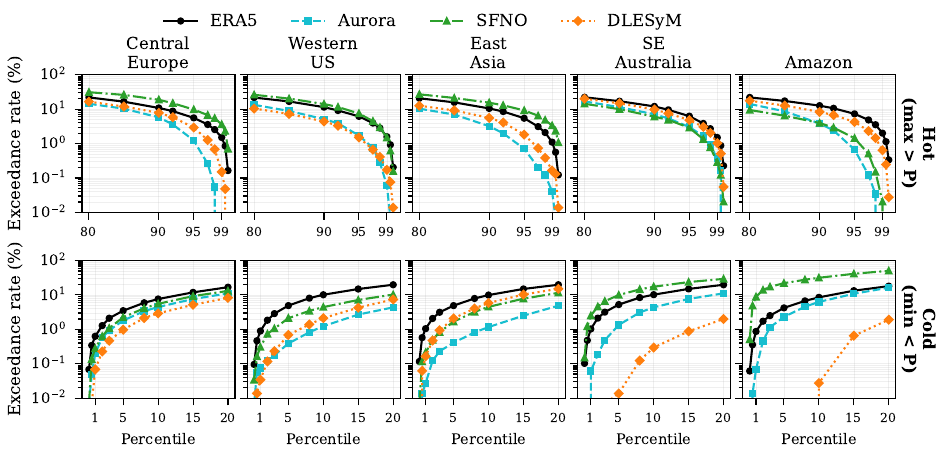}
        \caption{Multi-threshold exceedance rates. For thresholds defined from the ERA5 distribution (P80--P99.9 for hot, P0.1--P20 for cold), the fraction of time steps exceeding each threshold is shown.}
        \label{fig:exceedance}
    \end{figure*}

    \paragraph{Frequency analysis.} We analyze the frequency of hot and cold events during the four-year time window for Aurora, SFNO, and DLESyM and compare it with ERA5. In this analysis, the threshold defining an extreme event is varied from the 80th to the 99.9th percentile for hot events (from 0.1st to 20th percentile for cold events). Results show that no model consistently reproduces extreme event frequency of ERA5 and no model is clearly better than the others (Fig.~\ref{fig:exceedance}). At the 90th percentile, the ratio of predicted hot events compared to ERA5 is between 0.25 and 1.39, with a clear tendency for ratios lower than 1.
    Zhang et al.~\cite{zhang_physicsbased_2026} also find that AI weather models underestimate the frequency and intensity of record-breaking extremes, but in a different setting: initialized forecasts up to 10 days from GraphCast, Pangu-Weather, and FuXi, whereas we evaluate multi-year statistics of free-running rollouts.

\begin{figure}[h]
        \centering
        \includegraphics[width=0.75\linewidth]{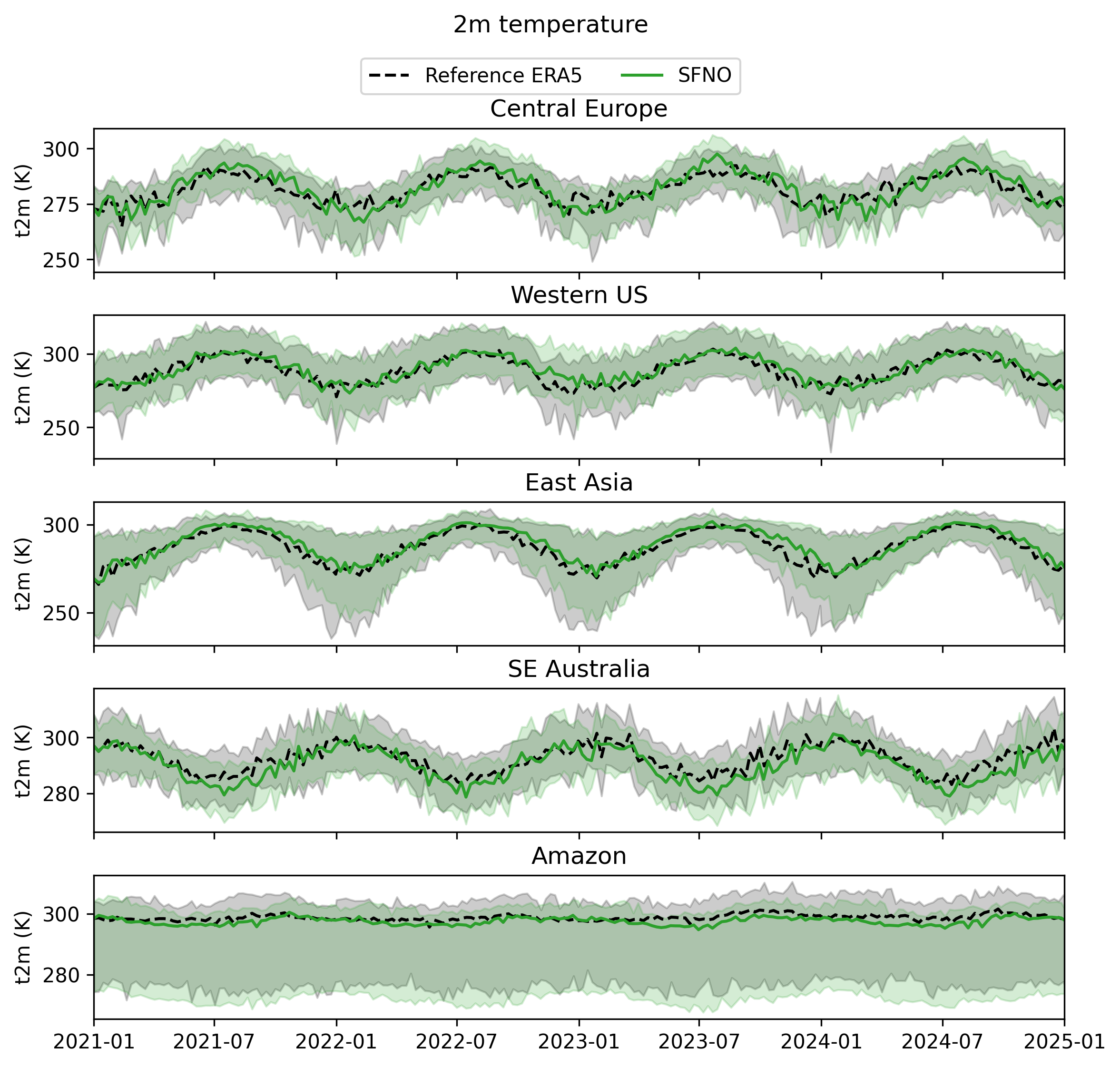}
        \caption{2-meter temperature in five regions for SFNO initialized on January 1, 2021. The solid lines show the regional average, the shaded areas extend from the minimal to the maximal temperature over the region. The subplots show the first two months. Time series are depicted at a weekly resolution for easier visualization.}
        \label{fig:rollout_sfno_extreme}
    \end{figure}

    \paragraph{Calendar-day thresholds.} To separate weather variability from the seasonal cycle, we also use thresholds that depend on the calendar day. The 6-hourly regional maxima (minima) are aggregated to daily values, and P90 (P10) is computed from ERA5 1990--2019 within a $\pm$7-day window around each calendar day. Table~\ref{tab:extreme_calendar_day} reports the number of hot and cold event days over 2021--2024. Most model--region combinations (21 of 30) underestimate the ERA5 event frequency, and overestimation occurs mostly for SFNO (hot events in the Northern regions, cold events in the Southern regions, where SFNO has a cold bias; Fig.~\ref{fig:rollout_sfno_extreme}). Our conclusions are therefore not specific to the all-season thresholds.

    \begin{table}[h]
      \caption{Number of days in 2021--2024 (1,461 days) with a hot event (daily regional maximum above the calendar-day P90) or a cold event (daily regional minimum below the calendar-day P10), with thresholds from ERA5 1990--2019.}
      \label{tab:extreme_calendar_day}
      \begin{center}
        \begin{small}
          \begin{tabular}{lrrrrrrrr}
            \toprule
             & \multicolumn{4}{c}{Hot events} & \multicolumn{4}{c}{Cold events} \\
            \cmidrule(lr){2-5} \cmidrule(lr){6-9}
            Region & ERA5 & Aurora & SFNO & DLESyM & ERA5 & Aurora & SFNO & DLESyM \\
            \midrule
            Central Europe & 273 & 115 & 364 & 130 & 81 & 38 & 51 & 18 \\
            Western US & 233 & 166 & 478 & 16 & 103 & 4 & 18 & 11 \\
            East Asia & 299 & 130 & 469 & 38 & 94 & 212 & 8 & 29 \\
            SE Australia & 143 & 147 & 36 & 237 & 130 & 64 & 235 & 3 \\
            Amazon & 237 & 8 & 2 & 116 & 124 & 127 & 1,177 & 0 \\
            \bottomrule
          \end{tabular}
        \end{small}
      \end{center}
    \end{table}

\end{document}